\documentclass[10pt,twocolumn,letterpaper]{article}

\usepackage[pagenumbers]{cvpr} % To force page numbers, e.g. for an arXiv version

\usepackage[dvipsnames]{xcolor}

\definecolor{cvprblue}{rgb}{0.21,0.49,0.74}
\usepackage[pagebackref,breaklinks,colorlinks,citecolor=cvprblue]{hyperref}

\def\paperID{8238} % *** Enter the Paper ID here
\def\confName{CVPR}
\def\confYear{2024}

\usepackage{amsmath,amsfonts,bm}

\def\eqref#1{Equation~(\ref{#1})}
\def\1{\bm{1}}

\def\eps{{\epsilon}}

\def\vmu{{\bm{\mu}}}

\def\ve{{\bm{e}}}

\def\vp{{\bm{p}}}
\def\vq{{\bm{q}}}

\def\vs{{\bm{s}}}

\def\vv{{\bm{v}}}
\def\vw{{\bm{w}}}
\def\vx{{\bm{x}}}
\def\vy{{\bm{y}}}
\def\vz{{\bm{z}}}

\DeclareMathAlphabet{\mathsfit}{\encodingdefault}{\sfdefault}{m}{sl}
\SetMathAlphabet{\mathsfit}{bold}{\encodingdefault}{\sfdefault}{bx}{n}

\DeclareMathOperator*{\argmin}{arg\,min}

\usepackage{amssymb}
\usepackage{amsthm}
\usepackage{cancel}
\usepackage{amsmath}
\usepackage[normalem]{ulem} % [normalem] option prevents ulem from changing the default behavior of \emph command
\usepackage{mathtools}
\usepackage{multirow}
\usepackage{tabularx}
\usepackage{makecell}
\usepackage{afterpage}
\usepackage{arydshln} %for \cdashline, \hdashline
\usepackage{bbm} % for \mathbbm{1}

\usepackage{tikz}
\definecolor{crimson}{rgb}{0.85, 0.0, 0.0}
\definecolor{grass}{rgb}{0.1, 0.7, 0.1}

\renewcommand\checkmark[1][]{%
  \tikz[scale=0.4,#1]{\fill(0,.35) -- (.25,0) -- (1,.7) -- (.25,.15) -- cycle;}%
}
\newcommand\crossmark[1][]{%
  \tikz[scale=0.4,#1]{
    \fill(0,0)--(0.1,0) .. controls (0.5,0.4) .. (1,0.7)--(0.9,0.7) ..  controls (0.5,0.5) ..(0,0.1) --cycle;
    \fill(1,0.1)--(0.9,0.1) .. controls (0.5,0.3) .. (0,0.7)--(0.1,0.7) .. controls (0.5,0.4) ..(1,0.2) --cycle;
  }%
}
\newtheorem{propo}{Proposition}
\newtheorem{defi}{Definition}

\def\Obs{{\scriptstyle O}}
\newcommand{\Dg}{\Delta g}
\newcommand{\D}{\mathbf{D}}
\newcommand{\Grad}{\nabla}
\newcommand{\Lie}{\mathcal{L}}
\newcommand{\basis}{\hat{\ve}}
\newcommand{\tablefootnotemarkone}{\mathsection}

\newcommand{\notavailable}{(n/a)}
\newcommand{\LinvKernel}{K}
\newcommand{\BrownianKernel}{\mathcal{B}}

\def\sceneFrame{s}
\def\eefFrame{e}
\def\refFrame{d}
\def\Oscene{\Obs_{\sceneFrame}}
\def\Ograsp{\Obs_{\eefFrame}}
\def\gref{g_{\eefFrame \refFrame}}
\def\pref{\vp_{\eefFrame \refFrame}}
\def\Rref{R_{\eefFrame \refFrame}}
\def\FrameSelection{P}
\def\OriginSelection{P}

\newcommand{\propNo}{\checkmark[grass, scale=1.]}
\newcommand{\propYes}{\crossmark[crimson, scale=0.7]}

\newcommand{\intOver}[2][]{%
  \ifx\\#1\\%
  \else
    \hspace{#1}%
  \fi
  \int\limits_{\scriptscriptstyle #2}%
  \ifx\\#1\\%
  \else
    \hspace{#1}%
  \fi
}
\newcommand{\intOverSEthree}{\intOver[-4pt]{SE(3)}}
\newcommand{\intOverRthree}{
    \hspace{-3pt}\int_{\mathbb{R}^3}\hspace{-4pt}
}
\newcommand{\cross}{\wedge}

\def\Supp{Supp.}
\def\Fig{Fig.}
\def\Sec{Sec.}
\def\Table{Tab.}
\def\Proposition{Proposition}
\renewcommand{\eqref}{\cref}

\title{Diffusion-EDFs: Bi-equivariant Denoising Generative \\Modeling on SE(3) for Visual Robotic Manipulation}

\author{%
  Hyunwoo~Ryu$^{1}$,\ \ Jiwoo~Kim$^{1}$,\ \ Hyunseok~An$^{1}$,\ \ Junwoo~Chang$^{1}$,\ \ Joohwan~Seo$^{2}$,\\
  Taehan~Kim$^{3}$,\ \ Yubin~Kim$^{4}$,\ \ Chaewon~Hwang$^{5,6}$,\ \ Jongeun~Choi$^{1,2}$\thanks{Corresponding author: Jongeun Choi (jongeunchoi@yonsei.ac.kr)}, \ \ Roberto~Horowitz$^{2}$\\ 
  \normalsize $^{1}$Yonsei University,\ \ $^{2}$University of California, Berkeley,\ \ $^{3}$Samsung Research,\\
  \normalsize $^{4}$Massachusetts Institute of Technology, \ \ $^{5}$Ewha Womans University, \ \ $^{6}$Work done at Yonsei University\\
  \texttt{\small\{tomato1mule,nfsshift9801,junwoochang,hs991210,jongeunchoi\}@yonsei.ac.kr} \\
  \texttt{\small\{joohwan\_seo,  horowitz\}@berkeley.edu}\\
  \texttt{\small taehan11.kim@samsung.com}, \quad \texttt{\small ybkim95@media.mit.edu}, \quad \texttt{\small hcw0221@ewhain.net}
}

\begin{document}
\maketitle
\begin{abstract}
  Diffusion generative modeling has become a promising approach for learning robotic manipulation tasks from stochastic human demonstrations.
  In this paper, we present \emph{Diffusion-EDFs}, a novel $SE(3)$-equivariant diffusion-based approach for visual robotic manipulation tasks. 
  We show that our proposed method achieves remarkable data efficiency, requiring only 5 to 10 human demonstrations for effective end-to-end training in less than an hour. 
  Furthermore, our benchmark experiments demonstrate that our approach has superior generalizability and robustness compared to state-of-the-art methods.
  Lastly, we validate our methods with real hardware experiments.
  % The codes will be released upon acceptance.
  %By incorporating $SE(3)$-equivariance into our proposed model architectures, we show that our proposed method is %significantly much more data-efficient ($5\sim 10$ task demonstrations are enough for end-to-end training) and %generalizable than previous diffusion based manipulation methods. 
  % Codes are available at: \url{https://github.com/tomato1mule/diffusion_edf}
  Project Website: \url{https://sites.google.com/view/diffusion-edfs/home}
\end{abstract}    
\section{Introduction}
\label{sec:intro}

Diffusion models are increasingly being recognized as superior methods for modeling stochastic and multimodal policies \citep{janner2022planning,black2023training,pearce2023imitating,chi2023diffusion,simeonov2023shelving,liu2022structdiffusion,urain2022se3dif, ajay2022conditional, brehmer2023edgi, brehmer2023geometric,mishra2023reorientdiff}. 
In particular, \emph{$SE(3)$-Diffusion Fields} \citep{urain2022se3dif} apply diffusion-based learning on the $SE(3)$ manifold to generate grasp poses of the end-effector.
However, these methods require numerous demonstrations and do not generalize well on novel task configurations that are not provided during training.

% On the other hand, extensive work has shown that incorporating equivariance can significantly enhance the data efficiency, generalizability, and robustness in robot learning \citep{zeng2020transporter,simeonov2022neural,simeonov2023se,chun2023local,huang2022equivariant,huang2023edge,wang2022equivariant,ryu2023equivariant, kim2023robotic,lin2023mira, brehmer2023edgi, brehmer2023geometric}. 
In contrast, equivariant methods are well known for their data efficiency and generalizability in learning robotic manipulation tasks
\citep{zeng2020transporter,simeonov2022neural,simeonov2023se,chun2023local,huang2022equivariant,huang2023edge,wang2022equivariant,ryu2023equivariant, kim2023robotic,lin2023mira, brehmer2023edgi, brehmer2023geometric}.
In particular, several recent works explore the use of $SE(3)$-equivariant models for learning 6-DoF manipulation tasks with point cloud observations \citep{chun2023local,simeonov2022neural,simeonov2023se,ryu2023equivariant,huang2023edge}. 
\emph{Equivariant Descriptor Fields} (EDFs) \citep{ryu2023equivariant} achieve data-efficient end-to-end learning on 6-DoF visual robotic manipulation tasks by employing $SE(3)$ \emph{bi-equivariant} \citep{ryu2023equivariant,kim2023robotic} energy-based models.
% \emph{Equivariant Descriptor Fields} (EDFs) \citep{ryu2023equivariant} achieve few-shot level data efficiency and generalizability in 6-DoF manipulation learning by employing an $SE(3)$ \emph{bi-equivariant} \citep{ryu2023equivariant,kim2023robotic} energy-based model (EBM). 
% EDFs are distinguished from other $SE(3)$-equivariant methods in that they can be trained and deployed in a fully end-to-end manner without requiring any pre-training or object segmentation.
However, EDFs require more than 10 hours to learn from only a few demonstrations due to the inefficient training of energy-based models.
In this paper, we present Diffuion-EDFs, a diffusion-based alternative to EDFs with a significantly reduced training time ($\times$15 faster). 
% Similarly to EDFs, our method is designed to be bi-equivariant (see \Supp~\ref{appndx:bi-equivariance}), enabling fully end-to-end training from only $5\small{\sim}10$ human demonstrations without any pre-training and object segmentation.
% Due to the bi-equivariance
Similarly to EDFs, we exploit the bi-equivariance (see \Supp~\ref{appndx:bi-equivariance}) and locality of robotic manipulation tasks in our method design.
This enables our method to be trained end-to-end from only $5\small{\sim}10$ human demonstrations without requiring any pre-training and object segmentation, yet are highly generalizable to out-of-distribution object configurations.
We validate Diffusion-EDFs through simulation and real-robot experiments.

\vspace{-1\baselineskip}
\paragraph{Our contributions are summarized as follows:}
\begin{itemize}
    \item This is the first work to address an $SE(3)$-equivariant diffusion model for visual robotic manipulation. We provide novel theories and practices to achieve equivariance for point cloud-conditioned diffusion models on $SE(3)$.
    \item Our method significantly reduces the training time of previous work, EDFs \citep{ryu2023equivariant}, while maintaining their end-to-end trainability, data-efficiency, and generalizability.
    % \item We propose a novel hierarchical architecture to incorporate a wide receptive field. This enables our model to understand scene-level contexts, distinguishing it from previous $SE(3)$-equivariant manipulation methods.
    \item We propose a novel hierarchical architecture to incorporate a wide receptive field. This enables our model to understand contexts at the scene level, distinguishing it from previous object-centric $SE(3)$-equivariant methods.
\end{itemize}

\begin{figure*}[ht!]
  \centering
  \includegraphics[width=0.95\textwidth]{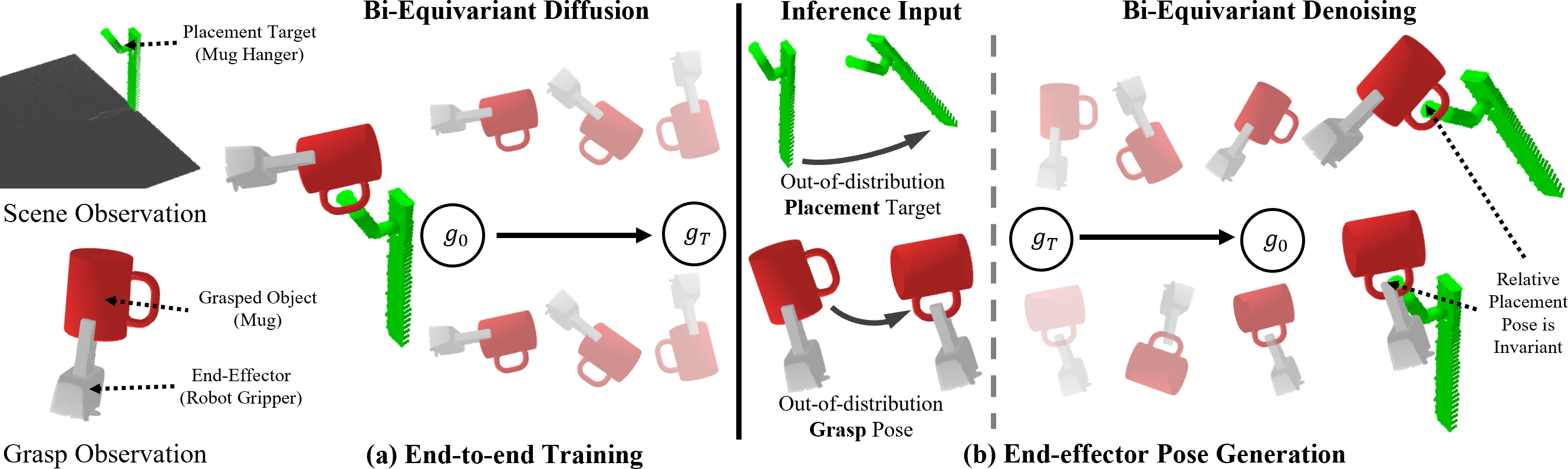}
  \caption{
    \textbf{Overview of Diffusion-EDFs.}
    \textbf{(a)} The target end-effector pose $g_0$ is bi-equivariantly diffused for the training of Diffusion-EDFs.
    \textbf{(b)} The end-effector pose is sampled from the policy by denoising with learned bi-equivariant score function. Due to the bi-equivariance, the trained policy can be effectively generalized to previously unseen configurations in the observation of the scene and the grasp.
  }
  \label{fig:Front}
\end{figure*}

\section{Preliminaries}
\subsection{SO(3) Group Representation Theory}
% A \emph{representation} $\D(g)$ is a map from a group $\mathcal{G}$ to a linear map on a vector space $\mathcal{W}$ that is homomorphic to the group action such that
A \emph{representation} $\D(g)$ is a map from a group $\mathcal{G}$ to a linear map on a vector space $\mathcal{W}$ that satisfies
\vspace{-0.2\baselineskip}
\begin{equation}
    \label{eqn:homomorphism}
    \D (g)\D (h)=\D (gh)\quad\forall g, h \in \mathcal{G}
    \vspace{-0.2\baselineskip}
\end{equation}
The vector space $\mathcal{W}$ where $\D(g)$ acts on is called the \emph{representation space} of $\D(g)$.
It is known that any representation of the special orthogonal group $SO(3)$ can be block-diagonalized into smaller representations by a change of basis.
\emph{Irreducible representations} are representations that cannot be reduced anymore, and hence constitute the building blocks of any larger representation.
% Following the convention of E3NN \citep{e3nn_paper}, we refer to vectors in irreducible representation spaces as \emph{irreducible vectors}.

According to the representation theory of $SO(3)$, all irreducible representations are classified according to their angular frequency $l \in \{0,1,2,\ldots\}$, a non-negative integer number called \emph{type}, or \emph{spin}.
Any type-$l$, or spin-$l$ representations are equivalent representations of the \emph{real Wigner D-matrix} of degree $l$, denoted as $\D_l(R):SO(3)\rightarrow\mathbb{R}^{(2l+1)\times(2l+1)}$.
We refer to the vectors in the representation space of $\D_l(R)$ as \emph{type-$l$}, or \emph{spin-$l$ vectors}.
Type-$0$ representations have zero angular frequency, i.e. $\D_0(R)=1$, meaning that type-$0$ vectors are \emph{scalars} that are invariant under rotations.
On the other hand, type-$1$ representations are identical when rotated by $360^{\circ}$, as their angular frequency is $1$.
Following the convention of E3NN \citep{e3nn_paper}, we use the $x$-$y$-$z$ basis in which $\D_{1}(R)=R$.
Therefore, type-$1$ vectors are typical spatial vectors in $\mathbb{R}^3$.
In general, $\D_l(R)$ is identical when rotated by $\theta=2\pi/l$, making higher-type vectors more suitable for encoding high-frequency details.
% In general, $\D_l(R)$ is identical when rotated by $\theta=2\pi/l$, making vectors of type-$2$ or higher to be better suited for encoding high frequency details in orientational features than type-$0$ scalars or type-$1$ vectors.

\subsection{Equivariant Descriptor Fields}
An Equivariant Descriptor Field (EDF)~\citep{ryu2023equivariant} $\boldsymbol{\varphi}(\vx|O)$ is 
% an $SE(3)$-equivariant\footnote{\label{footnote:edf}More precisely, it is $SO(3)$-equivariant and translation-invariant, forming a subset of $SE(3)$-equivariance.} 
an $SO(3)$-equivariant and translation-invariant 
vector field on $\mathbb{R}^3$ generated by a point cloud $\Obs\in\mathcal{O}$.
EDFs are decomposed into the direct sum of irreducible subspaces
\vspace{-0.3\baselineskip}
\begin{equation}
    \boldsymbol{\varphi}(\vx|\Obs)=\bigoplus_{n=1}^{N}\boldsymbol{\varphi}^{(n)}(\vx|\Obs)
    \vspace{-0.3\baselineskip}
\end{equation}
where $\boldsymbol{\varphi}^{(n)}(\vx|\Obs):\mathbb{R}^3\times\mathcal{O}\rightarrow\mathbb{R}^{2l_n+1}$ is 
% an $SE(3)$-equivariant\footref{footnote:edf} type-$l_n$ vector field generated by $O$.
a translation-invariant type-$l_n$ vector field generated by $\Obs$.
Therefore, an EDF $\boldsymbol{\varphi}(\vx|\Obs)$ is transformed according to $\Delta g=(\Delta \vp, \Delta R)\in SE(3),\  \Delta\vp\in\mathbb{R}^3,\  \Delta R\in SO(3)$ as
\begin{equation}
\label{eqn:edf_steer}
% \begin{split}
%     \boldsymbol{\varphi}(\Dg\,\vx|\Dg\cdot \Obs)&=\D (\Delta R)\boldsymbol{\varphi}(\vx|\Obs)\\
%     \forall\ \Delta g&=(\Delta \vp, \Delta R)\in SE(3)
% \end{split}
\boldsymbol{\varphi}(\Dg\,\vx|\Dg\cdot \Obs)=\D (\Delta R)\boldsymbol{\varphi}(\vx|\Obs)
\end{equation}
where $\D(R)$ is the block-diagonal matrix whose sub-matrices are Wigner D-matrices $\left\{\D_{l_n}(R)\right\}_{n=1}^{n=N}$.
% \citet{ryu2023equivariant} proposed to use EDFs to construct a bi-equivariant energy-based model for robotic object rearrangement tasks. 
% We provide more explanations on bi-equivariance in \Supp~\ref{appndx:bi-equivariance}.

\subsection{Brownian Diffusion on the SE(3) Manifold}
\label{sec:iso-diffusion}
Let $g_t\in SE(3)$ be generated by diffusing $g_0\in SE(3)$ for time $t$. The Brownian diffusion process is defined by the following Lie group stochastic differential equation (SDE)
\begin{equation}
    \label{eqn:iso_diff_sde}
    g_{t+dt}=g_t\exp{\left[dW\right]}
\end{equation}
% where $dW$ is the standard Wiener process on the Lie algebra $\mathfrak{se}(3)$, which is the 6-dimensional vector space that serves as the tangent space at the identity of $SE(3)$.
where $dW$ is the standard Wiener process on $\mathfrak{se}(3)$ Lie algebra.
The Brownian diffusion kernel $P_{t|0}(g_t|g_0) = \BrownianKernel_t(g_0^{-1}g_t)$ for the SDE in \eqref{eqn:iso_diff_sde} can be decomposed into rotational and translational parts \citep{yim2023se, corso2023diffdock} such that
\begin{align}
    &\BrownianKernel_t(g)=\ \mathcal{N}(\vp; \vmu=\boldsymbol{0}, \Sigma=tI)\ \mathcal{IG}_{SO(3)}(R; \eps=t/2) \label{eqn:iso_diff_kernel}
    \\
    % &\mathcal{IG}_{SO(3)}(R; \epsilon) = \sum_{l=0}^{\infty}(2l+1)e^{-\epsilon l (l+1)}\frac{\sin{((2l+1)\theta/2)}}{\sin{\theta/2}}\label{eqn:igso3}
    &\mathcal{IG}_{SO(3)}(R; \epsilon) = \sum_{l=0}^{\infty}(2l+1)e^{-\epsilon l (l+1)}\frac{\sin{(l\theta+\frac{\theta}{2})}}{\sin{\theta/2}}\label{eqn:igso3}
\end{align}
where $\mathcal{N}$ is the normal distribution on $\mathbb{R}^3$, $\mathcal{IG}_{SO(3)}$ is the isotropic Gaussian on $SO(3)$ \citep{nikolayev1970normal,savyolova1994normal,leach2022denoising,jagvaral2022diffusion}, $g=(\vp,R)\in SE(3),\  \vp\in\mathbb{R}^3,\  R\in SO(3)$, and $\theta$ is the rotation angle of $SO(3)$ in the axis-angle parameterization.
CDF sampling is used for the sampling of $\mathcal{IG}_{SO(3)}$ \citep{leach2022denoising}.
% $R$ can be sampled from $\mathcal{IG}_{SO(3)}$ with the following procedure \citep{leach2022denoising,ryu2023equivariant,jagvaral2022diffusion}. First, the rotation axis is sampled uniformly from the unit sphere $S^2$. Next, $\theta$ is sampled using the inverse-CDF method, ranging from $\theta=0$ to $\pi$. When calculating the cumulative distribution function (CDF), \eqref{eqn:igso3} must be multiplied by $(1-\cos{\theta})/\pi$. This is because the invariant volume element of $SO(3)$ is $dR=(1-\cos{\theta})/\pi\,d\theta dS^2$ in the axis-angle parameterization.

\subsection{Langevin Dynamics on the SE(3) Manifold}
\label{sec:AL-MCMC}
Let $\mathfrak{se}(3)$ be the Lie algebra that generates $SE(3)$.
A \emph{Lie derivative} $\Lie_{\mathcal{V}}$ along $\mathcal{V}\in\mathfrak{se}(3)$ of a differentiable function $f(g)$ on $SE(3)$ is defined as
\vspace{-0.3\baselineskip}
\begin{equation}
    \Lie_{\mathcal{V}}f(g)=\left.\frac{d}{d\epsilon}\right|_{\epsilon=0}f(g\exp\left[\epsilon\mathcal{V}\right])
    \vspace{-0.3\baselineskip}
\end{equation}
Let $dP(g)=P(g)dg$ be a distribution on $SE(3)$ with the invariant probability distribution function $P(g)$. 
The \emph{Langevin dynamics} for $dP(g)$ is defined as follows \citep{brockett1997notes, chirik}:
\begin{align}
    g_{\tau+d\tau}&=g_{\tau}\exp{\left[ \frac{1}{2}\Grad \log{P(g)}d\tau+dW\right]}\\[-5pt]
    \Grad \log P(g)&=\sum_{i=1}^{6}\Lie_i \log P(g)\,\basis_i
\end{align}
where in the last line we denote the Lie derivative along the $i$-th basis $\basis_i\in\mathfrak{se}(3)$ as $\Lie_{i}$ instead of $\Lie_{\basis_i}$ for brevity.
We denote the time for the Langevin dynamics as $\tau$, as we reserve the notation $t$ for the diffusion time. 
It is known that under mild assumptions, this process converges to $dP(g)$ as $\tau\rightarrow\infty$ regardless of the initial distribution. 
Thus, one may sample from $dP(g)$ with Langevin dynamics if the \emph{score function} $\vs(g)=\Grad \log P(g):SE(3)\rightarrow\mathfrak{se}(3)$ is known.

\section{Bi-equivariant Score Matching on the SE(3) Manifold}
\subsection{Problem Formulation}
Let the target policy distribution\footnote{For notational simplicity, we do not distinguish the probability distribution $dP=Pdg$ from the probability distribution function (PDF) $P$ where $dg$ denotes the bi-invariant volume form \citep{chirik,murray2017mathematical,zee2016group} on $SE(3)$.} be $P_{0}(g_0|\Oscene,\Ograsp)$, where $g_0\in SE(3)$ is the target end-effector pose, and $\Oscene$ and $\Ograsp$ are the observed point clouds of the scene and the grasped object, respectively. 
Note that $\Oscene$ is observed in the scene frame $\sceneFrame$, and $\Ograsp$ in the end-effector frame $\eefFrame$.
Following \citet{ryu2023equivariant}, we model $P_0$ to be bi-equivariant (see \Supp~\ref{appndx:bi-equivariance}):
\vspace{-0.3\baselineskip}
\begin{equation}
    \label{eqn:bi-equiv}
    \begin{split}
        P_{0}(g|\Oscene,\Ograsp)&=P_{0}(\Dg\,g|\Dg\cdot \Oscene,\Ograsp)\\
        &=P_{0}(g\Dg^{-1}|\Oscene,\Dg\cdot \Ograsp)
    \end{split}
    \vspace{-0.3\baselineskip}
\end{equation}

Now let $g_t\in SE(3)$ be the samples that are noised from $g_0$ by some diffusion process, where $t$ denotes the diffusion time. 
A detailed explanation of this diffusion process will be deferred to a  subsequent section.
Our goal is to train a model that denoises $g_t$, which is sampled from the diffused marginal distribution $P_{t}(g_t|\Oscene,\Ograsp)$, into a denoised sample $g$, which follows the target distribution $P_{0}(g|\Oscene,\Ograsp)$. This can be achieved with Annealed Langevin MCMC \citep{song2019generative,yim2023se,corso2023diffdock,bortoli2022riemannian,huang2022riemannian,jagvaral2022diffusion} if the \emph{score function} (see \Sec~\ref{sec:AL-MCMC}) of $P_t$ is known. See \Fig~\ref{fig:Front} for the overview of Diffusion-EDFs.

%%%%%%%%%%%%%%%%%%%%%%%%%%%%%%%%%%%%%%%%%%%%%%%%%%%%%%%%%%%%%%%%%%%%%%%%%%%%%%%%%%%%%%%%%%%%%%%%%%%%
% Section: Bi-equivariant Score Function
%%%%%%%%%%%%%%%%%%%%%%%%%%%%%%%%%%%%%%%%%%%%%%%%%%%%%%%%%%%%%%%%%%%%%%%%%%%%%%%%%%%%%%%%%%%%%%%%%%%%
\subsection{Bi-equivariant Score Function}
Let $\vs(g|\Oscene,\Ograsp) = \Grad \log{P(g|\Oscene,\Ograsp)}$ be the score function of a probability distribution $P(g|\Oscene,\Ograsp)$.
% \begin{equation}
%     \begin{split}
%         \vs(g|\Oscene,\Ograsp) &= \Grad \log{P(g|\Oscene,\Ograsp)} \\
%         &= \sum_{i=1}^{6}\Lie_i \log{P(g|\Oscene,\Ograsp)}\,\basis_i
%     \end{split}
% \end{equation}
% \eqref{eqn:bi-equiv} induces specific equivariance conditions for the score function that can be exploited in our model.
\begin{propo}
    \label{propo:score_left_and_right}
    $\vs(g|\Oscene,\Ograsp)$ satisfies the following conditions for all $\Dg\in SE(3)$ if $P(g|\Oscene,\Ograsp)$ is bi-equivariant:
    \begin{align}
        \vs(\Dg\,g|\Dg\cdot \Oscene,\Ograsp) &= \vs(g|\Oscene,\Ograsp)\label{eqn:left-inv}\\ 
        \vs(g\,\Dg^{-1}|\Oscene,\Dg\cdot \Ograsp) &= \left[\mathrm{Ad}_{\Dg}\right]^{-T} \vs(g|\Oscene,\Ograsp) \label{eqn:right-equiv}
    \end{align}
\end{propo}
\noindent $\mathrm{Ad}_{g}$ is the \textit{adjoint representation} \citep{chirik,murray2017mathematical,lynch2017modern} of $SE(3)$ with $g=(\vp, R)$, $\vp\in \mathbb{R}^3$, and $R\in SO(3)$
\begin{equation}
    \label{eqn:adjoint_mat}
    Ad_g=
        \begin{bmatrix}
        R & [\vp]^{\wedge}R\\
        \emptyset & R
        \end{bmatrix}
        % =
        % \begin{bmatrix}
        % R & \emptyset\\
        % R & R
        % \end{bmatrix}
\end{equation}
where $[\vp]^{\wedge}$ denotes the skew-symmetric $3\times 3$ matrix of $\vp$. 
% We follow the notation from \citet{murray2017mathematical} and place the translational part in the upper rows.
See \Supp~\ref{proof:score_left_and_right} for the proof of \Proposition~\ref{propo:score_left_and_right}. 
% We refer to \eqref{eqn:left-inv} as the \emph{left invariance} of the score, since $\Dg$ comes to the left side of $g$, and the score remains invariant.
% Likewise, we refer to \eqref{eqn:right-equiv} as the \emph{right equivariance} of the score, as the inverse of $\Dg$ comes to the right side of $g$, and the score is adjoint-transformed accordingly.

%%%%%%%%%%%%%%%%%%%%%%%%%%%%%%%%%%%%%%%%%%%%%%%%%%%%%%%%%%%%%%%%%%%%%%%%%%%%%%%%%%%%%%%%%%%%%%%%%%%%
% Section: Bi-equivariant Diffusion Process
%%%%%%%%%%%%%%%%%%%%%%%%%%%%%%%%%%%%%%%%%%%%%%%%%%%%%%%%%%%%%%%%%%%%%%%%%%%%%%%%%%%%%%%%%%%%%%%%%%%%
\subsection{Bi-equivariant Diffusion Process}
\label{sec:bi-equiv-diffusion}
% One may consider learning the score function of the diffused marginal $P_{t}(g_t|\Oscene,\Ograsp)$ for the denoising generative modeling. 
Let the point cloud conditioned diffusion kernel under time~$t$ be $P_{t|0}(g|g_0,\Oscene,\Ograsp)$ such that the diffused marginal $P_t(g|\Oscene,\Ograsp)$ for $P_0(g|\Oscene,\Ograsp)$ is defined as follows:
\vspace{-0.3\baselineskip}
\begin{equation}
    \label{eqn:marginal}
    % \begin{split}
    %     &P_t(g|\Oscene,\Ograsp)\\
    %     &=\intOverSEthree dg_0\;P_{t|0}(g|g_0,\Oscene,\Ograsp)P_0(g_0|\Oscene,\Ograsp)
    % \end{split}
    % P_t(g|\Oscene,\Ograsp)
    % =\intOverSEthree dg_0\;P_{t|0}(g|g_0,\Oscene,\Ograsp)P_0(g_0|\Oscene,\Ograsp)
    \begin{split}
        P_t(g|\Oscene,\Ograsp)
        =\intOverSEthree dg_0\;P_{t|0}(g|g_0,\Oscene,\Ograsp)P_0(g_0|\Oscene,\Ograsp) 
    \end{split}  \raisetag{10pt}
\end{equation} 
If the diffused marginal $P_t(g|\Oscene,\Ograsp)$ is bi-equivariant, one may leverage \Proposition~\ref{propo:score_left_and_right} in the score model design.

\begin{defi}
     % A bi-equivariant diffusion kernel $P_{t|0}(g|g_0,\ \Oscene,\Ograsp)$ is a square-integrable kernel that satisfies the following equations for all $\Dg\in SE(3)$, except on a set of measure zero:
     A bi-equivariant diffusion kernel $P_{t|0}$ is a square-integrable kernel that satisfies the following equations for all $\Dg\in SE(3)$, except on a set of measure zero:
    \begin{equation}
        \label{eqn:bi-equiv-kernel}
        \begin{split}
            &P_{t|0}(g|g_0,\ \Oscene,\Ograsp)=P_{t|0}(\Dg\,g|\Dg\,g_0,\Dg\cdot \Oscene,\Ograsp) \\
            &\phantom{111111111\,1}=P_{t|0}(g\,\Dg^{-1}|g_0\,\Dg^{-1},\Oscene,\Dg\cdot \Ograsp)
        \end{split}
    \end{equation}
\end{defi}
\begin{propo}
    \label{propo:marginal_biequiv}
    The diffused marginal $P_t$ is guaranteed to be bi-equivariant for all bi-equivariant initial distribution $P_0$ if and only if the diffusion kernel $P_{t|0}$ is bi-equivariant.
\end{propo}
\noindent See \Supp~\ref{proof:marginal_biequiv} for the proof of \Proposition~\ref{propo:marginal_biequiv}. 
Note that the Brownian diffusion kernel $P_{t|0}(g|g_0)=\BrownianKernel_t(g_0^{-1}g)$ in \eqref{eqn:iso_diff_kernel} is left invariant\footref{footnote:invariance} but not right invariant\footnote{
    \label{footnote:invariance}
    We use the term \emph{invariance} instead of equivariance since the kernel is neither conditioned by $\Oscene$ nor $\Ograsp$.
}, that is
% \begin{equation}
%     \begin{split}
%        P_{t|0}(\Dg \, g |\Dg \, g_0)&= P_{t|0}(g|g_0)\ \forall\,\Dg\in SE(3)\\
%        P_{t|0}(g\,\Dg^{-1}|g_0\,\Dg^{-1})&\neq P_{t|0}(g|g_0)\ \forall\,\Dg\in SE(3)
%     \end{split}
% \end{equation}
\begin{equation}
    \begin{split}
       &\forall\,\Dg\in SE(3),\ P_{t|0}(\Dg \, g |\Dg \, g_0)= P_{t|0}(g|g_0)\\
       &\exists\,\Dg\in SE(3),\ P_{t|0}(g\,\Dg^{-1}|g_0\,\Dg^{-1})\neq P_{t|0}(g|g_0) \phantom{1 1 1 1}
    \end{split} \raisetag{1.6\baselineskip}
\end{equation}
In fact, there exist no square-integrable kernel on $SE(3)$ that is bi-invariant\footref{footnote:invariance} (see \Supp~\ref{proof:non-existence-of-bi-invariant-kernel}). 
% Therefore, we propose to let   the diffusion kernel   be at least dependent on either $\Oscene$ or $\Ograsp$ to absorb the left or right group action of $\Dg$.
Therefore, a bi-equivariant diffusion kernel must be dependent on either $\Oscene$ or $\Ograsp$ to absorb the left or right action of $\Dg$.

% For generality, we keep both $\Oscene$ and $\Ograsp$ in the following. 

% While there may be numerous ways to implement such bi-equivariant diffusion kernels, we use a diffusion kernel with an equivariant \emph{diffusion frame selection mechanism} $\FrameSelection(\gref|g_0^{-1}\cdot\Oscene, \Ograsp)$ such that
To implement such bi-equivariant diffusion kernels, we use an equivariant \emph{diffusion frame selection mechanism} $\FrameSelection(\gref|g_0^{-1}\cdot\Oscene, \Ograsp)$ where $\gref\in SE(3)$ is the pose of the diffusion frame $\refFrame$ with respect to the end-effector frame $\eefFrame$
\vspace{-0.3\baselineskip}
\begin{equation}
    \label{eqn:frame_selection_marginal}
    \begin{split}
        &P_{t|0}(g|g_0,\Oscene,\Ograsp)\\
        &=\intOverSEthree d\gref\FrameSelection(\gref|g_0^{-1}\cdot\Oscene,\Ograsp)\LinvKernel_{t}(\gref^{-1}g_0^{-1}g\gref)
    \end{split}
\end{equation}
where $\LinvKernel_{t}(g_0^{-1}g)$ is any left invariant kernel (see \Supp~\ref{proof:non-existence-of-bi-invariant-kernel}).
% \footnote{Note that any left invariant kernel $\LinvKernel_{t}(g,g_0)$ can be written as $\LinvKernel_{t}(g_0^{-1}g)$ (see \Supp~\ref{proof:non-existence-of-bi-invariant-kernel}).}.
The diffusion procedure is as follows:
\begin{itemize}
    \item[D1.] A target pose $g_0\sim P_0(g_0|\Oscene,\Ograsp)$ is sampled. %from the demonstration policy.
    \item[D2.] A diffusion frame $\gref\sim\FrameSelection(\gref|g_0^{-1}\cdot\Oscene, \Ograsp)$ is sampled.% from the diffusion frame selection mechanism.
    \item[D3.] A diffusion displacement $\Dg_{t|0}\sim \LinvKernel_{t}(\Dg_{t|0})$ is sampled. %from $\LinvKernel_t$ in \eqref{eqn:frame_selection_marginal}.
    \item[D4.] $\Dg_{t|0}$ is applied to the demonstrated end-effector pose $g_0$ in the diffusion frame $\refFrame$, that is, $g_{t} = g_{0}\,\gref\,\Dg_{t|0}\,\gref^{-1}$ where $g_t\sim P_t$ is the diffused end-effector pose.
\end{itemize}
\begin{propo}
    \label{propo:bi-equiv-of-diffusion-marginal-kernel}
    The diffusion kernel $P_{t|0}$ in \eqref{eqn:frame_selection_marginal} is bi-equivariant if the diffusion frame selection mechanism $\FrameSelection(\gref|g_0^{-1}\cdot\Oscene, \Ograsp)$ satisfies the following property:
    \begin{equation}
        \label{eqn:frame_selection}
        % \begin{split}
        %     &\FrameSelection(\gref|g_0^{-1}\cdot\Oscene, \Ograsp)\\
        %     &=\FrameSelection(\Dg\,\gref|(\Dg\,g_0^{-1})\cdot\Oscene, \Dg\cdot \Ograsp)
        % \end{split}
        \begin{split}
            \FrameSelection(\gref|g_0^{-1}\cdot\Oscene, \Ograsp)
            =\FrameSelection(\Dg\,\gref|(\Dg\,g_0^{-1})\cdot\Oscene, \Dg\cdot \Ograsp)
        \end{split}
    \end{equation}
    % where $\gref\in SE(3)$ is the frame of the diffusion relative to $g_0$. 
\end{propo}
See \Supp~\ref{proof:frame_selection_marginal} for the proof. 
In practice, however, the orientational part of the frame selection mechanism may be difficult to implement.
% Remarkably, only the translation part of the frame selection mechanism is required for the bi-equivariance of \eqref{eqn:frame_selection_marginal} if $\LinvKernel_{t}$ is the Brownian diffusion kernel $\BrownianKernel_t$. 
Remarkably, for the specific case in which $\LinvKernel_{t}$ is the Brownian diffusion kernel $\BrownianKernel_t$,
only the translation part of the frame selection is required for \eqref{eqn:frame_selection_marginal} to be bi-equivariant.
% For example, consider the following frame selection mechanism
Therefore, we modify our diffusion frame selection mechanism as follows:
% \vspace{-0.3\baselineskip}
\begin{equation}
    \label{eqn:only-translation}
    % \begin{split}
    %     &\FrameSelection(\gref|g_0^{-1}\cdot\Oscene, \Ograsp)\\
    %     &=\OriginSelection(\pref|g_0^{-1}\cdot\Oscene, \Ograsp)\,\delta_{SO(3)}(R_{ref})
    % \end{split}
    \FrameSelection(\gref|g_0^{-1}\cdot\Oscene, \Ograsp)
    =\OriginSelection(\pref|g_0^{-1}\cdot\Oscene, \Ograsp)\,\delta(\Rref)
    % \vspace{-0.3\baselineskip}
\end{equation}
where $\delta(R)$ is the Dirac delta on $SO(3)$ 
% such that the orientation part of $\gref$ is always same as the identity, 
and $\OriginSelection(\pref|g_0^{-1}\cdot\Oscene, \Ograsp)$ is the \emph{diffusion origin selection mechanism}.
\begin{propo}
    \label{propo:only-translation}
    The diffusion kernel $P_{t|0}$ in \eqref{eqn:frame_selection_marginal} with the frame selection mechanism in \eqref{eqn:only-translation} is bi-equivariant if $\LinvKernel_t$ in \eqref{eqn:frame_selection_marginal} is the Brownian diffusion kernel and the origin selection mechanism in \eqref{eqn:only-translation} is equivariant that
    % \vspace{-0.3\baselineskip}
    \begin{equation}
        \label{eqn:origin_selection}
        \begin{split}
        &\OriginSelection\left(
            \pref|g_0^{-1}\cdot \Oscene,\Ograsp
        \right) \\
        &=\OriginSelection\left(
            \Dg\,\pref|(\Dg\,g_0^{-1})\cdot \Oscene, \Dg\cdot \Ograsp
        \right)
        \end{split}
        % \OriginSelection\left(
        %     \pref|g_0^{-1}\cdot \Oscene,\Ograsp
        % \right) 
        % =\OriginSelection\left(
        %     \Dg\,\pref|(\Dg\,g_0^{-1})\cdot \Oscene, \Dg\cdot \Ograsp
        % \right)
    \end{equation}
\end{propo} %\vspace{-0.3\baselineskip}
\noindent We provide the proof in \Supp~\ref{proof:only-translation}.
A concrete realization of such equivariant diffusion origin selection mechanism $\OriginSelection(\pref|g_0^{-1}\cdot\Oscene, \Ograsp)$ is discussed in \Sec~\ref{sec:contact-heuristic}.

%%%%%%%%%%%%%%%%%%%%%%%%%%%%%%%%%%%%%%%%%%%%%%%%%%%%%%%%%%%%%%%%%%%%%%%%%%%%%%%%%%%%%%%%%%%%%%%%%%%%
% Section: Score Matching Objectives
%%%%%%%%%%%%%%%%%%%%%%%%%%%%%%%%%%%%%%%%%%%%%%%%%%%%%%%%%%%%%%%%%%%%%%%%%%%%%%%%%%%%%%%%%%%%%%%%%%%%
\subsection{Score Matching Objectives}
In contrast to \citet{song2019generative,urain2022se3dif}, our diffusion kernel $P_{t|0}(g|g_0,\Oscene, \Ograsp)$ in \eqref{eqn:frame_selection_marginal} is not the Brownian kernel. Still, the following mean squared error (MSE) loss can be used to train our score model $\vs_t(g|\Oscene, \Ograsp)$ without requiring the integration of \eqref{eqn:frame_selection_marginal}:
% to approximate the score function of the marginal $P_t$:
% we require the Lie-derivative of the log-likelihood of the diffusion kernel $P_{t|0}(g|g_0,\Oscene, \Ograsp)$ \citep{song2019generative,urain2022se3dif}. However, \eqref{eqn:frame_selection_marginal} may not allow tractable calculation depending on the frame selection mechanism $\FrameSelection$. Still, the score model can be efficiently trained by minimizing the following objective
\begin{equation}
    \label{eqn:score_matching_loss}
    \begin{split}
        \mathcal{J}_t&=\mathbb{E}_{g, g_0, \gref, \Oscene,\Ograsp}\left[
            J_t
        \right] \\
        J_t &= \frac{1}{2}  \left \|\vs_t(g|\Oscene,\Ograsp)-\Grad \log \LinvKernel_{t}(\gref^{-1}g_0^{-1}g\gref)  \right \|^2
    \end{split}
\end{equation}
where $g_0\sim P_0(g_0|\Oscene,\Ograsp)$, $\ \gref\sim \FrameSelection(\gref|g_0^{-1}\cdot \Oscene,\Ograsp)$, and $\ g\sim P_{t|0}(g|g_0,\Oscene,\Ograsp)$.
We optimize $\mathcal{J}_t$ for sampled reference frame $\gref$ and diffusion time $t$.
The minimizer of $\mathcal{J}_t$ is neither $\Grad\log\LinvKernel_{t}$ nor $\Grad\log P_{t|0}$ but the score function of the diffused marginal $\Grad\log P_t$, that is
\begin{equation}
    \label{eqn:mse-minimizer}
    % \begin{split}
    %     \argmin_{\vs_t(g|\Oscene,\Ograsp)} \mathcal{J}_t &= \vs_t^*(g|\Oscene,\Ograsp)\\[-8pt]
    %     &=\Grad \log{P_t({g|\Oscene,\Ograsp})}
    % \end{split}
    \begin{split}
        \argmin_{\vs_t(g|\Oscene,\Ograsp)} \mathcal{J}_t = \vs_t^*(g|\Oscene,\Ograsp)
        =\Grad \log{P_t({g|\Oscene,\Ograsp})}
    \end{split}
\end{equation}
Although \eqref{eqn:mse-minimizer} is a straightforward adaptation of the  MSE minimizer formula \citep{song2019generative},  we still provide the derivation in \Supp~\ref{proof:mse-minimizer} for completeness. 
In practice, we use the Brownian diffusion kernel $\BrownianKernel_t$ for $\LinvKernel_t$ to exploit \Proposition~\ref{propo:only-translation}.
Therefore, training with \eqref{eqn:score_matching_loss} requires the computation of $\Grad \log \BrownianKernel_{t}(\gref^{-1}g_0^{-1}g\gref)$. 
While autograd packages can be used for this computation \citep{leach2022denoising,ryu2023equivariant,jagvaral2022diffusion,urain2022se3dif,corso2023diffdock,yim2023se}, we use a more stable explicit form in \Supp~\ref{appndx:analytic-target-score}.

%%%%%%%%%%%%%%%%%%%%%%%%%%%%%%%%%%%%%%%%%%%%%%%%%%%%%%%%%%%%%%%%%%%%%%%%%%%%%%%%%%%%%%%%%%%%%%%%%%%%
\begin{figure*}[t]
  \centering
  %\framebox{\parbox{\textwidth}{\includegraphics[width=\textwidth]{figure11_solid.png}}}
  \includegraphics[width=0.9\textwidth]{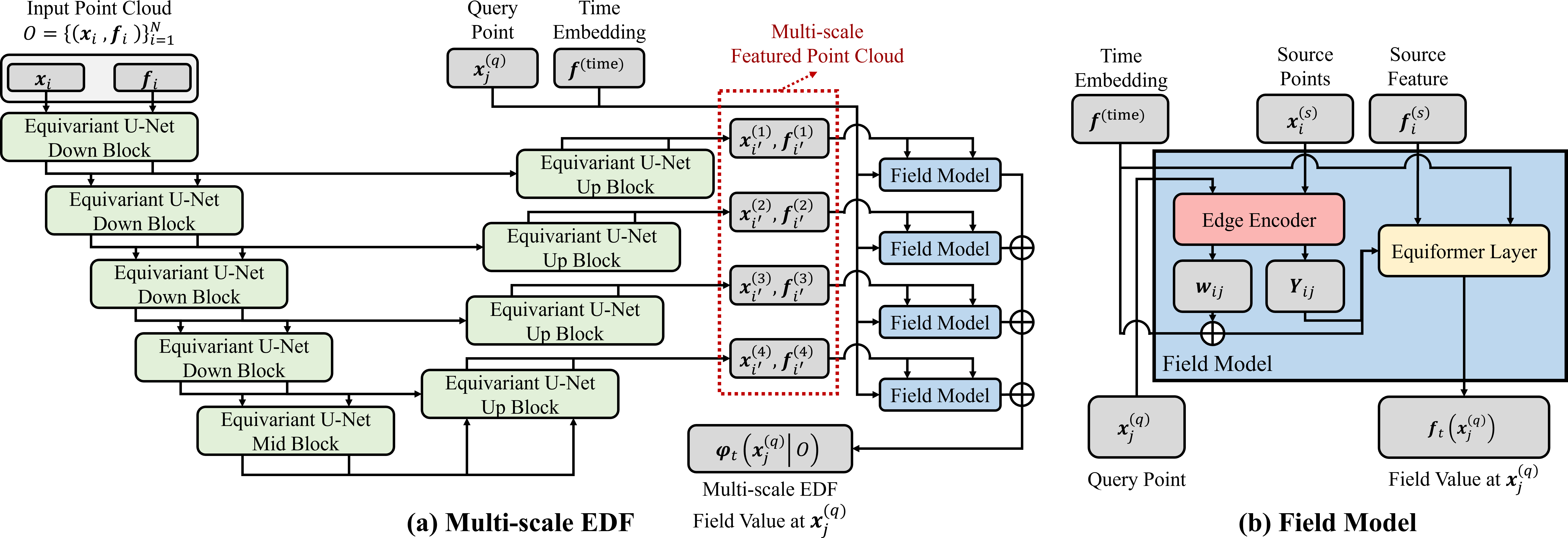}
  \caption{
    \textbf{Architecture of multiscale EDF.} 
    Our multiscale EDF model is composed of a feature extracting part and a field model part.
    See \Fig~\ref{fig:modules} in \Supp~\ref{appndx:architecture} for details on each module in the architecture.
    \textbf{(a)} The feature extractor encodes the input point cloud into multiscale featured point clouds.
    We use an U-Net-like GNN architecture for the feature extractor part.
    \textbf{(b)} The encoded multiscale point clouds are passed into the field model part along with the query point and time embedding. The field model outputs the time-conditioned EDF field value at the query point. We simply sum up the output from each scale to obtain the EDF field value at the query point.
    % See \cref{fig:modules} for specific design of each blocks.
  }
  \label{fig:Unet}
\end{figure*}
%%%%%%%%%%%%%%%%%%%%%%%%%%%%%%%%%%%%%%%%%%%%%%%%%%%%%%%%%%%%%%%%%%%%%%%%%%%%%%%%%%%%%%%%%%%%%%%%%%%%

%%%%%%%%%%%%%%%%%%%%%%%%%%%%%%%%%%%%%%%%%%%%%%%%%%%%%%%%%%%%%%%%%%%%%%%%%%%%%%%%%%%%%%%%%%%%%%%%%%%%
% Section: Bi-equivariant Score Model
%%%%%%%%%%%%%%%%%%%%%%%%%%%%%%%%%%%%%%%%%%%%%%%%%%%%%%%%%%%%%%%%%%%%%%%%%%%%%%%%%%%%%%%%%%%%%%%%%%%%
\subsection{Bi-equivariant Score Model}
\label{sec:bi-equiv-score-model}
% To model the left invariance and the right equivariance of the score function as in \eqref{eqn:left-inv} and \eqref{eqn:right-equiv}, we propose the following score model on $SE(3)$:
% We propose the following score model that satisfies the equivariance conditions in \eqref{eqn:left-inv} and \eqref{eqn:right-equiv}:
We split our score model $\vs_t(\cdot|\Oscene,\Ograsp):SE(3)\rightarrow\mathfrak{se}(3)\cong \mathbb{R}^6$ into the direct sum of translational and rotational parts
\begin{equation}
    \label{eqn:score_model}
    \vs_t(g|\Oscene,\Ograsp) = \left[\vs_{\nu;t} \oplus \vs_{\omega;t}\right](g|\Oscene,\Ograsp) 
\end{equation}
% \newpage
\noindent where we denote the translational part with subscript $\nu$ and rotational part with subscript $\omega$.
Thus, $\vs_{\nu;t}(\cdot|\Oscene,\Ograsp):SE(3)\rightarrow \mathbb{R}^3$ is the translational score and $\vs_{\omega;t}(\cdot|\Oscene,\Ograsp):SE(3)\rightarrow \mathfrak{so(3)}\cong \mathbb{R}^3$ is the rotational score.
To satisfy the equivariance conditions in \eqref{eqn:left-inv} and \eqref{eqn:right-equiv}, we propose the following models:
\begin{align}
    % \label{eqn:score_model}
    % \begin{split}
    %     &\vs_t(g|\Oscene,\Ograsp) \\
    %     &= \vs_{\nu;t}(g|\Oscene,\Ograsp) \oplus \vs_{\omega;t}(g|\Oscene,\Ograsp) 
    % \end{split} \\
    %%%%%%%%%%%%%%%%%%%%%%%
    % \hspace{-5pt}
    % \label{eqn:score_model}
    % &\vs_t(g|\Oscene,\Ograsp) = \left[\vs_{\nu;t} \oplus \vs_{\omega;t}\right](g|\Oscene,\Ograsp) 
    % \\ 
    %%%%%%%%%%%%%%%%%%%%%%%
    \label{eqn:nu_model}
    % \hspace{-10pt}
    \begin{split}
        &\vs_{\nu;t}(g|\Oscene,\Ograsp) 
        \hspace{-1pt}
        = \intOverRthree d^3\vx\ \rho_{\nu;t}(\vx|\Ograsp)\ \widetilde{\vs}_{\nu;t}(g,\vx|\Oscene,\Ograsp) \phantom{11111}
    \end{split} \raisetag{40pt}
    % \hspace{-20pt}
    \\
    %%%%%%%%%%%%%%%%%%%%%%%
    \label{eqn:omg_model}
    \begin{split}
        &\vs_{\omega;t}(g|\Oscene,\Ograsp) 
        \hspace{-1pt}
        =\uwave{\intOverRthree d^3\vx\ \rho_{\omega;t}(\vx|\Ograsp)\ \widetilde{\vs}_{\omega;t}(g,\vx|\Oscene,\Ograsp)}_{\raisebox{0pt}{\hspace{-31pt} $_\text{Spin term}$}}
        % \hspace{-5pt}
        \\[-5pt]
        &\phantom{\qquad\qquad\ }
        +\uwave{\intOverRthree d^3\vx\ \rho_{\nu;t}(\vx|\Ograsp)\ \vx\cross\widetilde{\vs}_{\nu;t}(g,\vx|\Oscene,\Ograsp)}_{\raisebox{0pt}{\hspace{-39pt} $_\text{Orbital term}$}}
        % \hspace{-20pt}
    \end{split} \raisetag{40pt}
\end{align}
where $\cross$ denotes the cross product (wedge product).
In these models, we compute the translational and rotational score using two different types of equivariant fields: 1) the equivariant density field $\rho_{_\square;t}(\cdot|\Ograsp):\mathbb{R}^3\rightarrow\mathbb{R}_{\geq 0}$, and 2) the time-conditioned score field $\widetilde{\vs}_{_\square;t}(\cdot|\Oscene,\Ograsp):SE(3)\times \mathbb{R}^3\rightarrow\mathbb{R}^3$, where $\square$ is either $\omega$ or $\nu$.

\begin{propo}
    \label{propo:score-model}
    The score model in \eqref{eqn:score_model} satisfies \eqref{eqn:left-inv} and \eqref{eqn:right-equiv} if for $\square=\omega,\nu$ the density and  score fields satisfy the following conditions for all $\Dg \in SE(3)$
    % \begin{align}
    %     \rho_{_\square}(\Dg\,g|\Dg\cdot \Ograsp)&= \rho_{_\square}(g|\Ograsp)&\forall\ \Dg \in SE(3) \tag{Invariance of density}\\
    %     \vs_{_\square;t}(\Dg\,g, \Dg\,\vx|\Dg \cdot \Oscene,\Ograsp)&= \vs_{_\square;t}(g,x|\Oscene,\Ograsp)&\forall\ \Dg \in SE(3)  \tag{Left invarance of score}\\
    %     \vs_{_\square;t}(g\,{\Dg}^{-1}, \Dg\,\vx|\Oscene,\Dg \cdot \Ograsp)&= R\,\vs_{_\square;t}(g,x|\Oscene,\Ograsp)&\forall\ \Dg \in SE(3) \tag{Right equivariane of score}
    % \end{align}
    \begin{align}
        \label{eqn:density_equiv}
        &\rho_{_\square;t}(\Dg\,\vx|\Dg\cdot \Ograsp)= \rho_{_\square;t}(\vx|\Ograsp)\\
        \label{eqn:score_field_left_equiv}
        % &\widetilde{\vs}_{_\square;t}(\Dg\,g, \vx|\Dg \cdot \Oscene,\Ograsp)= \widetilde{\vs}_{_\square;t}(g,\vx|\Oscene,\Ograsp)\\
        \begin{split}
            &\widetilde{\vs}_{_\square;t}(\Dg\,g, \vx|\Dg \cdot \Oscene,\Ograsp) 
            = \widetilde{\vs}_{_\square;t}(g,\vx|\Oscene,\Ograsp)
        \end{split} \\
        \label{eqn:score_field_right_equiv}
        \begin{split}
            &\widetilde{\vs}_{_\square;t}(g\,{\Dg}^{-1}, \Dg\,\vx|\Oscene,\Dg \cdot \Ograsp) 
            = \Delta R\,\widetilde{\vs}_{_\square;t}(g,\vx|\Oscene,\Ograsp )
        \end{split}
    \end{align}
\end{propo}
\noindent See \Supp~\ref{proof:score-model} for the proof. To achieve the left invariance (\eqref{eqn:score_field_left_equiv}) and right equivariance (\eqref{eqn:score_field_right_equiv}) of the score field, we propose using the following model with two EDFs:
\begin{equation}
\label{eqn:score_field_model}
    \begin{split}
        &\widetilde{\vs}_{\square;t}(g,\vx|\Oscene,\Ograsp)\\
        &=\boldsymbol{\psi}_{\square;t}(\vx|\Ograsp)\ \otimes_{\square;t}^{(\rightarrow 1)}\ \D (R^{-1})\,\boldsymbol{\varphi}_{\square;t}(g\,\vx|\Oscene)
    \end{split}
\end{equation}
where $\boldsymbol{\varphi}_{\square;t}$ and $\boldsymbol{\psi}_{\square;t}$ are two different EDFs that respectively encode the point clouds $\Oscene$ and $\Ograsp$, and $\otimes_{\square;t}^{(\rightarrow 1)}$ is the time-conditioned equivariant tensor product~\citep{thomas2018tensor,fuchs2020se} with \emph{Clebsch-Gordan coefficients} that maps the highly over-parametrized equivariant descriptors into a type-$1$ vector.
%
% \\ \phantom{This Line is required to prevent Latex bug} \vspace{-\baselineskip}
\begin{propo}
    \label{propo:edf-score}
    The score field model
    in \eqref{eqn:score_field_model} satisfies \eqref{eqn:score_field_left_equiv} and \eqref{eqn:score_field_right_equiv}.
\end{propo}
\noindent We provide the proof of \Proposition~\ref{propo:edf-score} in \Supp~\ref{proof:score-field-biequiv}. 
% Note that the resulting score field model $\widetilde{\vs}_{\square;t}(g,\vx|\Oscene,\Ograsp)$ can be considered as the type-$1$ generalization to the energy function of \citet{ryu2023equivariant}, which is a type-$0$ (rotation invariant scalar) function. 

\section{Implementation}
In this section, we first provide the specific implementation of the bi-equivariant diffusion frame selection mechanism, which was postponed in \Sec~\ref{sec:bi-equiv-diffusion}.
% We then propose novel multiscale EDF architectures in \Sec~\ref{sec:bi-equiv-score-model}.
We then provide a novel multiscale EDF architecture, and the query points model.
% Especially, we focus on designing our method to respect the highly local nature of manipulation tasks. 
% % Many objects can be decomposed into local sub-geometries, and for the most of the cases, specific sub-geometries are much more significant than others in terms of deciding the target pose. 
% % For example, when hanging a mug on a rack by the handle (see Fig. TODO), the target placement pose should only be dependent on the local geometry of handle, not the whole mug.
% % Therefore, manipulation methods should be able to take into account such locality. 
% % Such importance of incorporating locality in equivariant methods has recently been reported in various works \citep{chun2023local,chatzipantazis2022se} in the name of \emph{local equivariance} \citep{ryu2023equivariant,kim2023robotic} or \emph{part equivariance} \citep{deng2023banana}. 
% % While recently there have been some attempts to formally define this concepts, we confine our discussion to rather empirical property.
% Such importance of incorporating locality in equivariant methods has recently been reported in various works \citep{chun2023local,ryu2023equivariant,kim2023robotic,chatzipantazis2022se,deng2023banana}. 
Further details such as non-dimensionalization and denoising schedule are provided in \Supp~\ref{appndx:imple_detail}

\subsection{Diffusion Origin Selection Mechanism}
\label{sec:contact-heuristic}
For most manipulation tasks, specific local sub-geometries are more significant than the global geometry of the target object in determining its pose. 
Several works have addressed the importance of incorporating such locality in equivariant methods \citep{chun2023local,ryu2023equivariant,kim2023robotic,chatzipantazis2022se,deng2023banana}.
In manipulation tasks, contact-rich sub-geometries are more likely to be important than the others.
We exploit this property by selecting the origin of diffusion near contact-rich sub-geometries. 

Let $n_r(\vx,\Obs)$ be the number of points in a point cloud $\Obs$ that is within a contact radius $r$ from a point $\vx\in\mathbb{R}^3$. 
We use the following diffusion origin selection mechanism with $r$ as a hyperparameter.
\begin{equation}
    \label{eqn:neighbor_origin_selection}
    %
    % \begin{split}
    %     &\OriginSelection\left(
    %         \vp_{ref}|g_0^{-1}\cdot \Oscene,\Ograsp
    %     \right)\\
    %     &\propto 
    %     \begin{cases}
    %         n_r\left(
    %             \vp_{ref},\ g_0^{-1}\cdot \Oscene
    %         \right) &  \text{if }\  \vp_{ref}\in\Ograsp 
    %         \\ 0 & \text{else} 
    %     \end{cases}
    % \end{split}
    %
    % \OriginSelection\left(
    %     \pref|g_0^{-1}\cdot \Oscene,\Ograsp
    % \right)
    % \propto 
    % n_r\left(
    %     \pref,\ g_0^{-1}\cdot \Oscene
    % \right)
    % \mathbbm{1}(\pref\in\Ograsp)
    %
    % \OriginSelection\left(
    %     \pref|g_0^{-1}\cdot \Oscene,\Ograsp
    % \right)
    % =\sum_{\vp\in\Ograsp}
    % n_r\left(
    %     \vp,\ g_0^{-1}\cdot \Oscene
    % \right)
    % \delta^{(3)}(\pref-\vp)
    %
    \begin{split}
        &\OriginSelection\left(
            \pref|g_0^{-1}\cdot \Oscene,\Ograsp
        \right) \\
        &\propto\sum_{\vp\in\Ograsp}
        n_r\left(
            \vp,\ g_0^{-1}\cdot \Oscene
        \right)
        \delta^{(3)}(\pref-\vp)
    \end{split}
\end{equation}
% For each point in $\Ograsp$, we first calculate the number of neighboring points in $g_0^{-1}\cdot \Oscene$. We then randomly sample a point from $\Ograsp$ as the origin of diffusion with the probability proportional to the number of neighbors. Although this specific frame origin selection mechanism allows tractable integration of \eqref{eqn:frame_selection_marginal}, we efficiently approximate it by  using the loss function in~\eqref{eqn:score_matching_loss} with a minibatch of a few $\vp_{ref}$. Note that this approach is more general and can be applied for an arbitrary frame selection mechanism whose integration of \eqref{eqn:frame_selection_marginal} is intractable. 
% where $\mathbbm{1}$ is the indicator function of the points in $\Ograsp$.
where $\delta^{(3)}(\vp)$ is the Dirac delta function on $\mathbb{R}^3$.
We find that this strategy enables our models to pay more attention to such contact-rich and relevant sub-geometries without explicit supervision.
%(see \Fig~\ref{fig:query_points} in \Supp~\ref{appndx:frame_selection}).
See \Supp~\ref{appndx:frame_selection} for more details.

\subsection{Architecture of Equivariant Descriptor Fields}
For faster sampling, we separate our implementation of EDFs into the feature extractor and the field model (see \Fig~\ref{fig:Unet}) as \citet{ryu2023equivariant} and \citet{chatzipantazis2022se}. The feature extractor is a deep $SE(3)$-equivariant GNN encoder that is run only once at the beginning of the denoising process. 
On the other hand, the field model is much shallower and faster GNN that is utilized for each denoising step.
It takes the encoded feature points from the feature extractor as input and computes the field value at a given query point.

For denoising, the receptive field of our model should cover the whole scene.
However, the original EDFs \citep{ryu2023equivariant} have small receptive fields due to memory constraints.
We address this issue with our U-Net-like multiscale architecture, which maintains a wide receptive field without losing local high-frequency details.
This increased receptive field enables Diffusion-EDFs to understand scene-level context.

In our multiscale EDF architecture, we use smaller message passing radius for small-scale points and larger radius for large-scale points.
To keep the number of graph edges constant, we apply point pooling to larger-scale points with \emph{Farthest Point Sampling} (FPS) algorithm \citep{qi2017pointnet++}. 
% The output multiscale feature points from this network is then fed into the field model.
For the field model, we find that a single layer is sufficient, although it is possible to stack multiple layers as \citet{chatzipantazis2022se}.
We use Equiformer \citep{liao2022equiformer} as the $SE(3)$-equivariant backbone GNN, with the addition of skip connections through point pooling layers.
See \Fig~\ref{fig:Unet} for an illustration of our architecture.
More details can be found in \Supp~\ref{appndx:architecture}.

\subsection{Score Model}
We use the weighted query points model similar to \citet{ryu2023equivariant} for $\rho(\vx|O)$
% \begin{align}
%     \rho(\vx|\Ograsp)=\sum_{n=1}^{N_{q}}w(\vx|\Ograsp)\delta^{(3)}(\vx-\vq_{n}(\Ograsp))
% \end{align}
\begin{align}
    \rho(\vx|\Ograsp)=\sum_{\vq\in Q(\Ograsp)}w(\vx|\Ograsp)\delta^{(3)}(\vx-\vq)
\end{align}
where $Q(\cdot):\Ograsp\mapsto\left\{\vq_{n}\right\}_{n=1}^{N_q}$ is the \emph{query points function} which outputs the set of $N_q$ query points, and $w(\cdot|\Ograsp):\mathbb{R}^3\rightarrow\mathbb{R}_{\geq 0}$ is the \emph{query weight field} that assigns weights to each query point. 
The query points function and query weight field are $SE(3)$-equivariant such that
\begin{align*}
    Q(\Dg\cdot \Ograsp)&=\left\{\Dg\,\vq_n|\vq_n\in Q(\Ograsp)\right\}&\forall \Dg\in SE(3)
    \\w(\vx|\Ograsp)&=w(\Dg\,\vx|\Dg\cdot\Ograsp)&\forall \Dg\in SE(3)
\end{align*}
We use FPS algorithm for $Q(\Ograsp)$. Although it is not strictly deterministic, we observe negligible impact from this stochasticity. For the implementation of the query weight field $w(\vx|\Obs)$, we use an EDF with a single scalar (type-0) output.
With this query points model, \eqref{eqn:nu_model} and \eqref{eqn:omg_model} become tractable summation forms 
\begin{align}
    \label{eqn:score_sum_lin}
    \begin{split}
        &\vs_{\nu;t}(g|\Oscene,\Ograsp)=
        \sum_{\vq\in Q(\Ograsp)}
        w(\vq|\Ograsp)\,\widetilde{\vs}_{\nu;t}(g,\vq|\Oscene,\Ograsp)
    \end{split} \\
    \label{eqn:score_sum_ang}
    \begin{split}
        &\vs_{\omega;t}(g|\Oscene,\Ograsp)=\sum_{\vq\in Q(\Ograsp)} w(\vq|\Ograsp)\,\widetilde{\vs}_{\omega;t}(g,\vq|\Oscene,\Ograsp) \\
        &\phantom{\qquad\qquad\,}+\sum_{\vq\in Q(\Ograsp)} w(\vq|\Ograsp)\,\vq\wedge\widetilde{\vs}_{\nu;t}(g,\vq|\Oscene,\Ograsp)
    \end{split}
\end{align}
% where $Q(\Ograsp)=\left\{\vq_{n}\left(\Ograsp\right)\right\}_{n=1}^{N_q}$ is the set of $N_q$ query points.

\section{Experiments and Results}
\label{sec:exp} 
\label{sec:sim_exp}
\label{sec:real_exp}

\begin{figure}[t]
  \centering
   \includegraphics[width=0.95\linewidth]{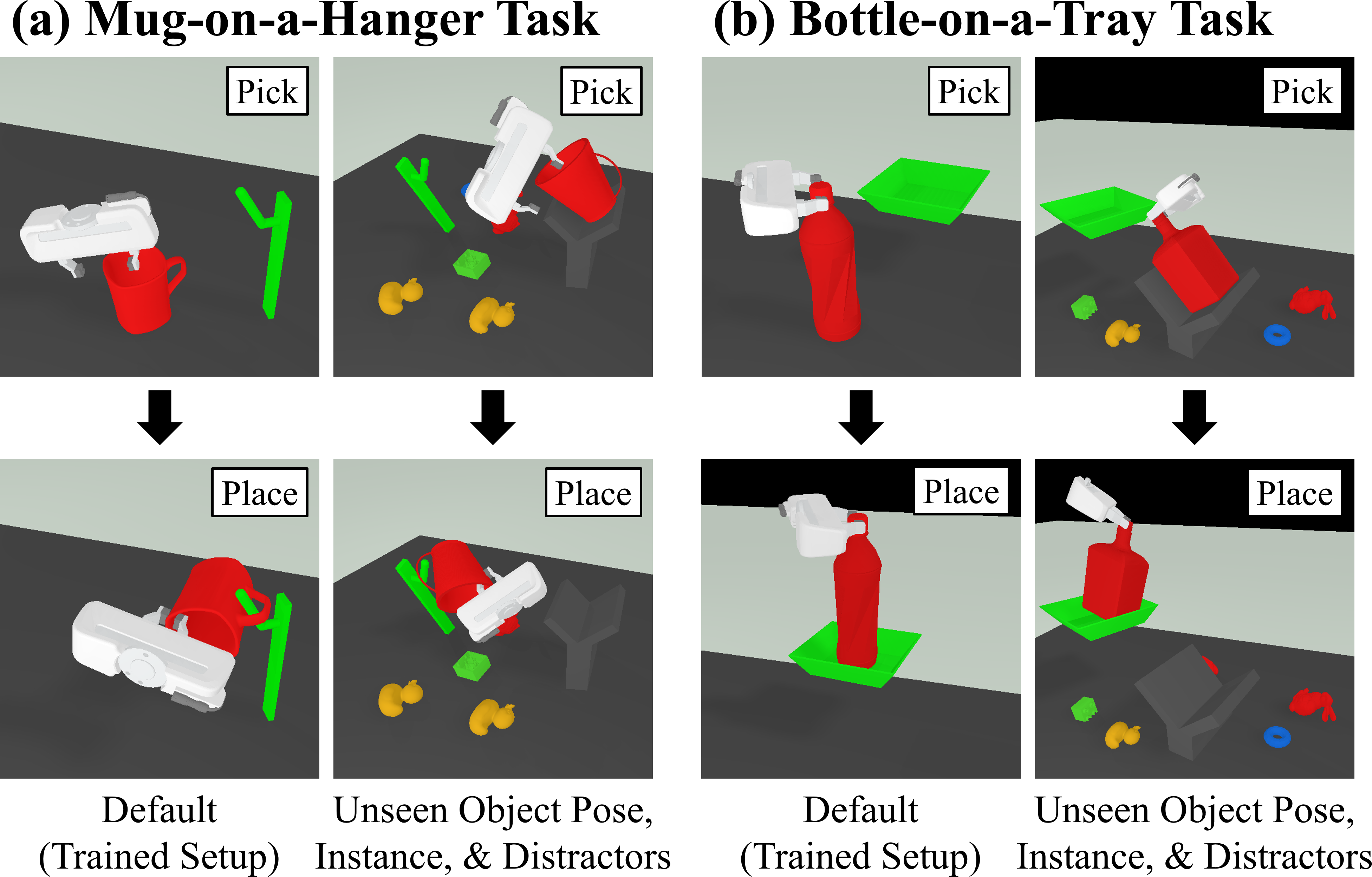}
   \caption{
        \textbf{Simulation Experiments.} (a) In the \emph{Mug-on-a-Hanger} task, a red mug should be picked up by its rim and placed on a green hanger by its handle. (b) In the \emph{Bottle-on-a-Tray} task, a red bottle should be picked up by its cap and placed on a green tray.
   }
   \label{fig:sim_exp}
   \vspace{-0.7\baselineskip}
\end{figure}

% \subsection{Simulation Benchmarks}
% \label{sec:sim_exp}
\paragraph{Simulation Benchmarks.}
We compare diffusion-EDFs with a state-of-the-art $SE(3)$-equivariant method (R-NDFs \citep{simeonov2023se}) and a state-of-the-art denoising diffusion-based method ($SE(3)$-Diffusion Fields \citep{urain2022se3dif}) under an evaluation protocol similar to \citet{simeonov2022neural,simeonov2023se}, \citet{ryu2023equivariant}, and \citet{biza2023one}.
% We evaluate the performance of each model following a similar protocol to that of \citet{simeonov2022neural,simeonov2023se,ryu2023equivariant,biza2023one}.
In particular, we measure the pick-and-place success rate for two different object categories: mugs and bottles (see \Fig~\ref{fig:sim_exp}).
We assess the generalizability of each method under four previously unseen scenarios: 1) novel object instances, 2) novel object poses, 3) novel clutters of distracting objects, and 4) all three combined. 
See \Supp~\ref{appndx:sim_exp} for more details on the experimental setup.

All the models are trained with ten task demonstrations performed by humans.
We train Diffusion-EDFs in a fully end-to-end manner without using any pre-training or object segmentation.
In contrast, we evaluate R-NDFs and $SE(3)$-Diffusion Fields for both with and without object segmentation pipelines.
% For $SE(3)$-Diffusion Fields, we use rotational augmentation for a fair comparison with $SE(3)$-equivariant methods.
For $SE(3)$-Diffusion Fields, we use rotational augmentation as they lack $SE(3)$-equivariance.
For R-NDFs, we additionally use category-specific pre-trained weights from the original implementation \citep{simeonov2023se}.
%, which is necessary to achieve reasonable success rates.
% Still, we observe that R-NDFs fail to place the mug on a rack due to the discrepancy of the shape of the rack in our experiment and the pre-trained ones. 
% Therefore, we do R-NDFs an additional favor of using the pre-trained rack instead of our rack for the evaluation.
% We turn off collision with the environment (except for the target pick/place objects) and observation noise to remove unnecessary influence of motion planning and point cloud processing pipelines, which are orthogonal to our research.
It took 20$\sim$45 minutes to train Diffusion-EDFs for single pick or place task with RTX 3090 GPU and i9-12900k CPU.

%%%%%JIWOOKIM%%%%%%
%%%%%CHANGED THE SEQUENCE?%%%%
% All the models are trained with ten task demonstrations performed by humans.
%We evaluate R-NDFs and $SE(3)$-Diffusion Fields for both with and without object segmentation pipelines.
% For $SE(3)$-Diffusion Fields, we use rotational augmentation as they lack $SE(3)$-equivariance.
% For R-NDFs, we additionally use category-specific pre-trained weights from the original implementation \citep{simeonov2023se}.
% In contrast, we trained Diffusion-EDFs in a fully end-to-end manner without using any pre-training or object segmentation. It took less than an hour to train Diffusion-EDFs in our local machine with RTX 3090 GPU and i9-12900k CPU.
%%%%%JIWOOKIM%%%%%%

%%%%%%%%%%%%%%%%%%%%%%%%%%%%%%%%%%%%%%%%%%%%%%%%%%%%%%%%%%%%%%%%%%%%%%%%%%%%%%%%%%%%%%%%%%%%%%%%%%%%
%%%%%%%%%%%%%%%%%%%%%%%%%%%%%%%%%%%%%%%%%%%%%%%%%%%%%%%%%%%%%%%%%%%%%%%%%%%%%%%%%%%%%%%%%%%%%%%%%%%%
% \afterpage{
    %%%%%%%%%%%%%%%%%%%%%%%%%%%%%%%%%%%%%%%%%%%%%%%%%%%%%%%%%%%%%%%%%%%%%%%%%%%%%%%%%%%%%%%%%%%%%%%%%%%%
\begin{table*}[t]
% \caption{Pick-and-place success rates in various out-of-distribution settings in simulated environment.}
% \vspace{-15pt}
% \label{tab:sota}
\begin{centering}
\setlength\tabcolsep{2pt}%
\setlength\dashlinedash{0.65pt}
\setlength\dashlinegap{2pt}
\small
\begin{tabularx}{\textwidth}{ 
  % Scenario
  c| %!{\vrule width 1pt}
  % Methods
  l 
  % Property
  |>{\centering\arraybackslash}X 
  >{\centering\arraybackslash}X 
  >{\centering\arraybackslash}X|
  % !{\vrule width 1pt}
  % Task 1
  >{\centering\arraybackslash}X 
  >{\centering\arraybackslash}X 
  >{\centering\arraybackslash}X 
  % Task 2
  >{\centering\arraybackslash}X 
  >{\centering\arraybackslash}X 
  >{\centering\arraybackslash}X }

%%%%%%%%%%%%%%%%%%%%%%%%%%%%%%%%%%%%%%%%%%%%%%%%%%%%%%%%%%%%%%%%%%
% hline
%%%%%%%%%%%%%%%%%%%%%%%%%%%%%%%%%%%%%%%%%%%%%%%%%%%%%%%%%%%%%%%%%%
\Xhline{0\arrayrulewidth}
% \cdashline{2-2} \cdashline{3-5} \cdashline{6-11}
&
\phantom{method}
&
\phantom{O} & \phantom{O} & \phantom{O}
&
\phantom{0.00} & \phantom{0.00} & \phantom{0.00}
& 
\phantom{0.00} & \phantom{0.00} & \phantom{0.00}
\\[-9pt]
%%%%%%%%%%%%%%%%%%%%%%%%%%%%%%%%%%%%%%%%%%%%%%%%%%%%%%%%%%%%%%%%%%
% Task Names
%%%%%%%%%%%%%%%%%%%%%%%%%%%%%%%%%%%%%%%%%%%%%%%%%%%%%%%%%%%%%%%%%%
\Xhline{3\arrayrulewidth}
\rule{0pt}{10pt}
\multirow{2}{*}{\textbf{Scenario}}
&
\multicolumn{1}{c|}{\multirow{2}{*}{\textbf{Method}}}
& 
% \multirow{2}{*}{\shortstack[c]{\textbf{Pre-}\\ \textbf{Training}}} 
\multirow{2}{*}{\scriptsize \shortstack[c]{Without \\ Pretraining}}
& 
% \multirow{2}{*}{\shortstack[c]{\textbf{Obj.}\\ \textbf{Seg.}}}
\multirow{2}{*}{\scriptsize \shortstack[c]{Without \\ Obj. Seg.}}
& 
% \multirow{2}{*}{\shortstack[c]{\textbf{Rot.}\\ \textbf{Aug.}}} 
\multirow{2}{*}{\scriptsize \shortstack[c]{Without \\ Rot. Aug.}}
& 
\multicolumn{3}{c}{\textbf{Mug}} 
& 
\multicolumn{3}{c}{\textbf{Bottle}}
\\

%%%%%%%%%%%%%%%%%%%%%%%%%%%%%%%%%%%%%%%%%%%%%%%%%%%%%%%%%%%%%%%%%%
% Subtask Names
%%%%%%%%%%%%%%%%%%%%%%%%%%%%%%%%%%%%%%%%%%%%%%%%%%%%%%%%%%%%%%%%%%
% \cline{2-9} \multicolumn{9}{c}{}\\[-9pt] % Subtask seperator

% Scenario
&
% Method
& 
% Pretrained
& 
% Obj. Seg.
&
% Rot. Aug.
&
\multicolumn{1}{c}{Pick} & \multicolumn{1}{c}{Place} & \multicolumn{1}{c}{\textbf{Total}} 
& 
\multicolumn{1}{c}{Pick} & \multicolumn{1}{c}{Place} & \multicolumn{1}{c}{\textbf{Total}} 
\\

%%%%%%%%%%%%%%%%%%%%%%%%%%%%%%%%%%%%%%%%%%%%%%%%%%%%%%%%%%%%%%%%%%
% Default (Trained Setup)
%%%%%%%%%%%%%%%%%%%%%%%%%%%%%%%%%%%%%%%%%%%%%%%%%%%%%%%%%%%%%%%%%%
    %%%%%%%%%%%%%%%%%%%%%%%%%%%%%%%%%%%%%%%%%%%%%%%%%%%%%%%%%%%%%%%%%%
    % hline
    %%%%%%%%%%%%%%%%%%%%%%%%%%%%%%%%%%%%%%%%%%%%%%%%%%%%%%%%%%%%%%%%%%
    % \Xhline{3\arrayrulewidth}
    \hline
    \multirow{8}{*}{
        \shortstack[c]{
            \textbf{Default}
            \\ 
            \textbf{(Trained Setup)}
            \\
            \rule{0pt}{20pt} 
        }
    }
    &
    \phantom{method}
    &
    \phantom{O} & \phantom{O} & \phantom{O}
    &
    \phantom{0.00} & \phantom{0.00} & \phantom{0.00}
    & 
    \phantom{0.00} & \phantom{0.00} & \phantom{0.00}
    \\[-9pt]
    %%%%%%%%%%%%%%%%%%%%%%%%%%%%%%%%%%%%%%%%%%%%%%%%%%%%%%%%%%%%%%%%%%
    % R-NDFs
    %%%%%%%%%%%%%%%%%%%%%%%%%%%%%%%%%%%%%%%%%%%%%%%%%%%%%%%%%%%%%%%%%%
    &
    \multirow{2}{*}{ R-NDFs \citep{simeonov2023se}}
    &
    \propYes & \propYes & \propNo
    &
    0.83 & \textbf{0.97} & 0.81 
    & 
    0.91 & 0.73 & 0.67 
    \\
    &
    & 
    \propYes & \propNo & \propNo
    &
    0.00 & 0.00 & 0.00   
    &
    0.00 & 0.00 & 0.00   
    \\
    %%%%%%%%%%%%%%%%%%%%%%%%%%%%%%%%%%%%%%%%%%%%%%%%%%%%%%%%%%%%%%%%%%
    % hline
    %%%%%%%%%%%%%%%%%%%%%%%%%%%%%%%%%%%%%%%%%%%%%%%%%%%%%%%%%%%%%%%%%%
    \cdashline{2-11}
    % \cdashline{2-2} \cdashline{3-5} \cdashline{6-11}
    &
    \phantom{method}
    &
    \phantom{O} & \phantom{O} & \phantom{O}
    &
    \phantom{0.00} & \phantom{0.00} & \phantom{0.00}
    & 
    \phantom{0.00} & \phantom{0.00} & \phantom{0.00}
    \\[-9pt]
    %%%%%%%%%%%%%%%%%%%%%%%%%%%%%%%%%%%%%%%%%%%%%%%%%%%%%%%%%%%%%%%%%%
    % SE(3)-DiffusionFields
    %%%%%%%%%%%%%%%%%%%%%%%%%%%%%%%%%%%%%%%%%%%%%%%%%%%%%%%%%%%%%%%%%%
    &
    \multirow{2}{*}{ SE(3)-DiffusionFields \citep{urain2022se3dif}}
    & 
    \propNo & \propYes & \propYes
    &
    0.75 & {\notavailable} & {\notavailable} 
    & 
    0.47 & {\notavailable} & {\notavailable} 
    \\
    &
    & 
    \propNo & \propNo & \propYes
    &
    0.11 & {\notavailable} & {\notavailable} 
    & 
    0.01 & {\notavailable} & {\notavailable}   
    \\
    %%%%%%%%%%%%%%%%%%%%%%%%%%%%%%%%%%%%%%%%%%%%%%%%%%%%%%%%%%%%%%%%%%
    % hline
    %%%%%%%%%%%%%%%%%%%%%%%%%%%%%%%%%%%%%%%%%%%%%%%%%%%%%%%%%%%%%%%%%%
    \cdashline{2-11}
    &
    \phantom{method}
    &
    \phantom{O} & \phantom{O} & \phantom{O}
    &
    \phantom{0.00} & \phantom{0.00} & \phantom{0.00}
    & 
    \phantom{0.00} & \phantom{0.00} & \phantom{0.00}
    \\[-9pt]
    %%%%%%%%%%%%%%%%%%%%%%%%%%%%%%%%%%%%%%%%%%%%%%%%%%%%%%%%%%%%%%%%%%
    % Diffusion-EDFs
    %%%%%%%%%%%%%%%%%%%%%%%%%%%%%%%%%%%%%%%%%%%%%%%%%%%%%%%%%%%%%%%%%%
    &
    { \textbf{Diffusion-EDFs} \textbf{(Ours)}}
    & 
    \propNo & \propNo & \propNo
    &
    \textbf{0.99} & 0.96 & \textbf{0.95} 
    & 
    \textbf{0.97} & \textbf{0.85} & \textbf{0.83} 
    \\

%%%%%%%%%%%%%%%%%%%%%%%%%%%%%%%%%%%%%%%%%%%%%%%%%%%%%%%%%%%%%%%%%%
% Previously Unseen Instances
%%%%%%%%%%%%%%%%%%%%%%%%%%%%%%%%%%%%%%%%%%%%%%%%%%%%%%%%%%%%%%%%%%
    %%%%%%%%%%%%%%%%%%%%%%%%%%%%%%%%%%%%%%%%%%%%%%%%%%%%%%%%%%%%%%%%%%
    % hline
    %%%%%%%%%%%%%%%%%%%%%%%%%%%%%%%%%%%%%%%%%%%%%%%%%%%%%%%%%%%%%%%%%%
    % \Xhline{3\arrayrulewidth}
    \hline
    \multirow{8}{*}{
        \shortstack[c]{
            \textbf{Previously}
            \\ 
            \textbf{Unseen}
            \\
            \textbf{Instances}
            \\
            \rule{0pt}{20pt} 
        }
    }
    &
    \phantom{method}
    &
    \phantom{O} & \phantom{O} & \phantom{O}
    &
    \phantom{0.00} & \phantom{0.00} & \phantom{0.00}
    & 
    \phantom{0.00} & \phantom{0.00} & \phantom{0.00}
    \\[-9pt]
    %%%%%%%%%%%%%%%%%%%%%%%%%%%%%%%%%%%%%%%%%%%%%%%%%%%%%%%%%%%%%%%%%%
    % R-NDFs
    %%%%%%%%%%%%%%%%%%%%%%%%%%%%%%%%%%%%%%%%%%%%%%%%%%%%%%%%%%%%%%%%%%
    &
    \multirow{2}{*}{ R-NDFs \citep{simeonov2023se}}
    &
    \propYes & \propYes & \propNo
    &
    0.73 & 0.70 & 0.51 
    & 
    0.90 & 0.87 & 0.79
    \\
    &
    & 
    \propYes & \propNo & \propNo
    &
    0.00 & 0.00 & 0.00   
    &
    0.00 & 0.00 & 0.00   
    \\
    %%%%%%%%%%%%%%%%%%%%%%%%%%%%%%%%%%%%%%%%%%%%%%%%%%%%%%%%%%%%%%%%%%
    % hline
    %%%%%%%%%%%%%%%%%%%%%%%%%%%%%%%%%%%%%%%%%%%%%%%%%%%%%%%%%%%%%%%%%%
    \cdashline{2-11}
    % \cdashline{2-2} \cdashline{3-5} \cdashline{6-11}
    &
    \phantom{method}
    &
    \phantom{O} & \phantom{O} & \phantom{O}
    &
    \phantom{0.00} & \phantom{0.00} & \phantom{0.00}
    & 
    \phantom{0.00} & \phantom{0.00} & \phantom{0.00}
    \\[-9pt]
    %%%%%%%%%%%%%%%%%%%%%%%%%%%%%%%%%%%%%%%%%%%%%%%%%%%%%%%%%%%%%%%%%%
    % SE(3)-DiffusionFields
    %%%%%%%%%%%%%%%%%%%%%%%%%%%%%%%%%%%%%%%%%%%%%%%%%%%%%%%%%%%%%%%%%%
    &
    \multirow{2}{*}{ SE(3)-DiffusionFields \citep{urain2022se3dif}}
    & 
    \propNo & \propYes & \propYes
    &
    0.55 & {\notavailable} & {\notavailable} 
    & 
    0.57 & {\notavailable} & {\notavailable} 
    \\
    &
    & 
    \propNo & \propNo & \propYes
    &
    0.14 & {\notavailable} & {\notavailable} 
    & 
    0.00 & {\notavailable} & {\notavailable}   
    \\
    %%%%%%%%%%%%%%%%%%%%%%%%%%%%%%%%%%%%%%%%%%%%%%%%%%%%%%%%%%%%%%%%%%
    % hline
    %%%%%%%%%%%%%%%%%%%%%%%%%%%%%%%%%%%%%%%%%%%%%%%%%%%%%%%%%%%%%%%%%%
    \cdashline{2-11}
    &
    \phantom{method}
    &
    \phantom{O} & \phantom{O} & \phantom{O}
    &
    \phantom{0.00} & \phantom{0.00} & \phantom{0.00}
    & 
    \phantom{0.00} & \phantom{0.00} & \phantom{0.00}
    \\[-9pt]
    %%%%%%%%%%%%%%%%%%%%%%%%%%%%%%%%%%%%%%%%%%%%%%%%%%%%%%%%%%%%%%%%%%
    % Diffusion-EDFs
    %%%%%%%%%%%%%%%%%%%%%%%%%%%%%%%%%%%%%%%%%%%%%%%%%%%%%%%%%%%%%%%%%%
    &
    { \textbf{Diffusion-EDFs} \textbf{(Ours)}}
    & 
    \propNo & \propNo & \propNo
    &
    \textbf{0.96} & \textbf{0.96} & \textbf{0.92} 
    & 
    \textbf{0.99} & \textbf{0.91} & \textbf{0.90} 
    \\

%%%%%%%%%%%%%%%%%%%%%%%%%%%%%%%%%%%%%%%%%%%%%%%%%%%%%%%%%%%%%%%%%%
% Previously Unseen Poses
%%%%%%%%%%%%%%%%%%%%%%%%%%%%%%%%%%%%%%%%%%%%%%%%%%%%%%%%%%%%%%%%%%
    %%%%%%%%%%%%%%%%%%%%%%%%%%%%%%%%%%%%%%%%%%%%%%%%%%%%%%%%%%%%%%%%%%
    % hline
    %%%%%%%%%%%%%%%%%%%%%%%%%%%%%%%%%%%%%%%%%%%%%%%%%%%%%%%%%%%%%%%%%%
    % \Xhline{3\arrayrulewidth}
    \hline
    \multirow{8}{*}{
        \shortstack[c]{
            \textbf{Previously}
            \\ 
            \textbf{Unseen}
            \\
            \textbf{Poses}
            \\
            \rule{0pt}{20pt} 
        }
    }
    &
    \phantom{method}
    &
    \phantom{O} & \phantom{O} & \phantom{O}
    &
    \phantom{0.00} & \phantom{0.00} & \phantom{0.00}
    & 
    \phantom{0.00} & \phantom{0.00} & \phantom{0.00}
    \\[-9pt]
    %%%%%%%%%%%%%%%%%%%%%%%%%%%%%%%%%%%%%%%%%%%%%%%%%%%%%%%%%%%%%%%%%%
    % R-NDFs
    %%%%%%%%%%%%%%%%%%%%%%%%%%%%%%%%%%%%%%%%%%%%%%%%%%%%%%%%%%%%%%%%%%
    &
    \multirow{2}{*}{ R-NDFs \citep{simeonov2023se}}
    &
    \propYes & \propYes & \propNo
    &
    0.84 & 0.93 & 0.78 
    & 
    0.65 & 0.72 & 0.47 
    \\
    &
    & 
    \propYes & \propNo & \propNo
    &
    0.00 & 0.00 & 0.00   
    &
    0.00 & 0.00 & 0.00   
    \\
    %%%%%%%%%%%%%%%%%%%%%%%%%%%%%%%%%%%%%%%%%%%%%%%%%%%%%%%%%%%%%%%%%%
    % hline
    %%%%%%%%%%%%%%%%%%%%%%%%%%%%%%%%%%%%%%%%%%%%%%%%%%%%%%%%%%%%%%%%%%
    \cdashline{2-11}
    % \cdashline{2-2} \cdashline{3-5} \cdashline{6-11}
    &
    \phantom{method}
    &
    \phantom{O} & \phantom{O} & \phantom{O}
    &
    \phantom{0.00} & \phantom{0.00} & \phantom{0.00}
    & 
    \phantom{0.00} & \phantom{0.00} & \phantom{0.00}
    \\[-9pt]
    %%%%%%%%%%%%%%%%%%%%%%%%%%%%%%%%%%%%%%%%%%%%%%%%%%%%%%%%%%%%%%%%%%
    % SE(3)-DiffusionFields
    %%%%%%%%%%%%%%%%%%%%%%%%%%%%%%%%%%%%%%%%%%%%%%%%%%%%%%%%%%%%%%%%%%
    &
    \multirow{2}{*}{ SE(3)-DiffusionFields \citep{urain2022se3dif}}
    & 
    \propNo & \propYes & \propYes
    &
    0.75 & {\notavailable} & {\notavailable} 
    & 
    0.47 & {\notavailable} & {\notavailable} 
    \\
    &
    & 
    \propNo & \propNo & \propYes
    &
    0.00 & {\notavailable} & {\notavailable} 
    & 
    0.04 & {\notavailable} & {\notavailable}   
    \\
    %%%%%%%%%%%%%%%%%%%%%%%%%%%%%%%%%%%%%%%%%%%%%%%%%%%%%%%%%%%%%%%%%%
    % hline
    %%%%%%%%%%%%%%%%%%%%%%%%%%%%%%%%%%%%%%%%%%%%%%%%%%%%%%%%%%%%%%%%%%
    \cdashline{2-11}
    &
    \phantom{method}
    &
    \phantom{O} & \phantom{O} & \phantom{O}
    &
    \phantom{0.00} & \phantom{0.00} & \phantom{0.00}
    & 
    \phantom{0.00} & \phantom{0.00} & \phantom{0.00}
    \\[-9pt]
    %%%%%%%%%%%%%%%%%%%%%%%%%%%%%%%%%%%%%%%%%%%%%%%%%%%%%%%%%%%%%%%%%%
    % Diffusion-EDFs
    %%%%%%%%%%%%%%%%%%%%%%%%%%%%%%%%%%%%%%%%%%%%%%%%%%%%%%%%%%%%%%%%%%
    &
    { \textbf{Diffusion-EDFs} \textbf{(Ours)}}
    & 
    \propNo & \propNo & \propNo
    &
    \textbf{0.98} & \textbf{0.98} & \textbf{0.96}
    & 
    \textbf{0.98} & \textbf{0.81} & \textbf{0.79}
    \\

%%%%%%%%%%%%%%%%%%%%%%%%%%%%%%%%%%%%%%%%%%%%%%%%%%%%%%%%%%%%%%%%%%
% Previously Unseen Clutters
%%%%%%%%%%%%%%%%%%%%%%%%%%%%%%%%%%%%%%%%%%%%%%%%%%%%%%%%%%%%%%%%%%
    %%%%%%%%%%%%%%%%%%%%%%%%%%%%%%%%%%%%%%%%%%%%%%%%%%%%%%%%%%%%%%%%%%
    % hline
    %%%%%%%%%%%%%%%%%%%%%%%%%%%%%%%%%%%%%%%%%%%%%%%%%%%%%%%%%%%%%%%%%%
    % \Xhline{3\arrayrulewidth}
    \hline
    \multirow{6}{*}{
        \shortstack[c]{
            \textbf{Previously}
            \\ 
            \textbf{Unseen}
            \\
            \textbf{Clutters}$^{\tablefootnotemarkone}$
            \\
            \rule{0pt}{20pt} 
        }
    }
    &
    \phantom{method}
    &
    \phantom{O} & \phantom{O} & \phantom{O}
    &
    \phantom{0.00} & \phantom{0.00} & \phantom{0.00}
    & 
    \phantom{0.00} & \phantom{0.00} & \phantom{0.00}
    \\[-9pt]
    %%%%%%%%%%%%%%%%%%%%%%%%%%%%%%%%%%%%%%%%%%%%%%%%%%%%%%%%%%%%%%%%%%
    % R-NDFs
    %%%%%%%%%%%%%%%%%%%%%%%%%%%%%%%%%%%%%%%%%%%%%%%%%%%%%%%%%%%%%%%%%%
    &
    \multirow{1}{*}{ R-NDFs \citep{simeonov2023se}}
    &
    \propYes & \propNo & \propNo
    &
    0.00 & 0.00 & 0.00 
    & 
    0.00 & 0.00 & 0.00 
    \\
    %%%%%%%%%%%%%%%%%%%%%%%%%%%%%%%%%%%%%%%%%%%%%%%%%%%%%%%%%%%%%%%%%%
    % hline
    %%%%%%%%%%%%%%%%%%%%%%%%%%%%%%%%%%%%%%%%%%%%%%%%%%%%%%%%%%%%%%%%%%
    \cdashline{2-11}
    % \cdashline{2-2} \cdashline{3-5} \cdashline{6-11}
    &
    \phantom{method}
    &
    \phantom{O} & \phantom{O} & \phantom{O}
    &
    \phantom{0.00} & \phantom{0.00} & \phantom{0.00}
    & 
    \phantom{0.00} & \phantom{0.00} & \phantom{0.00}
    \\[-9pt]
    %%%%%%%%%%%%%%%%%%%%%%%%%%%%%%%%%%%%%%%%%%%%%%%%%%%%%%%%%%%%%%%%%%
    % SE(3)-DiffusionFields
    %%%%%%%%%%%%%%%%%%%%%%%%%%%%%%%%%%%%%%%%%%%%%%%%%%%%%%%%%%%%%%%%%%
    &
    \multirow{1}{*}{ SE(3)-DiffusionFields \citep{urain2022se3dif}}
    & 
    \propNo & \propNo & \propYes
    &
    0.06 & {\notavailable} & {\notavailable} 
    & 
    0.03 & {\notavailable} & {\notavailable} 
    \\
    %%%%%%%%%%%%%%%%%%%%%%%%%%%%%%%%%%%%%%%%%%%%%%%%%%%%%%%%%%%%%%%%%%
    % hline
    %%%%%%%%%%%%%%%%%%%%%%%%%%%%%%%%%%%%%%%%%%%%%%%%%%%%%%%%%%%%%%%%%%
    \cdashline{2-11}
    &
    \phantom{method}
    &
    \phantom{O} & \phantom{O} & \phantom{O}
    &
    \phantom{0.00} & \phantom{0.00} & \phantom{0.00}
    & 
    \phantom{0.00} & \phantom{0.00} & \phantom{0.00}
    \\[-9pt]
    %%%%%%%%%%%%%%%%%%%%%%%%%%%%%%%%%%%%%%%%%%%%%%%%%%%%%%%%%%%%%%%%%%
    % Diffusion-EDFs
    %%%%%%%%%%%%%%%%%%%%%%%%%%%%%%%%%%%%%%%%%%%%%%%%%%%%%%%%%%%%%%%%%%
    &
    { \textbf{Diffusion-EDFs} \textbf{(Ours)}}
    & 
    \propNo & \propNo & \propNo
    &
    \textbf{0.91} & \textbf{1.00} & \textbf{0.91}
    & 
    \textbf{0.96} & \textbf{0.91} & \textbf{0.87} 
    \\

%%%%%%%%%%%%%%%%%%%%%%%%%%%%%%%%%%%%%%%%%%%%%%%%%%%%%%%%%%%%%%%%%%
% Previously Unseen Instances, Poses & Clutters
%%%%%%%%%%%%%%%%%%%%%%%%%%%%%%%%%%%%%%%%%%%%%%%%%%%%%%%%%%%%%%%%%%
    %%%%%%%%%%%%%%%%%%%%%%%%%%%%%%%%%%%%%%%%%%%%%%%%%%%%%%%%%%%%%%%%%%
    % hline
    %%%%%%%%%%%%%%%%%%%%%%%%%%%%%%%%%%%%%%%%%%%%%%%%%%%%%%%%%%%%%%%%%%
    % \Xhline{3\arrayrulewidth}
    \hline
    \multirow{8}{*}{
        \shortstack[c]{
            \textbf{Previously}
            \\ 
            \textbf{Unseen}
            \\
            \textbf{Instances,}
            \\
            \textbf{Poses,}
            \\
            \textbf{\& Clutters}$^{\tablefootnotemarkone}$
            \\
            \rule{0pt}{20pt} 
        }
    }
    &
    \phantom{method}
    &
    \phantom{O} & \phantom{O} & \phantom{O}
    &
    \phantom{0.00} & \phantom{0.00} & \phantom{0.00}
    & 
    \phantom{0.00} & \phantom{0.00} & \phantom{0.00}
    \\[-9pt]
    %%%%%%%%%%%%%%%%%%%%%%%%%%%%%%%%%%%%%%%%%%%%%%%%%%%%%%%%%%%%%%%%%%
    % R-NDFs
    %%%%%%%%%%%%%%%%%%%%%%%%%%%%%%%%%%%%%%%%%%%%%%%%%%%%%%%%%%%%%%%%%%
    &
    \multirow{2}{*}{ R-NDFs \citep{simeonov2023se}}
    &
    \propYes & \propYes & \propNo
    &
    0.71$^{\tablefootnotemarkone}$ & 0.75$^{\tablefootnotemarkone}$ & 0.53$^{\tablefootnotemarkone}$ 
    & 
    0.85$^{\tablefootnotemarkone}$ & 0.84$^{\tablefootnotemarkone}$ & 0.72$^{\tablefootnotemarkone}$
    \\
    &
    & 
    \propYes & \propNo & \propNo
    &
    0.00 & 0.00 & 0.00   
    &
    0.00 & 0.00 & 0.00   
    \\
    %%%%%%%%%%%%%%%%%%%%%%%%%%%%%%%%%%%%%%%%%%%%%%%%%%%%%%%%%%%%%%%%%%
    % hline
    %%%%%%%%%%%%%%%%%%%%%%%%%%%%%%%%%%%%%%%%%%%%%%%%%%%%%%%%%%%%%%%%%%
    \cdashline{2-11}
    % \cdashline{2-2} \cdashline{3-5} \cdashline{6-11}
    &
    \phantom{method}
    &
    \phantom{O} & \phantom{O} & \phantom{O}
    &
    \phantom{0.00} & \phantom{0.00} & \phantom{0.00}
    & 
    \phantom{0.00} & \phantom{0.00} & \phantom{0.00}
    \\[-9pt]
    %%%%%%%%%%%%%%%%%%%%%%%%%%%%%%%%%%%%%%%%%%%%%%%%%%%%%%%%%%%%%%%%%%
    % SE(3)-DiffusionFields
    %%%%%%%%%%%%%%%%%%%%%%%%%%%%%%%%%%%%%%%%%%%%%%%%%%%%%%%%%%%%%%%%%%
    &
    \multirow{2}{*}{ SE(3)-DiffusionFields \citep{urain2022se3dif}}
    & 
    \propNo & \propYes & \propYes
    &
    0.58$^{\tablefootnotemarkone}$ & {\notavailable} & {\notavailable}
    & 
    0.59$^{\tablefootnotemarkone}$ & {\notavailable} & {\notavailable}
    \\
    &
    & 
    \propNo & \propNo & \propYes
    &
    0.03& {\notavailable} & {\notavailable}
    & 
    0.00& {\notavailable} & {\notavailable}   
    \\
    %%%%%%%%%%%%%%%%%%%%%%%%%%%%%%%%%%%%%%%%%%%%%%%%%%%%%%%%%%%%%%%%%%
    % hline
    %%%%%%%%%%%%%%%%%%%%%%%%%%%%%%%%%%%%%%%%%%%%%%%%%%%%%%%%%%%%%%%%%%
    \cdashline{2-11}
    &
    \phantom{method}
    &
    \phantom{O} & \phantom{O} & \phantom{O}
    &
    \phantom{0.00} & \phantom{0.00} & \phantom{0.00}
    & 
    \phantom{0.00} & \phantom{0.00} & \phantom{0.00}
    \\[-9pt]
    %%%%%%%%%%%%%%%%%%%%%%%%%%%%%%%%%%%%%%%%%%%%%%%%%%%%%%%%%%%%%%%%%%
    % Diffusion-EDFs
    %%%%%%%%%%%%%%%%%%%%%%%%%%%%%%%%%%%%%%%%%%%%%%%%%%%%%%%%%%%%%%%%%%
    &
    { \textbf{Diffusion-EDFs} \textbf{(Ours)}}
    & 
    \propNo & \propNo & \propNo
    &
    \textbf{0.89} & \textbf{0.89} & \textbf{0.79} 
    & 
    \textbf{0.98} & \textbf{0.89} & \textbf{0.87} 
    \\

%%%%%%%%%%%%%%%%%%%%%%%%%%%%%%%%%%%%%%%%%%%%%%%%%%%%%%%%%%%%%%%%%%
\Xhline{3\arrayrulewidth}
\end{tabularx}
\end{centering}
\footnotesize{$^{\tablefootnotemarkone}$Models with segmented inputs are tested without cluttered objects to guarantee perfect object segmentation.}
\caption{Pick-and-place success rates in various out-of-distribution settings in simulated environment.}
\label{tab:sota}
\end{table*}
%%%%%%%%%%%%%%%%%%%%%%%%%%%%%%%%%%%%%%%%%%%%%%%%%%%%%%%%%%%%%%%%%%%%%%%%%%%%%%%%%%%%%%%%%%%%%%%%%%%%
% }
%%%%%%%%%%%%%%%%%%%%%%%%%%%%%%%%%%%%%%%%%%%%%%%%%%%%%%%%%%%%%%%%%%%%%%%%%%%%%%%%%%%%%%%%%%%%%%%%%%%%
% \afterpage{
    \begin{figure*}[t]
      \centering
      \includegraphics[width=1\textwidth]{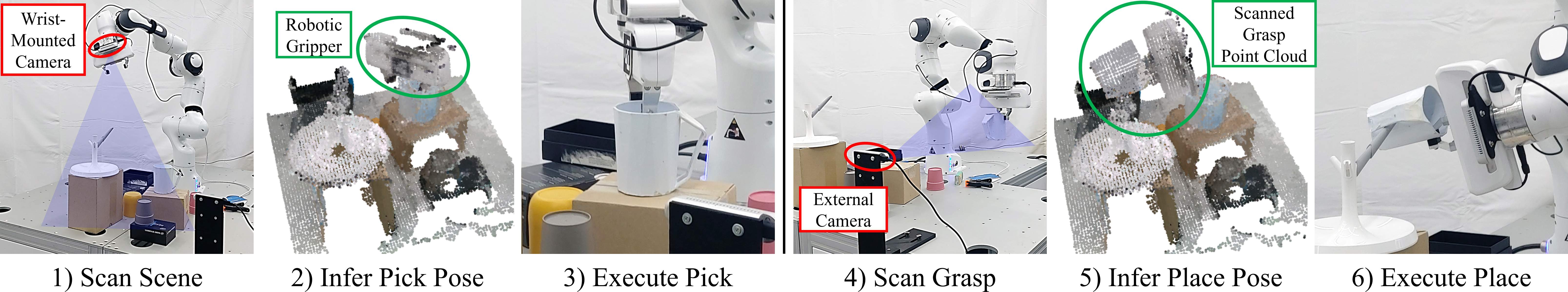}
      \caption{
        \textbf{Real Hardware Experiment Pipeline} 1)~The scene point cloud is observed via 3D SLAM algorithm with the wrist-mounted RGB-D Camera. 2)~Diffusion-EDFs infer the gripper pose to pick up the target object. 3)~The robot executes picking if the pose is reachable. 4)~The grasp point cloud is scanned with an external RGB-D camera. 
        5) Diffusion-EDFs infer the gripper pose to place the grasped object on the placement target.
        6) The robot executes placement if the pose is reachable. 
        See \Supp~\ref{appndx:real_exp} for more details.
      }
      \label{fig:pipelines}
    \end{figure*}
% }
%%%%%%%%%%%%%%%%%%%%%%%%%%%%%%%%%%%%%%%%%%%%%%%%%%%%%%%%%%%%%%%%%%%%%%%%%%%%%%%%%%%%%%%%%%%%%%%%%%%%
% %%%%%%%%%%%%%%%%%%%%%%%%%%%%%%%%%%%%%%%%%%%%%%%%%%%%%%%%%%%%%%%%%%%%%%%%%%%%%%%%%%%%%%%%%%%%%%%%%%%%
% \afterpage{
%     \begin{figure}[t]
%       \centering
%       \includegraphics[width=0.95\linewidth]{images/pipelines_onecol_proc.png}
      % \caption{
      %   \textbf{Real Hardware Experiment Pipeline} 1)~The scene point cloud is observed via 3D SLAM algorithm with the wrist-mounted RGB-D Camera. 2)~Diffusion-EDFs infer the gripper pose to pick up the target object. 3)~The robot executes picking if the pose is reachable. 4)~The grasp point cloud is scanned with an external RGB-D camera. 
      %   5) Diffusion-EDFs infer the gripper pose to place the grasped object on the placement target.
      %   6) The robot executes placement if the pose is reachable. 
      %   See \Supp~\ref{appndx:real_exp} for more details.
      % }
%       \label{fig:pipelines}
%     \end{figure}
% }
% %%%%%%%%%%%%%%%%%%%%%%%%%%%%%%%%%%%%%%%%%%%%%%%%%%%%%%%%%%%%%%%%%%%%%%%%%%%%%%%%%%%%%%%%%%%%%%%%%%%%

% \paragraph{Results.}
As shown in \Table~\ref{tab:sota}, Diffusion-EDFs consistently outperform both the $SE(3)$-equivariant baseline (R-NDFs \citep{simeonov2023se}) and diffusion model baseline ($SE(3)$-DiffusionFields \citep{urain2022se3dif}) in almost all scenarios, despite not being provided with pre-training or segmented inputs.
In particular, the baseline models completely fail with unsegmented observations.
Without object segmentation, R-NDFs achieve zero success rates due to the lack of locality in their method design \citep{ryu2023equivariant,kim2023robotic,chun2023local}.
% $SE(3)$-DiffusionFields are also unable to achieve a success rate higher than 15\% without object segmentation.
While slightly better than R-NDFs, $SE(3)$-DiffusionFields also record low success rates, presumably due to the lack of $SE(3)$-equivariance.
% Notably, the performance of unsegmented $SE(3)$-DiffusionFields significantly drops with unseen object poses, whereas its segmented counterpart is unaffected due to rotational augmentation during training.
On the other hand, Diffusion-EDFs maintain total success rates around 80\% even in the most adversarial scenarios due to the local equivariance \citep{ryu2023equivariant,kim2023robotic} inherited from EDFs and our local contact-based diffusion frame selection mechanism.

% It should be noted that the ability to infer without object segmentation is important not only because of its convenience.
% It allows the model to understand \emph{scene-level contexts} beyond a single target object.
% This benefit will become evident in our subsequent experiments with real hardware.

%%%%%%%%%%%%%%%%%%%%%%%%%%%%%%%%%%%%%%%%%%%%%%%%%%%%%%%%%%%%%%%%%%%%%%%%%%%%%%%%%%%%%%%%%%%%%%%%%%%%
% \afterpage{
    \begin{figure}[t]
      \centering
       \includegraphics[width=1\linewidth]{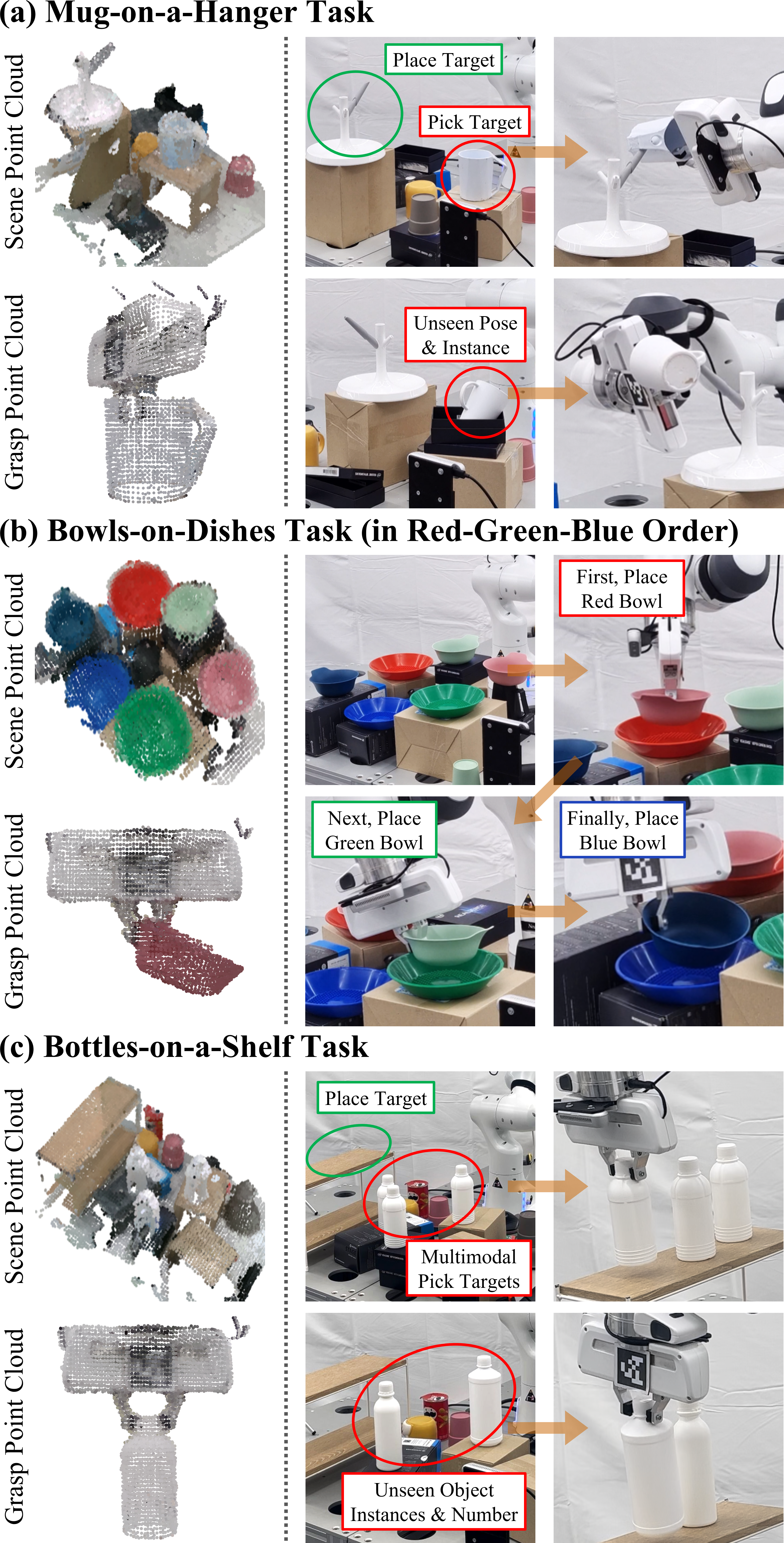}
       \caption{
            \textbf{Real Hardware Experiments.} (a) In the \emph{mug-on-a-hanger} task, the white mug must be picked and placed on the white hanger. (b) In the \emph{bowls-on-dishes} task, the bowls must be picked and placed on the dishes of matching color in red-green-blue order. (c) In the \emph{bottles-on-a-shelf} task, multiple bottles must be picked and placed on the shelf one by one.
            The experimental results can be found in the supplementary materials and our project website: \url{https://sites.google.com/view/diffusion-edfs/home}.
       }
       \label{fig:real_exp}
    \end{figure}
% }
%%%%%%%%%%%%%%%%%%%%%%%%%%%%%%%%%%%%%%%%%%%%%%%%%%%%%%%%%%%%%%%%%%%%%%%%%%%%%%%%%%%%%%%%%%%%%%%%%%%

% \subsection{Real Hardware Experiments}
% \label{sec:real_exp}
\vspace{-0.7\baselineskip}
\paragraph{Real Hardware Experiments.}
We further evaluate our Diffusion-EDFs on three real-world tasks: the \emph{mug-on-a-hanger} task, \emph{bowls-on-dishes} task, and \emph{bottles-on-a-shelf} task.
We illustrate these tasks in \Fig~\ref{fig:real_exp}, and the experiment pipeline in \Fig~\ref{fig:pipelines}.
More details on the training and evaluation setups can be found in \Supp~\ref{appndx:real_exp}.

The mug-on-a-hanger task is similar to the one in the simulation benchmark.
% The radius of the handle of the mugs in this task ranges from 1 to 3 centimeters. 
% Considering the noisiness of point cloud observation, this task requires sub-centimeter accuracy. 
% In addition, hanging a mug requires 6-DoF inference, which is heavily dependent on the posture of the grip.
In this task, even a minor error of a centimeter can result in complete failure due to noisy observation and the small size of mug handles.
In addition, the placement pose heavily depends on the posture of the grip, requiring full 6-DoF inference capability.
We also experiment with novel objects in oblique poses that were not presented during training.
% To train Diffusion-EDFs, we use two human demonstrations for each of the five training mug instances, resulting in a total of ten demonstrations.
% Therefore, this experiment demonstrates Diffusion-EDFs' ability to perform 1)~accurate 6-DoF manipulation tasks with 2)~previously unseen object instances and 3)~out-of-distribution poses.
% We use ten human demonstrations with mugs in upright poses.
% Ten human demonstrations with mugs in upright poses are used to train Diffusion-EDFs in this task.
Diffusion-EDFs successfully learned to solve this task from only ten human demonstrations, demonstrating their ability to perform 1)~accurate 6-DoF manipulation tasks with 2)~previously unseen object instances and 3)~out-of-distribution poses.

In the bowls-on-dishes task, the robot should pick up the bowls and place them on the dishes of matching colors in red-green-blue order.
Note that this sequential task requires scene-level comprehension, which is impossible for methods that rely on object segmentation.
For example, the robot should not pick up the blue bowl unless the red and green bowls are already on the dishes.
Diffusion EDFs successfully learned to solve this sequential task (in correct order) from only ten human demonstrations, which consists of red, green, and blue subtasks. 
This validates Diffusion-EDFs' ability to 1)~solve sequential problems; 2)~understand scene-level contexts; and 3)~process color-critical information.

Lastly, in the bottles-on-a-shelf task, the robot should pick up multiple bottles one by one and place them on a shelf.
In this task, we provide three identical bottle instances for both training and evaluation.
Non-probabilistic methods such as R-NDFs are known to suffer from such multimodalities in the task \citep{simeonov2023shelving}.
Methods that depend on object segmentation are also unable to solve this task, as they cannot differentiate between bottles that are already placed on the shelf and those that are not.
To evaluate generalization, we also experiment with object instances and quantities that were not presented during training.
Diffusion-EDFs successfully learned the task from four human demonstrations (consisting of three sequential pick-and-place subtasks for each bottle), showcasing their robustness to stochastic and multimodal tasks.

In conclusion, our experiments demonstrate that Diffusion-EDFs are capable of: 1) accurately generating 6-DoF poses; 2) understanding scene-level contexts; 3) learning from stochastic demonstrations; and 4) generalizing to novel object instances and poses in real-world robotic manipulation, despite being trained with a limited number of demonstrations.
We summarize the key challenges of each task in \Table~\ref{tab:challenge}. 
For the experimental results, please refer to the supplementary materials and our project website: \url{https://sites.google.com/view/diffusion-edfs/home}

% \begin{table}
%     \centering
%     \setlength\tabcolsep{2pt}%
%         \small
%         \begin{tabularx}{\linewidth}{|X|X|X|}
%         \hline
%         \textbf{Mug-on-a-hanger} & \textbf{Bowls-on-dishes} & \textbf{Bottles-on-a-shelf} \\ 
%         \hline
%         \multicolumn{1}{|l|}{\scriptsize Accurate 6-DoF inference} & \multicolumn{1}{l|}{\scriptsize Sequential problem} & \multicolumn{1}{l|}{\scriptsize Multimodal distribution} \\ 
%         \multicolumn{1}{|l|}{\scriptsize Unseen object pose} & \multicolumn{1}{l|}{\scriptsize Scene-level understanding} & \multicolumn{1}{l|}{\scriptsize Variable object number} \\ 
%         \multicolumn{1}{|l|}{\scriptsize Unseen object instance} & \multicolumn{1}{l|}{\scriptsize Color-critical} & \multicolumn{1}{l|}{\scriptsize Unseen object instance} \\ 
%         \hline
%         \end{tabularx}
%     \caption{Key challenges of each task}
%     \label{tab:challenge}
%     \vspace{-\baselineskip}
% \end{table}

\begin{table}
    \centering
    \setlength\tabcolsep{2pt}%
        \small
        \begin{tabularx}{\linewidth}{|X|X|X|}
        \hline
        \textbf{Mug-on-a-hanger} & \textbf{Bowls-on-dishes} & \textbf{Bottles-on-a-shelf} \\ 
        \hline
        {\scriptsize Accurate 6-DoF inference} & {\scriptsize Sequential problem} & {\scriptsize Multimodal distribution} \\ 
        {\scriptsize Unseen object pose} & {\scriptsize Scene-level understanding} & {\scriptsize Variable object number} \\ 
        {\scriptsize Unseen object instance} & {\scriptsize Color-critical} & {\scriptsize Unseen object instance} \\ 
        \hline
        \end{tabularx}
    \caption{Key challenges of each task}
    \label{tab:challenge}
    \vspace{-\baselineskip}
\end{table}

% \afterpage{
%     \input{sec/table_property}
% }

\section{Related Works}
\paragraph{Equivariant Robot Learning.}
% Recent works have shown that incorporating Equivariance in robot learning can significantly improve the data-efficiency, generalizability, and robustness. 
Several works in robot learning utilize $SE(2)$-equivariance to improve data-efficiency for behavior cloning \citep{zeng2020transporter,huang2022equivariant,seita2021learning,wu2022transporters,teng2022multidimensional,lim2022multi,jia2023seil} and reinforcement learning \citep{wang2022equivariant,wang2022so(2)RL,zhu2022sample,wang2022robot}. 
Although these methods can be extended to problems that are not strictly $SE(2)$-symmetric \citep{wang2022surprising,wang2023general}, they still suffer from highly spatial out-of-plane tasks \citep{ryu2023equivariant,lin2023mira}. To address this issue, $SE(3)$-equivariance has been explored in robotic manipulation learning~\citep{simeonov2022neural,simeonov2023se,ryu2023equivariant,chun2023local,kim2023robotic,huang2023edge,brehmer2023edgi,brehmer2023geometric}.
Equivariant modeling has also been shown to be effective in learning robot control \citep{kim2023robotic,zhao2023mathrm,seo2023robot,kohler2023symmetric}.

\vspace{-0.5\baselineskip}
\paragraph{SE(3)-Equivariant Graph Neural Networks.}
$SO(3)$- and $SE(3)$-equivariant graph neural networks (GNNs) \citep{thomas2018tensor,fuchs2020se,satorras2021n,deng2021vector,liao2022equiformer,du2022se,liao2023equiformerv2} are widely used to model the 3-dimensional roto-translation symmetry in various domains, including bioinformatics \citep{cramer2021alphafold2,lee2022equifold,ganea2021independent,corso2023diffdock,yim2023se}, chemistry \citep{thomas2018tensor,fuchs2020se,liao2022equiformer,batzner20223}, computer vision \citep{chatzipantazis2022se,deng2023banana,lin_ghaffari2023se,zhu_ghaffari2023e2pn,lei2023efem}, and robotics \citep{simeonov2022neural,ryu2023equivariant,huang2023edge,fu2023neuse}. 
% See Appendix~\ref{appndx:repr_theory_gnn} for more details on the subset of $SE(3)$-equivariant GNNs that we use in our architecture.
% In this paper, we use Equiformer~\citep{liao2022equiformer} for the $SE(3)$-equivariant GNN layers.

\vspace{-0.5\baselineskip}
\paragraph{Diffusion Models.}
Diffusion models are rapidly replacing previous generative models in various fields including computer vision \citep{ho2020denoising,song2020denoising,song2020score,dhariwal2021diffusion,ramesh2022hierarchical,nichol2021glide,rombach2022high},  bioinformatics \citep{yim2023se,corso2023diffdock,watson2022broadly,du2023reduce}, and robotics \citep{janner2022planning,black2023training,pearce2023imitating,chi2023diffusion,simeonov2023shelving,liu2022structdiffusion,urain2022se3dif, ajay2022conditional, brehmer2023edgi, brehmer2023geometric,mishra2023reorientdiff,chen2023planning,dai2023learning}. Recent works studied diffusion models on Riemannian manifolds \citep{bortoli2022riemannian,huang2022riemannian} such as Lie groups \citep{leach2022denoising,jagvaral2022diffusion,urain2022se3dif,corso2023diffdock,yim2023se,simeonov2023shelving}. In robotics, \citet{urain2022se3dif,simeonov2023shelving} utilized diffusion models to generate end-effector poses from $SE(3)$.
Several works also explore reward-guided diffusion policy \citep{janner2022planning,urain2022se3dif,ajay2022conditional,mishra2023reorientdiff}. 
Equivariant diffusion models on the $SE(3)$ manifold have been partially explored in bioinformatics \citep{yim2023se,corso2023diffdock} but not yet in robotics.
\section{Conclusion}
In this paper, we present Diffusion-EDFs, a bi-equivariant diffusion-based generative model on the $SE(3)$ manifold for visual robotic manipulation with point cloud observations.
% We first introduced specific bi-equivariance conditions for diffusion and denoising processes on the $SE(3)$ manifold.
% We then proposed novel multiscale EDF-based architectures for a bi-equivariant score-matching model with its validation on simulated and real robot experiments.
Diffusion-EDFs significantly improve the slow training time and small receptive field of EDFs without losing their benefits.
By thorough simulation and real hardware experiments, we validate Diffusion-EDFs' data efficiency and generalizability.
% In practice, however, the observed point clouds are often noisy and with occlusions. Future research may explore methods that work directly from images to address this issue.
One limitation of Diffusion-EDFs is the inability of control-level or trajectory-level inference. The application of geometric control framework \citep{seo2022geometric,seo2023robot} or guided diffusion with motion planning cost \citep{janner2022planning,urain2022se3dif} can be considered in subsequent work. 
The other limitation is the necessity of the grasp observation procedure, which prevents its application to closed-loop inference.
Future research may incorporate point cloud segmentation techniques to distinguish the grasp point cloud from the scene point cloud in a single observation.

\section*{Acknowledgments}
This work was supported by the National Research Foundation of Korea (NRF) grants funded by the Korea government (MSIT) (No.RS-2023-00221762 and No. 2021R1A2B5B01002620). This work was also supported by the Korea Institute of Science and Technology (KIST) intramural grants (2E31570), and a Berkeley Fellowship.

{
    \small
    \bibliographystyle{ieeenat_fullname}
    \bibliography{main}
}

% ```cvpr.sty```
% \def\maketitlesupplementary
%    {
%    \newpage
%        \twocolumn[
%         \centering
%         \Large
%         \textbf{\thetitle}\\
%         \vspace{0.5em}Supplementary Material \\
%         \vspace{1.0em}
%        ] %< twocolumn
%    }

\renewcommand\maketitlesupplementary
   {
   \newpage
       \onecolumn{
        \centering
        \Large
        \textbf{\thetitle}\\
        \vspace{0.5em}Supplementary Material \\
        \vspace{1.0em}
        }
   }

% ---------------------------------------------------------------

\clearpage
\maketitlesupplementary

\appendix
\section{Bi-equivariance}
\label{appndx:bi-equivariance}

For robust pick-and-place manipulation, the trained policy needs to be generalizable to previously unseen configurations of the target objects to pick/place.
This can be achieved by inferring end-effector poses that keep the relative pose between the grasped object and the placement target invariant.
Note that in our formulation, picking is essentially a special case of placing tasks, in which the gripper is \emph{placed} at appropriate grasp points of the target object to pick with an appropriate orientation.

\begin{figure*}[hbt!]
  \centering
  \includegraphics[width=0.9\textwidth]{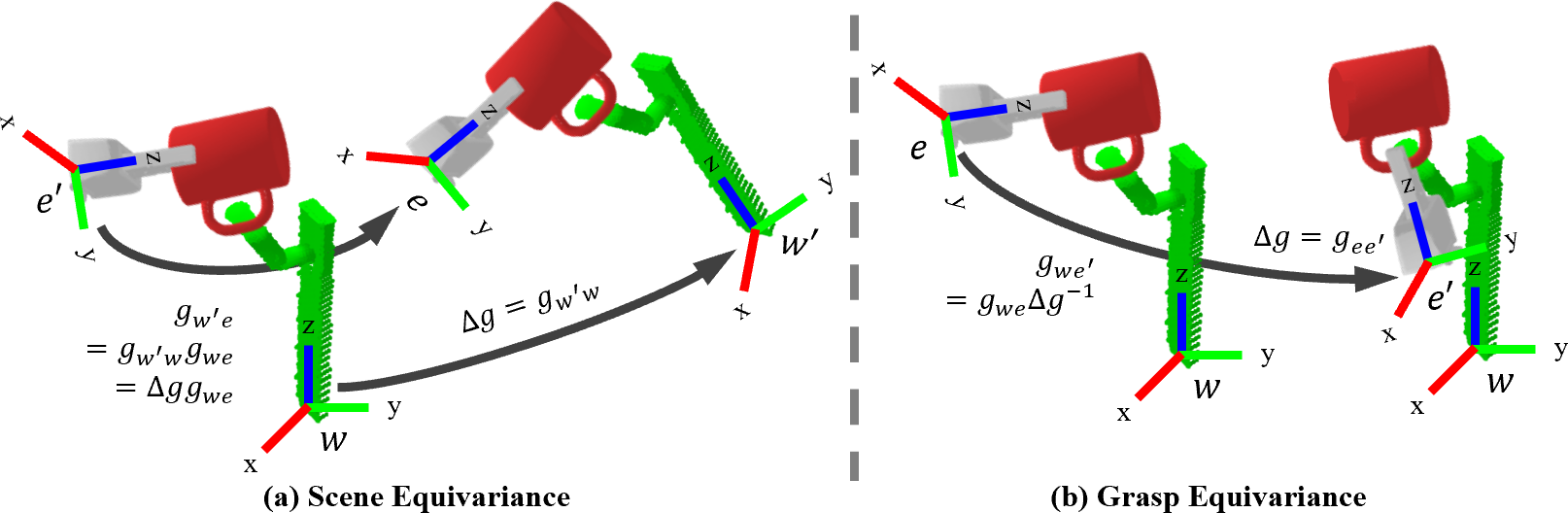}
  \caption{
    \textbf{Scene Equivariance and Grasp Equivariance.}
    \textbf{(a)} 
    % In order to keep the relative pose between the grasped object and the placement target be invariant, 
    The end-effector pose must follow the transformation of the placement target within the scene. This \textit{scene equivariance} can be achieved by multiplying the transformation $\Delta g$ on the left side of the end-effector pose. Therefore, we also refer to this property as the \textit{left equivariance}.
    \textbf{(b)} The end-effector pose must move contravariantly to the transformation of the grasped object to compensate for the changes. This \textit{grasp equivariance} involves the inverse transformation $\Delta g^{-1}$ coming to the right side of the end-effector pose. Therefore, we also refer to this property as the \textit{right equivariance}.
  }
  \label{fig:bi-equivariance}
\end{figure*}

Consider the scenario in which the policy is trained with a demonstration $\left(g_{we},\Oscene,\Ograsp\right)$ in which $g_{we}$ is the end-effector pose, and $\Oscene$ and $\Ograsp$ are respectively the point cloud observations of the scene and grasp.
We denote the world frame using subscript $w$ and the end-effector frame using subscript $e$. Note that $\Oscene$ is observed in frame $w$ and $\Ograsp$ in frame $e$. Now, let the placement target be moved by $\Dg=g_{w'w}$, inducing the transformation of the observation $\Oscene\rightarrow \Dg\,\Oscene$. This is equivalent to changing the world reference frame from $w$ to $w'$ with respect to the observation. Therefore, the end-effector pose should also be transformed equivariantly as $g_{we}\rightarrow g_{w'e}=\Dg\,g_{we}$ (see \Fig~\ref{fig:bi-equivariance}-(a)). This \emph{scene equivariance} is also referred to as \emph{left equivariance} \citep{ryu2023equivariant,kim2023robotic}, as the transformation $\Dg$ comes to the left side of $g_{we}$.

On the other hand, consider the transformation of the grasped object $\Dg=g_{e'e}$, which induces the transformation of the observation $\Ograsp\rightarrow \Dg\,\Ograsp$. This is equivalent to changing the end-effector reference frame from $e$ to $e'$ with respect to the observation. In the world frame, this corresponds to the transformation of the end-effector pose by $g_{we}\rightarrow g_{we'}=g_{we}\Dg^{-1}$  (see \Fig~\ref{fig:bi-equivariance}-(b)). This \emph{grasp equivariance} is also referred to as \emph{right equivariance} \citep{ryu2023equivariant,kim2023robotic}, as the transformation $\Dg^{-1}$ comes to the right side of $g_{we}$.
Combining these left and right equivariance conditions, we obtain the bi-equivariance condition, which can be formally expressed in a probabilistic form as~\eqref{eqn:bi-equiv}.

\section{Analytic Form of the Target Score in~\eqref{eqn:score_matching_loss}}
\label{appndx:analytic-target-score}
In this section, we provide the analytic form of the target score function in~\eqref{eqn:score_matching_loss}
\begin{equation}
    \label{eqn:target-score-to-derive}
    \Grad \log \BrownianKernel_{t}(\gref^{-1}g_0^{-1}g\gref)
\end{equation}
By definition, the $i$-th component of the target score function is calculated as follows:
\begin{equation}
    \begin{split}
   \Lie_i\log \BrownianKernel_{t}(\gref^{-1}g_0^{-1}g\gref)&=\left.\frac{d}{d\epsilon}\right|_{\epsilon=0}\log{\BrownianKernel_{t}\left(\gref^{-1}g_0^{-1}g\exp[\epsilon\basis_i]\gref\right)} \\
    &=\left.\frac{d}{d\epsilon}\right|_{\epsilon=0}\log{\BrownianKernel_{t}\left(\gref^{-1}g_0^{-1}g\, \gref\exp[\epsilon Ad_{\gref^{-1}}\basis_i]\right)} \\
    &=\left[\Lie_{Ad_{\gref^{-1}}\basis_i}\log \BrownianKernel_{t}\right]\left(\gref^{-1}g_0^{-1}g\gref\right) \\
    &=\left[\sum_{j=1}^{6}\left[Ad_{\gref^{-1}}\right]_{ji}\Lie_{j}\log \BrownianKernel_{t}\right]\left(\gref^{-1}g_0^{-1}g\gref\right) \\
    &=\sum_{j=1}^{6}\left[Ad_{\gref^{-1}}\right]_{ji}\left[\Lie_{j}
    \log \BrownianKernel_{t}\vphantom{1^{1^{1^{1^{1}}}}}           % vphantom to adjust the height of the square bracket
    \right]\left(\gref^{-1}g_0^{-1}g\gref\right)
    \end{split}
\end{equation}
\begin{equation}
    \Rightarrow\Grad \log \BrownianKernel_{t}(\gref^{-1}g_0^{-1}g\gref)=\left[Ad_{\gref}\right]^{-T} \left[\Grad \log \BrownianKernel_{t}\right](\gref^{-1}g_0^{-1}g\gref)
\end{equation}
Therefore, all we need is the score of the Brownian diffusion kernel $\Grad \log \BrownianKernel_{t}(g)$ which can be decomposed into its translation and rotation parts using \eqref{eqn:iso_diff_kernel}
\begin{equation}
    \Grad \log \BrownianKernel_{t}(g)=\Grad\log\mathcal{N}(\vp; \vmu=\boldsymbol{0}, \Sigma=tI)
    + \Grad \log \mathcal{IG}_{SO(3)}(R; \eps=t/2)
\end{equation}
where $\Grad\log\mathcal{N}(\vp; \vmu=\boldsymbol{0}, \Sigma=tI)=-\vp/t$ can be easily computed.
A common practice for the calculation of the rotational part $\Grad \log \mathcal{IG}_{SO(3)}(R; \eps=t/2)$ is to use automatic differentiation packages \citep{leach2022denoising,ryu2023equivariant,jagvaral2022diffusion,urain2022se3dif,corso2023diffdock,yim2023se}. 
However, the explicit form can be easily calculated without automatic differentiation packages. 
\begin{equation}
    \Lie_i\ \log\mathcal{IG}_{SO(3)}(R; \eps) = \frac{\Lie_i\ \mathcal{IG}_{SO(3)}(R; \eps)}{\mathcal{IG}_{SO(3)}(R; \eps)}
\end{equation}
\begin{equation}
    \begin{split}
        \Lie_i\ \mathcal{IG}_{SO(3)}(R; \eps)
        =\sum_{l=0}^\infty(2l+1)\exp\left[-l(l+1)\epsilon\right]\left[\frac{(l+1)\sin(l\theta)-l\sin((l+1)\theta)}{\cos(\theta)-1}\right]
        \left[\frac{-\mathrm{tr}\left[R[\basis_i]^{\wedge}\right]}{2\sin{\theta}}\right] \label{eqn:explicit_lie_igso3}
    \end{split}
\end{equation}
We denote the skew-symmetric matrix of the $i$-th $\mathfrak{so}(3)$ basis $\basis_i$ as $[\basis_i]^{\wedge}$, whose matrix element is $[\basis_i]^{\wedge}_{jk}=-\epsilon_{ijk}$ where $\epsilon_{ijk}$ is the Levi-Civita symbol.

The derivation is as follows.
First, we rewrite \eqref{eqn:igso3} with the \emph{character} $\mathcal{X}(R)$ of $SO(3)$ \citep{zee2016group}.
\begin{align}
    \mathcal{IG}_{SO(3)}(R; \eps)&=\sum_{l=0}^\infty(2l+1)\exp\left[-l(l+1)\epsilon\right]\mathcal{X}_l(R)\\
    % \mathcal{X}_l(R)=\mathrm{tr}\left[D_l(R)\right]=\frac{\sin\left((2l+1)\frac{\theta}{2}\right)}{\sin(\frac{\theta}{2})}&=\frac{\cos((l+1)\theta)-\cos(l\theta)}{\cos(\theta)-1}
    \mathcal{X}_l(R)=\mathrm{tr}\left[D_l(R)\right]&=
    \sin\left((2l+1)\frac{\theta}{2}\right)/\sin(\frac{\theta}{2})
\end{align}
$\theta\in(0,\pi)$ is the rotation angle of $R$. Now we calculate the Lie derivative of $\mathcal{IG}_{SO(3)}$ as follows:
\begin{align}
    \Lie_i\,\mathcal{IG}_{SO(3)}(R; \eps)&=\sum_{l=0}^\infty(2l+1)\exp\left[-l(l+1)\epsilon\right]\Lie_i\,\mathcal{X}_l(R) \\
    \Lie_i\,\mathcal{X}_l(R) &= \left[\frac{(l+1)\sin(l\theta)-l\sin((l+1)\theta)}{\cos(\theta)-1}\right]\Lie_i\,\theta \\
    \Lie_i\,\theta &= \left[\frac{-1}{\sin{\theta}}\right] \Lie_i\left[\cos{\theta}\right] \label{eqn:lie-of-theta}
\end{align}
The last line can be easily calculated using $\cos{\theta}=\frac{1}{2}\left(\mathrm{tr}\left[R\right]-1\right)$ and $\Lie_{\mathcal{V}}\,\mathrm{tr}\left[R\right]=\mathrm{tr}\left[R[\mathcal{V}]^{\wedge}\right]$.
\begin{equation}
    \Lie_i\left[\cos{\theta}\right]=\frac{1}{2}\left(\mathrm{tr}\left[R[\basis_i]^{\wedge}\right]\right)
\end{equation}
Combining these results, one can derive \eqref{eqn:explicit_lie_igso3}. In practice, the infinite sum in \eqref{eqn:explicit_lie_igso3} is approximated with $\sum_{l=0}^{l_{max}}$ where $l_{max}=1000\sim10000$, which can be computed within a millisecond when appropriately parallelized. Although we have derived \eqref{eqn:explicit_lie_igso3} for $\theta=(0,\pi)$, the result can be asymptotically extended to $\theta=0$ and $\pi$ as $\mathcal{IG}_{SO(3)}$ is an infinitely differentiable on $SO(3)$ \citep{nikolayev1970normal}.
\section{Proofs and Derivations}
\label{appndx:proofs}

\subsection{Proof of Proposition~\ref{propo:score_left_and_right}}
\label{proof:score_left_and_right}
% Let $\Lie_i$ be the Lie derivative along the $i$-th basis $\basis_i$ of the $\mathfrak{se}(3)$ Lie algebra.
% \begin{equation}
%     \vs(g|\Oscene,\Ograsp) = \Grad \log{P(g|\Oscene,\Ograsp)}
% \end{equation}

\begin{proof}[Proof of the left invariance of the score function]
    \begin{align*}
    \Lie_i \log{P(\Dg\,g|\Dg\cdot\Oscene,\Ograsp)}
    &=\left.\frac{d}{d\epsilon}\right|_{\epsilon=0}\log{P}(\Dg\,g\exp{\left[\epsilon \basis_i\right]}|\Dg\cdot\Oscene,\Ograsp)\\
    &=\left.\frac{d}{d\epsilon}\right|_{\epsilon=0}\log{P}(g\exp{\left[\epsilon \basis_i\right]}|\Oscene,\Ograsp)\\
    &=\Lie_i \log{P}(g|\Oscene,\Ograsp)
    \end{align*}
    where we used $P(\Dg\,g|\Dg\cdot\Oscene,\Ograsp)=P(g|\Oscene,\Ograsp)$ in the second line.
\end{proof}

\begin{proof}[Proof of the right equivariance of the score function]
    \begin{align*}
    \Lie_{i}\log{P}(g\,\Dg^{-1}|\Oscene,\Dg\cdot\Ograsp)
    &=\left.\frac{d}{d\epsilon}\right|_{\epsilon=0}\log{P}(g\,\Dg^{-1}\exp{\left[\epsilon \basis_i\right]}|\Oscene,\Dg\cdot\Ograsp)\\
    &=\left.\frac{d}{d\epsilon}\right|_{\epsilon=0}\log{P}(g\,\Dg^{-1}\exp{\left[\epsilon \basis_i\right]}\Dg|\Oscene,\Ograsp)\\
    % &=\left.\frac{d}{d\epsilon}\right|_{\epsilon=0}\log{P}(g\exp{\left[\epsilon \Dg^{-1}\basis_i \Dg\right]}|\Oscene,\Ograsp)\\
    &=\left.\frac{d}{d\epsilon}\right|_{\epsilon=0}\log{P}(g\exp{\left[\epsilon Ad_{\Dg^{-1}}\basis_i\right]}|\Oscene,\Ograsp)\\
    &=\Lie_{Ad_{\Dg^{-1}}\basis_i}\log{P}(g|\Oscene,\Ograsp)\\
    &=\Lie_{\sum_{j}{\left[Ad_{\Dg^{-1}}\right]_{ji}\basis_{j}}}\log{P}(g|\Oscene,\Ograsp)\\
    &=\sum_{j=1}^{6}{\left[Ad_{\Dg^{-1}}\right]_{ji}\Lie_{j}\log{P}(g|\Oscene,\Ograsp)} \tag{$\because$ Linearity of Lie-derivatives~\citep{chirik}\ \ $\mathcal{L}_{\sum_{i}v_i\basis_{i}}=\sum_{i}v_i\,\mathcal{L}_{i}$}\\
    \Rightarrow
    \Grad \log{P}(g\,\Dg^{-1}|\Oscene,\Dg\cdot\Ograsp)
    &=\left[Ad_{\Dg^{-1}}\right]^T \Grad \log{P}(g|\Oscene,\Ograsp)
    =\left[Ad_{\Dg}\right]^{-T} \Grad \log{P}(g|\Oscene,\Ograsp)
    \end{align*}
    where we denote the $(j,i)$-th matrix element of $Ad_{\Dg^{-1}}$ with $\left[Ad_{\Dg^{-1}}\right]_{ji}\,$. We used $P(g\,\Dg^{-1}|\Oscene,\Dg\cdot\Ograsp)=P(g|\Oscene,\Ograsp)$ in the second line.
\end{proof}

\subsection{Proof of Proposition~\ref{propo:marginal_biequiv}}
\label{proof:marginal_biequiv}
It is straightforward to prove the bi-equivariance of the diffused marginal using the bi-invariance of the integral measure (Haar measure) $dg$
\begin{equation}
    \label{eqn:integral-measure-bi-invariance}
    \int_{SE(3)}d(\Dg\,g)=\int_{SE(3)}dg=\int_{SE(3)}d(g\,\Dg)   \qquad\forall \Dg\in SE(3)
\end{equation}
where $dg=dR\,d\vp=\frac{1}{8\pi^2}\left(\sin{\beta}\right)\,d\alpha\,d\beta\,d\gamma\,dx\,dy\,dz$ in the rotation-translation coordinate with the Euler angles $\alpha,\beta,\gamma$ and the frame origin $x,y,z$. See \citet{chirikjian2015partial, chirik, chiriknonharm,murray2017mathematical} and Appendix~A of \citet{ryu2023equivariant} for more details on the bi-invariant integral measure of $SE(3)$.

We first prove that the marginal is bi-equivariant if the kernel is bi-equivariant.
\begin{proof}[Proof of left equivariance]
    \begin{align*}
        P_t(g|\Oscene,\Ograsp)&=\int_{SE(3)}dg_0\;P_{t|0}(g|g_0,\Oscene,\Ograsp)P_0(g_0|\Oscene,\Ograsp)
        \\&=\int_{SE(3)}dg_0\;P_{t|0}(g|g_0,\Oscene,\Ograsp)P_0(\Dg\,g_0|\Dg\cdot\Oscene,\Ograsp)    \tag{$\because$~\eqref{eqn:bi-equiv}}
        \\&=\int_{SE(3)}dg_0\;P_{t|0}(\Dg\,g|\Dg\,g_0,\Dg\cdot\Oscene,\Ograsp)P_0(\Dg\,g_0|\Dg\cdot\Oscene,\Ograsp)    \tag{$\because$~\eqref{eqn:bi-equiv-kernel}}
        \\&=\int_{SE(3)}dg_0\;P_{t|0}(\Dg\,g|g_0,\Dg\cdot\Oscene,\Ograsp)P_0(g_0|\Dg\cdot\Oscene,\Ograsp)    \tag{$\because$~\eqref{eqn:integral-measure-bi-invariance},\, $\Dg\,g_0\rightarrow g_0$}
        \\&=P_t(\Dg\,g|\Dg\cdot\Oscene,\Ograsp)
    \end{align*}
\end{proof}
\begin{proof}[Proof of right equivariance]
    \begin{align*}
        P_t(g|\Oscene,\Ograsp)&=\int_{SE(3)}dg_0\;P_{t|0}(g|g_0,\Oscene,\Ograsp)P_0(g_0|\Oscene,\Ograsp)
        \\&=\int_{SE(3)}dg_0\;P_{t|0}(g|g_0,\Oscene,\Ograsp)P_0(g_0\,\Dg^{-1}|\Oscene,\Dg\cdot\Ograsp) \tag{$\because$~\eqref{eqn:bi-equiv}}
        \\&=\int_{SE(3)}dg_0\;P_{t|0}(g\,\Dg^{-1}|g_0\,\Dg^{-1},\Oscene,\Dg\cdot\Ograsp)P_0(g_0\,\Dg^{-1}|\Oscene,\Dg\cdot\Ograsp) \tag{$\because$~\eqref{eqn:bi-equiv-kernel}}
        \\&=\int_{SE(3)}dg_0\;P_{t|0}(g\,\Dg^{-1}|g_0,\Oscene,\Dg\cdot\Ograsp)P_0(g_0|\Oscene,\Dg\cdot\Ograsp) \tag{$\because$~\eqref{eqn:integral-measure-bi-invariance},\, $g_0\,\Dg^{-1}\rightarrow g_0$}
        \\&=P_t(g\,\Dg^{-1}|\Oscene,\Dg\cdot\Ograsp)
    \end{align*}
\end{proof}
Similarly, it can be proven that the kernel must be bi-equivariant (up to measure zero) to guarantee the bi-equivariance of the diffused marginal for any arbitrary initial distribution $dP_0=P_0\,dg_0$. 
% It suffices to show only for every possible initial distributions $dP_0=P_0\,dg_0$ that are nonzero everywhere. 
\begin{proof}[Proof]
    \begin{align*}
        P_t(g|\Oscene,\Ograsp)&=\int_{SE(3)}dg_0\;P_{t|0}(g|g_0,\Oscene,\Ograsp)P_0(g_0|\Oscene,\Ograsp)
        \\
        P_t(\Dg\,g|\Dg\cdot\Oscene,\Ograsp)
        &=\int_{SE(3)}dg_0\;P_{t|0}(\Dg\,g|g_0,\Dg\cdot\Oscene,\Ograsp)P_0(g_0|\Dg\cdot\Oscene,\Ograsp)
        \\
        &=\int_{SE(3)}dg_0\;P_{t|0}(\Dg\,g|g_0,\Dg\cdot\Oscene,\Ograsp)P_0(\Dg^{-1}\,g_0|\Oscene,\Ograsp)
        \tag{$\because$~\eqref{eqn:bi-equiv}}
        \\
        &=\int_{SE(3)}dg_0\;P_{t|0}(\Dg\,g|\Dg\,g_0,\Dg\cdot\Oscene,\Ograsp)P_0(g_0|\Oscene,\Ograsp)
        \tag{$\because$~\eqref{eqn:integral-measure-bi-invariance},\, $g_0\rightarrow \Dg\,g_0$}
    \end{align*}
    \begin{align*}
        P_t(g\,\Dg^{-1}|\Oscene,\Dg\cdot\Ograsp)
        &=\int_{SE(3)}dg_0\;P_{t|0}(g\,\Dg^{-1}|g_0,\Oscene,\Dg\cdot\Ograsp)P_0(g_0|\Oscene,\Dg\cdot\Ograsp)
        \\
        &=\int_{SE(3)}dg_0\;P_{t|0}(g\,\Dg^{-1}|g_0,\Oscene,\Dg\cdot\Ograsp)P_0(g_0\,\Dg|\Oscene,\Ograsp)
        \tag{$\because$~\eqref{eqn:bi-equiv}}
        \\
        &=\int_{SE(3)}dg_0\;P_{t|0}(g\,\Dg^{-1}|g_0\,\Dg^{-1},\Oscene,\Dg\cdot\Ograsp)P_0(g_0|\Oscene,\Ograsp)
        \tag{$\because$~\eqref{eqn:integral-measure-bi-invariance},\, $g_0\rightarrow g_0\,\Dg^{-1}$}
    \end{align*}
    \begin{align*}
        \Rightarrow\ &\int_{SE(3)} dg_0\;P_0(g_0|\Oscene,\Ograsp)\times\left[P_{t|0}(g|g_0,\Oscene,\Ograsp)-P_{t|0}(\Dg\,g|\Dg\,g_0,\Dg\cdot\Oscene,\Ograsp)\right] = 0 \\
        &\int_{SE(3)} dg_0\;P_0(g_0|\Oscene,\Ograsp)\times\left[P_{t|0}(g|g_0,\Oscene,\Ograsp)-P_{t|0}(g\,\Dg^{-1}|g_0\,\Dg^{-1},\Oscene,\Dg\cdot\Ograsp)\right] = 0
    \end{align*}
    Therefore, for this equation to hold for any arbitrary bi-equivariant initial distribution $dP_0=P_0\,dg_0$, the diffusion kernel must be bi-equivariant $\forall g,\Dg\in SE(3)$
    \begin{align}
        P_{t|0}(g|g_0,\Oscene,\Ograsp)-P_{t|0}(\Dg\,g|\Dg\,g_0,\Dg\cdot\Oscene,\Ograsp) &= 0 \\
        P_{t|0}(g|g_0,\Oscene,\Ograsp)-P_{t|0}(g\,\Dg^{-1}|g_0\,\Dg^{-1},\Oscene,\Dg\cdot\Ograsp) &= 0 \\
        \Rightarrow\ P_{t|0}(g|g_0,\Oscene,\Ograsp)=P_{t|0}(\Dg\,g|\Dg\,g_0,\Dg\cdot\Oscene,\Ograsp)
        &=P_{t|0}(g\,\Dg^{-1}|g_0\,\Dg^{-1},\Oscene,\Dg\cdot\Ograsp)
    \end{align}
\end{proof}

% \subsection{Proof of Proposition~\ref{propo:no-naive-diffusion}}
% \label{proof:no-naive-diffusion}

\subsection{Non-existence of Bi-Invariant Diffusion Kernels on SE(3)}
\label{proof:non-existence-of-bi-invariant-kernel}
Note that any left invariant kernel $P_{t|0}(g|g_0)$ can be written in a univariate form $\LinvKernel_{t}(g_0^{-1}g)$.
% Due to the left invariance of the kernel, it must depend solely on the group difference $g_0^{-1}g$
\begin{equation}
    P_{t|0}(\Dg\,g|\Dg\,g_0)=P_{t|0}(g|g_0)\quad\forall\,\Dg,g\quad\Rightarrow\quad P_{t|0}(g|g_0)=P_{t|0}(g_0^{-1}g|I)\quad\forall\,g
\end{equation}
% Therefore, we can simply denote any left-invariant kernel as $\LinvKernel_t(g)=P_{t|0}(g|I)$.
The right invariance requires this kernel to satisfy $\LinvKernel_t(\Dg\,g\,\Dg^{-1})=\LinvKernel_t(g)$, meaning that it is a \emph{class function}, which does not exist for $L^2(SE(3))$ \citep{kyatkin1998regularized,chiriknonharm}.
% For the case $\Delta R=I$, this means that
% \begin{equation*}
%     \BrownianKernel_{t}\left(
%     \begin{bmatrix}
%         R& -R\Delta\vp+R\vp+\Delta\vp \\
%         0& 1
%     \end{bmatrix}
%     \right)=\BrownianKernel_{t}\left(
%     \begin{bmatrix}
%         R& \vp \\
%         0& 1
%     \end{bmatrix}
%     \right)
% \end{equation*}
% Note that $-R\Delta\vp+R\vp+\Delta\vp$ can reach any point on $\mathbb{R}^3$ by adjusting $\mathbb{R}^3$ (NOT TRUE)

\subsection{Proof of Proposition~\ref{propo:bi-equiv-of-diffusion-marginal-kernel}}
\label{proof:frame_selection_marginal}
\begin{proof}
The right equivariance can be proved as follows.
    \begin{align*}
        P_{t|0}(g|g_0,\Oscene,\Ograsp)&=\int_{SE(3)} d\gref\FrameSelection(\gref|g_0^{-1}\cdot\Oscene,\Ograsp)\LinvKernel_{t}(\gref^{-1}g_0^{-1}g\gref)
        \\&=\int_{SE(3)} d\gref\FrameSelection(\Dg\,\gref|(\Dg\,g_0^{-1})\cdot\Oscene,\Dg\cdot\Ograsp)\LinvKernel_{t}(\gref^{-1}g_0^{-1}g\gref)\tag{$\because $~\eqref{eqn:frame_selection}}
        \\&=\int_{SE(3)} d\gref\FrameSelection(\gref|(\Dg\,g_0^{-1})\cdot\Oscene,\Dg\cdot\Ograsp)\LinvKernel_{t}(\gref^{-1}\left(g_0\,\Dg^{-1}\right)^{-1}\left(g\,\Dg^{-1}\right)\gref)\tag{$\because$ invariance of integral $\int d\gref$ under $\gref\rightarrow\Dg^{-1}\,\gref$}
        \\&=P_{t|0}(g\,\Dg^{-1}|g_0\,\Dg^{-1},\Oscene,\Dg\cdot\Ograsp)
    \end{align*}
    The left equivariance proof is straightforward using the following equations:
    \begin{align}
        g_0^{-1}\,g&=\left(\Dg\,g_0\right)^{-1}\left(\Dg\,g\right)
        \\g_0^{-1}\cdot\Oscene&=\left(\Dg\,g_0\right)^{-1}\cdot\left(\Dg\cdot\Oscene\right)
    \end{align}
\end{proof}

\subsection{Proof of Proposition~\ref{propo:only-translation}}
\label{proof:only-translation}
Note that the Brownian diffusion kernel $\BrownianKernel_{t}(g)$ is right-invariant to  rotation, that is,
\begin{equation}
    \label{eqn:right-rot-inv-of-kt}
    \begin{split}
        \BrownianKernel_{t}\left((g_0\,\Delta R)^{-1}\,(g\,\Delta R)\right)&=\BrownianKernel_{t}(g_0^{-1}\,g) \\
        \Rightarrow\BrownianKernel_{t}(\Delta R^{-1}\,g\,\Delta R)&=\BrownianKernel_{t}(g) \qquad\forall \Delta R\in SO(3)
    \end{split}
\end{equation}
where we abuse the notation to denote the action of a pure rotation $\Delta R$ on $g=(\vp,R)$ as $\Delta R\,g=(\Delta R\,\vp, \Delta R\,R)$ and $g\,\Delta R=(\vp, R\,\Delta R)$. \eqref{eqn:right-rot-inv-of-kt} holds because the Gaussian distribution in \eqref{eqn:iso_diff_kernel} is rotation-invariant and $\mathcal{IG}_{SO(3)}$ in \eqref{eqn:igso3} is a linear combination of \emph{characters} of $SO(3)$, which are \emph{class functions} due to the permutation invariance of trace operations (see \Supp~\ref{appndx:analytic-target-score} and ~\ref{proof:non-existence-of-bi-invariant-kernel}).
Consider the following diffusion kernel with the equivariant origin selection mechanism in \eqref{eqn:origin_selection}:
\begin{equation}
    P_{t|0}(g|g_0,\Oscene,\Ograsp)=\int_{\mathbb{R}^3} d\pref\OriginSelection\left(\pref|g_0^{-1}\cdot \Oscene,\Ograsp\right)\BrownianKernel_{t}\left(\left(g_0\triangleleft\pref\right)^{-1}\left(g\triangleleft\pref\right)\right)
\end{equation}
where $ \triangleleft\,\pref:SE(3)\rightarrow SE(3)$ denotes the right action of a pure translation $\pref\in\mathbb{R}^3$ onto $g=(\vp,R)\in SE(3)$ such that
\begin{equation}
    g\,\triangleleft\,\pref = 
    \begin{bmatrix}
        R & \vp \\ \emptyset & 1
    \end{bmatrix}
    \begin{bmatrix}
        I & \pref \\ \emptyset & 1
    \end{bmatrix} = 
    \begin{bmatrix}
        R & R\,\pref + \vp \\ \emptyset & 1
    \end{bmatrix}
\end{equation}
Note that the following equation holds for all $g_1,g_2\in SE(3)$ and $\pref\in\mathbb{R}^{3}$:
\begin{equation}
    \label{eqn:split-formula}
    \begin{split}
        (g_1\,g_2)\,\triangleleft\,\pref
        &= 
        \begin{bmatrix}
            R_1 & \vp_1 \\ \emptyset & 1
        \end{bmatrix}
        \begin{bmatrix}
            R_2 & \vp_2 \\ \emptyset & 1
        \end{bmatrix}
        \begin{bmatrix}
            I & \pref \\ \emptyset & 1
        \end{bmatrix} 
        \\&= 
        \begin{bmatrix}
            R_1\,R_2 & R_1\left(R_2\pref+\vp_2\right) + \vp_1 \\ \emptyset & 1
        \end{bmatrix}
        \\&= 
        \begin{bmatrix}
            R_1 & \vp_1 \\ \emptyset & 1
        \end{bmatrix}
        \begin{bmatrix}
            I & g_2\,\pref \\ \emptyset & 1
        \end{bmatrix}
        \begin{bmatrix}
            R_2 & \mathbf{0} \\ \emptyset & 1
        \end{bmatrix}
        \\&= \left(g_1 \triangleleft \left(g_2\,\pref\right)\right)\,\Delta R_2
    \end{split}
\end{equation}

The bi-equivariance of $P_{t|0}$ can be proved using \eqref{eqn:right-rot-inv-of-kt} and \eqref{eqn:split-formula}.
\begin{proof} 
The proof of left equivariance is straightforward as $g_0^{-1}\cdot\Oscene=\left(\Dg\,g_0\right)^{-1}\cdot\left(\Dg\cdot\Oscene\right)$.
The proof of right equivariance is as follows:
\begin{align*}
    &P_{t|0}(g\,\Dg^{-1}|g_0\,\Dg^{-1},\Oscene,\Dg\cdot\Ograsp)
    \\&=
    \int_{\mathbb{R}^3} d\pref
    \OriginSelection\left(
        \pref | (\Delta g\,g_0^{-1})\cdot \Oscene,\Delta g\cdot\Ograsp\right
    )
    \BrownianKernel_{t}\left(
        \left((g_0\,\Dg^{-1})\triangleleft\pref\right)^{-1}    \left((g\,\Dg^{-1})\triangleleft\pref\right)
    \right)
    \\&=
    \int_{\mathbb{R}^3} d\pref
    \OriginSelection\left(
        \Dg^{-1}\,\pref | g_0^{-1}\cdot \Oscene,\Ograsp\right
    )
    \BrownianKernel_{t}\left(
        \left((g_0\,\Dg^{-1})\triangleleft\pref\right)^{-1}    \left((g\,\Dg^{-1})\triangleleft\pref\right)
    \right)
    \tag{$\because $\ \eqref{eqn:origin_selection}}
    \\&=
    \int_{\mathbb{R}^3} d\pref
    \OriginSelection\left(
        \Dg^{-1}\,\pref | g_0^{-1}\cdot \Oscene,\Ograsp\right
    )
    \BrownianKernel_{t}\left(
        \Delta R\,\left(               g_0 \triangleleft \left(\Dg^{-1}\,\pref\right)                 \right)^{-1}    
                  \left(                 g \triangleleft \left(\Dg^{-1}\,\pref\right)                 \right)\,\Delta R^{-1}
    \right)
    \tag{$\because $\ \eqref{eqn:split-formula}}
    \\&=
    \int_{\mathbb{R}^3} d\pref
    \OriginSelection\left(
        \Dg^{-1}\,\pref | g_0^{-1}\cdot \Oscene,\Ograsp\right
    )
    \BrownianKernel_{t}\left(
        \left(               g_0 \triangleleft \left(\Dg^{-1}\,\pref\right)                 \right)^{-1}    
        \left(                 g \triangleleft \left(\Dg^{-1}\,\pref\right)                 \right)
    \right)
    \tag{$\because $\ \eqref{eqn:right-rot-inv-of-kt}}
    \\&=
    \int_{\mathbb{R}^3} d\pref
    \OriginSelection\left(
        \pref | g_0^{-1}\cdot \Oscene,\Ograsp\right
    )
    \BrownianKernel_{t}\left(
        \left(               g_0 \triangleleft \left(\pref\right)                 \right)^{-1}    
        \left(                 g \triangleleft \left(\pref\right)                 \right)
    \right)
    \tag{$\because $\ invariance of Euclidean integral under roto-translation, $\pref\rightarrow\Dg\,\pref$}
    \\&=
    P_{t|0}(g|g_0,\Oscene,\Ograsp)
\end{align*}
% where in the second last line, we used the fact that for $g=(\vp,R)$ and $\Dg=(\Delta\vp,\Delta R)$
% \begin{align*}
%     \left(g\triangleleft \Dg\,\pref\right)\Delta R &= 
%     \begin{bmatrix}
%         R & \vp \\ \emptyset & 1
%     \end{bmatrix}
%     \begin{bmatrix}
%         I & \Delta R\,\pref + \Delta\vp \\ \emptyset & 1
%     \end{bmatrix}
%     \begin{bmatrix}
%         \Delta R & \emptyset \\ \emptyset & 1
%     \end{bmatrix}
%     \\&=
%     \begin{bmatrix}
%         R\,\Delta R & R\,\Delta R\,\pref+R\,\Delta\vp+\vp \\ \emptyset & 1
%     \end{bmatrix}
%     \\&=
%     \begin{bmatrix}
%         R\,\Delta R & R\,\Delta\vp+\vp \\ \emptyset & 1
%     \end{bmatrix}
%     \begin{bmatrix}
%         I & \pref \\ \emptyset & 1
%     \end{bmatrix}
%     \\&= g\,\Dg\triangleleft\pref
% \end{align*}
\end{proof}
In fact, any left-invariant kernel that is also right-invariant to rotation as in~\eqref{eqn:right-rot-inv-of-kt} can be used.

\subsection{Derivation of \eqref{eqn:mse-minimizer}}
\label{proof:mse-minimizer}
We first show that $\vs_t^*(g|\Oscene,\Ograsp) = \mathbb{E}_{g_0,\gref|g,\Oscene,\Ograsp}\left[\Grad \log \LinvKernel_{t}(\gref^{-1}g_0^{-1}g\gref)\right]$ using a simple variational calculus. 

\begin{proof}
Let $\delta \vs_t(g|\Oscene,\Ograsp)$ be a perturbation of the score model $\vs_t(g|\Oscene,\Ograsp)$. For the optimal score model $\vs^*_t(g|\Oscene,\Ograsp)$, any small perturbation would result in zero perturbation of the objective.
\begin{equation}
    \begin{split}
         &\vs_t^*(g|\Oscene,\Ograsp)\ \ =\  \argmin_{\vs_t(g|\Oscene,\Ograsp)} \mathcal{J}_t\left[\vs_t(g|\Oscene,\Ograsp) \right] \\
        \Rightarrow \ \  &\delta\mathcal{J}_t\left[\vs^*_t(g|\Oscene,\Ograsp) \right] \ = \ 0\qquad \forall\ \  \delta \vs^*_t(g|\Oscene,\Ograsp)
    \end{split}
\end{equation}
The explicit form of the perturbation of the objective with regard to $\delta \vs_t(g|\Oscene,\Ograsp)$ is written as follows:
\begin{equation}
    \begin{split}
    &\delta\mathcal{J}_t\left[\vs_t(g|\Oscene,\Ograsp) \right] = 
    \delta\left(\mathbb{E}_{g, g_0, \gref, \Oscene,\Ograsp}\left[\frac{1}{2} \left \|\vs_t(g|\Oscene,\Ograsp)-\Grad \log \LinvKernel_{t}(\gref^{-1}g_0^{-1}g\gref) \right \|^2\right]\right)\\
    =\ \ &
    \mathbb{E}_{g, \Oscene,\Ograsp}\left[
    \delta \vs_t(g|\Oscene,\Ograsp)\cdot\left[\vs_t(g|\Oscene,\Ograsp)-\mathbb{E}_{g_0,\gref|g,\Oscene,\Ograsp}\left[\Grad \log \LinvKernel_{t}(\gref^{-1}g_0^{-1}g\gref)\right]\right]
    \right]
    \end{split}
\end{equation}
Therefore, assuming $P_t(g|\Oscene,\Ograsp) > 0\ \  \forall g, \Oscene,\Ograsp$, the optimal score model must be
\begin{equation}
    \vs^*_t(g|\Oscene,\Ograsp)=\mathbb{E}_{g_0,\gref|g,\Oscene,\Ograsp}\left[\Grad \log \LinvKernel_{t}(\gref^{-1}g_0^{-1}g\gref)\right]
\end{equation}
\end{proof}

We now show that $\mathbb{E}_{g_0,\gref|g,\Oscene,\Ograsp}\left[\Grad \log \LinvKernel_{t}(\gref^{-1}g_0^{-1}g\gref)\right] = \Grad \log P_t(g|\Oscene,\Ograsp)$.

\begin{proof}
    \begin{align*}
    &\mathbb{E}_{g_0,\gref|g,\Oscene,\Ograsp}\left[\Grad \log \LinvKernel_{t}(\gref^{-1}g_0^{-1}g\gref)\right]\\
    &=\int dg_0\int d\gref\ P(g_0,\gref|g,\Oscene,\Ograsp;t)\frac{\Grad \LinvKernel_{t}(\gref^{-1}g_0^{-1}g\gref)}{\LinvKernel_{t}(\gref^{-1}g_0^{-1}g\gref)}\\
    &=\int dg_0\int d\gref\ \left[P(g|g_0,\gref,\Oscene,\Ograsp;t)\frac{P(g_0,\gref|\Oscene,\Ograsp)}{P_t(g|\Oscene,\Ograsp)}\right]\frac{\Grad \LinvKernel_{t}(\gref^{-1}g_0^{-1}g\gref)}{\LinvKernel_{t}(\gref^{-1}g_0^{-1}g\gref)}\\
    &=\int dg_0\int d\gref\ \cancel{P(g|g_0,\gref,\Oscene,\Ograsp;t)}\frac{P(g_0,\gref|\Oscene,\Ograsp)}{P_t(g|\Oscene,\Ograsp)}\frac{\Grad \LinvKernel_{t}(\gref^{-1}g_0^{-1}g\gref)}{\cancel{\LinvKernel_{t}(\gref^{-1}g_0^{-1}g\gref)}} \tag{ $\because\ P(g|g_0,\gref,\Oscene,\Ograsp;t)=P(g|g_0,\gref;t)=\LinvKernel_{t}(\gref^{-1}g_0^{-1}g\gref)\ $ }\\
    &=\frac{1}{P_t(g|\Oscene,\Ograsp)}\int dg_0\int d\gref\ P(g_0,\gref|\Oscene,\Ograsp)\Grad \LinvKernel_{t}(\gref^{-1}g_0^{-1}g\gref)\\
    &=\frac{1}{P_t(g|\Oscene,\Ograsp)}\Grad \int dg_0\int d\gref\ P(g_0,\gref|\Oscene,\Ograsp)\LinvKernel_{t}(\gref^{-1}g_0^{-1}g\gref)\\
    &=\frac{1}{P_t(g|\Oscene,\Ograsp)}\Grad \int dg_0\,P_0(g_0|\Oscene,\Ograsp)\int d\gref\ P(\gref|g_0^{-1}\cdot\Oscene,\Ograsp)\LinvKernel_{t}(\gref^{-1}g_0^{-1}g\gref)\\
    &=\frac{1}{P_t(g|\Oscene,\Ograsp)}\Grad P_t(g|\Oscene,\Ograsp)\tag{$\because$\ \eqref{eqn:frame_selection_marginal} and \eqref{eqn:marginal}}\\
    &=\frac{\Grad P_t(g|\Oscene,\Ograsp)}{P_t(g|\Oscene,\Ograsp)}=\Grad \log P_t(g|\Oscene,\Ograsp)
    \end{align*}
\end{proof}
Therefore, we prove that $\vs_t^*(g|\Oscene,\Ograsp) = \Grad \log P_t(g|\Oscene,\Ograsp)$.

\subsection{Proof of Proposition~\ref{propo:score-model}}
\label{proof:score-model}
For readers' convenience, we reproduce the bi-equivariance conditions for the score functions in Proposition~\ref{propo:score_left_and_right} with explicit components.
\begin{align}
    \ \vs(\Dg\,g|\Dg\cdot \Oscene,\Ograsp) &= \vs(g|\Oscene,\Ograsp) \label{eqn_appndx:left_score}\\
    \begin{split}
        \ \vs(g\,\Dg^{-1}|\Oscene,\Dg\cdot \Ograsp) &=\left[\mathrm{Ad}_{\Dg}\right]^{-T} \vs(g|\Oscene,\Ograsp)
        \\&=\begin{bmatrix}
            \Delta R & \emptyset\\
            [\Delta\vp]^{\wedge}\Delta R & \Delta R
            \end{bmatrix}
        \begin{bmatrix}
            \vs_{\nu}(g|\Oscene,\Ograsp) \\ \vs_{\omega}(g|\Oscene,\Ograsp)
        \end{bmatrix}
        \\&=\Delta R\,\vs_{\nu}(g|\Oscene,\Ograsp) \oplus \left[\Delta R\,\vs_{\omega}(g|\Oscene,\Ograsp) +\Delta\vp\cross\Delta R\,\vs_{\nu}(g|\Oscene,\Ograsp)\right]
    \end{split} \label{eqn_appndx:right_score}
\end{align}
where we used the fact that the inverse transpose of the adjoint matrix is as follows \citep{murray2017mathematical,lynch2017modern}:
\begin{equation}
    \left[\mathrm{Ad}_{\Dg}\right]^{-T}
            =\begin{bmatrix}
            \Delta R & \emptyset\\
            [\Delta\vp]^{\wedge}\Delta R & \Delta R
            \end{bmatrix}
\end{equation}

We begin by proving the bi-equivariance of the linear (translational) score term
\begin{proof}
    The left invariance of the linear score model is proved as
    \begin{align*}
        \vs_{\nu;t}(\Dg\,g|\Dg\cdot\Oscene,\Ograsp) &= \int_{\mathbb{R}^3} d^3\vx\ \rho_{\nu;t}(\vx|\Ograsp)\ \widetilde{\vs}_{\nu;t}(\Dg\,g,\vx|\Dg\cdot\Oscene,\Ograsp)
        \\&=\int_{\mathbb{R}^3} d^3\vx\ \rho_{\nu;t}(\vx|\Ograsp)\ \widetilde{\vs}_{\nu;t}(g,\vx|\Oscene,\Ograsp)\tag{$\because $~\eqref{eqn:score_field_left_equiv}}
        \\&=\vs_{\nu;t}(g|\Oscene,\Ograsp)
    \end{align*}
    
    The right equivariance of the linear score model is proved as
    \begin{align*}
        \vs_{\nu;t}(g\,\Dg^{-1}|\Oscene,\Dg\cdot\Ograsp) 
        &= \int_{\mathbb{R}^3} d^3\vx\ \rho_{\nu;t}(\vx|\Dg\cdot\Ograsp)\ \widetilde{\vs}_{\nu;t}(g\,\Dg^{-1},\vx|\Oscene,\Dg\cdot\Ograsp)
        \\&=\int_{\mathbb{R}^3} d^3\vx\ \rho_{\nu;t}(\Dg\,\vx|\Dg\cdot\Ograsp)\ \widetilde{\vs}_{\nu;t}(g\,\Dg^{-1},\Dg\,\vx|\Oscene,\Dg\cdot\Ograsp) \tag{$\because $\ invariance of Euclidean integral under roto-translation  $\vx\rightarrow\Dg\,\vx$}
        \\&=\int_{\mathbb{R}^3} d^3\vx\ \rho_{\nu;t}(\vx|\Ograsp)\ \Delta R\,\widetilde{\vs}_{\nu;t}(g,\vx|\Oscene,\Ograsp)\tag{$\because $~\eqref{eqn:density_equiv} and~\eqref{eqn:score_field_right_equiv}}
        \\&=\Delta R\,\int_{\mathbb{R}^3} d^3\vx\ \rho_{\nu;t}(\vx|\Ograsp)\ \widetilde{\vs}_{\nu;t}(g,\vx|\Oscene,\Ograsp)
        \\&=\Delta R\,\vs_{\nu;t}(g|\Oscene,\Ograsp)
    \end{align*}
\end{proof}
Let the angular (rotational) score model be decomposed into the spin term $\vs_{\textrm{spin};t}$ and the orbital term $\vs_{\textrm{orbital};t}$ as in \eqref{eqn:omg_model}.
The bi-equivariance of spin term in the angular (rotational) score model
\begin{align}
    \vs_{\textrm{spin};t}(g|\Oscene,\Ograsp)&=\int_{\mathbb{R}^3} d^3\vx\ \rho_{\omega;t}(\vx|\Ograsp)\ \widetilde{\vs}_{\omega;t}(g,\vx|\Oscene,\Ograsp)
    \\\vs_{\textrm{spin};t}(\Dg\,g|\Dg\cdot\Oscene,\Ograsp)&=\vs_{\textrm{spin};t}(g|\Oscene,\Ograsp)
    \\\vs_{\textrm{spin};t}(g\,\Dg^{-1}|\Oscene,\Dg\cdot\Ograsp)&=\Delta R\,\vs_{\textrm{spin};t}(g|\Oscene,\Ograsp)
\end{align}
 can be proven in a similar fashion to the linear score model. It can be shown that the orbital term satisfies the following bi-equivariance condition
\begin{align}
    \vs_{\textrm{orbital};t}(g|\Oscene,\Ograsp)&=\int_{\mathbb{R}^3} d^3\vx\ \rho_{\nu;t}(\vx|\Ograsp)\ \vx\cross\widetilde{\vs}_{\nu;t}(g,\vx|\Oscene,\Ograsp)
    \\\vs_{\textrm{orbital};t}(\Dg\,g|\Dg\cdot\Oscene,\Ograsp)&=\vs_{\textrm{orbital};t}(g|\Oscene,\Ograsp)
    \\\vs_{\textrm{orbital};t}(g\,\Dg^{-1}|\Oscene,\Dg\cdot\Ograsp)&=\Delta\vp\cross\Delta R\,\vs_{\textrm{orbital};t}(g|\Oscene,\Ograsp)
\end{align}
\begin{proof}
    The left invariance is straightforward, as the linear score field $\widetilde{\vs}_{\nu;t}$ is left-invariant as~\eqref{eqn:score_field_left_equiv}.
    The right equivariance can be proved as follows
\begin{align*}
    &\vs_{\textrm{orbital};t}(g\,\Dg^{-1}|\Oscene,\Dg\cdot\Ograsp)
    \\&=\int_{\mathbb{R}^3} d^3\vx\ \rho_{\nu;t}(\vx|\Dg\cdot\Ograsp)\ \vx\cross\widetilde{\vs}_{\nu;t}(g\,\Dg^{-1},\vx|\Oscene,\Dg\cdot\Ograsp)
    \\&=\int_{\mathbb{R}^3} d^3\vx\ \rho_{\nu;t}(\Dg^{-1}\,\vx|\Ograsp)\ 
    \vx\cross\Delta R\,\widetilde{\vs}_{\nu;t}(g,\Dg^{-1}\,\vx|\Oscene,\Ograsp)
    \tag{$\because $~\eqref{eqn:density_equiv} and~\eqref{eqn:score_field_right_equiv}}
    \\&=\int_{\mathbb{R}^3} d^3\vx\ \rho_{\nu;t}(\vx|\Ograsp)\ 
    \left(\Delta R\,\vx+\Delta\vp\right)\cross\Delta R\,\widetilde{\vs}_{\nu;t}(g,\vx|\Oscene,\Ograsp)
    \tag{$\because $\ invariance of Euclidean integral under roto-translation  $\vx\rightarrow\Dg\,\vx=\Delta R\,\vx+\Delta\vp$}
    \\&=\Delta R\,\left[\int_{\mathbb{R}^3} d^3\vx\ \rho_{\nu;t}(\vx|\Ograsp)\ 
    \vx\cross\widetilde{\vs}_{\nu;t}(g,\vx|\Oscene,\Ograsp)\right] 
    + \Delta\vp\cross\Delta R\,\int_{\mathbb{R}^3} d^3\vx\ \rho_{\nu;t}(\vx|\Ograsp)\ 
    \widetilde{\vs}_{\nu;t}(g,\vx|\Oscene,\Ograsp)
    \tag{$\because\,R\,\vx\cross R\,\vy = R\left(\vx\cross\vy\right)\ \ \forall\,\vx,\vy\in\mathbb{R}^{3}$}
    \\&=\Delta R\,\vs_{\textrm{orbital};t}(g|\Oscene,\Ograsp) + \Delta\vp\cross\Delta R\,\vs_{\nu;t}(g|\Oscene,\Ograsp)
\end{align*}
\end{proof}

As a result, the angular (rotational) score model 
\begin{equation}
    \vs_{\omega;t}(g|\Oscene,\Ograsp)=\vs_{\textrm{orbital};t}(g|\Oscene,\Ograsp)+\vs_{\textrm{spin};t}(g|\Oscene,\Ograsp)
\end{equation}
satisfies the following bi-equivariance
\begin{align}
    \vs_{\omega;t}(\Dg\,g|\Dg\cdot\Oscene,\Ograsp)=&\vs_{\omega;t}(g|\Oscene,\Ograsp)\\
    \begin{split}
        \vs_{\omega;t}(g\,\Dg^{-1}|\Oscene,\Dg\cdot\Ograsp)=&\Delta R\left[\vs_{\textrm{orbital};t}(g|\Oscene,\Ograsp)+\vs_{\textrm{spin};t}(g|\Oscene,\Ograsp)\right]
        + \Delta\vp\cross\Delta R\,\vs_{\nu;t}(g|\Oscene,\Ograsp)
        \\=&\Delta R\,\vs_{\omega;t}(g|\Oscene,\Ograsp)+ \Delta\vp\cross\Delta R\,\vs_{\nu;t}(g|\Oscene,\Ograsp)
    \end{split}
\end{align}
Hence, we have proven Proposition~\ref{propo:score-model} that the score model in \eqref{eqn:score_model} is bi-equivariant, satisfying \eqref{eqn_appndx:left_score} and \eqref{eqn_appndx:right_score}.

\subsection{Proof of Proposition~\ref{propo:edf-score}}
\label{proof:score-field-biequiv}
\begin{proof}
\begin{align*}
\widetilde{\vs}_{\square;t}(\Dg\,g,\vx|\Dg\cdot\Oscene,\Ograsp)
&=\boldsymbol{\psi}_{\square;t}(\vx|\Ograsp)\ \otimes_{\square;t}^{(\rightarrow 1)}\ \D (R^{-1}\Delta R^{-1})\,\boldsymbol{\varphi}_{\square;t}(\Dg\,g\,\vx|\Dg\cdot\Oscene)
\\&=\boldsymbol{\psi}_{\square;t}(\vx|\Ograsp)\ \otimes_{\square;t}^{(\rightarrow 1)}\ \D (R^{-1}\Delta R^{-1})\D (\Delta R)\,\boldsymbol{\varphi}_{\square;t}(g\,\vx|\Oscene) \tag{$\because $ \eqref{eqn:edf_steer}}
\\&=\boldsymbol{\psi}_{\square;t}(\vx|\Ograsp)\ \otimes_{\square;t}^{(\rightarrow 1)}\ \D (R^{-1}\cancel{\Delta R^{-1}\Delta R})\,\boldsymbol{\varphi}_{\square;t}(g\,\vx|\Oscene) \tag{$\because $~\eqref{eqn:homomorphism}}
\\&=\widetilde{\vs}_{\square;t}(g,\vx|\Oscene,\Ograsp)
\end{align*}

\begin{align*}
&\widetilde{\vs}_{\square;t}(g\,\Dg^{-1},\Dg\,\vx|\Oscene,\Dg\cdot\Ograsp)
\\&=\boldsymbol{\psi}_{\square;t}(\Dg\,\vx|\Dg\cdot\Ograsp)\ \otimes_{\square;t}^{(\rightarrow 1)}\ \D (\Delta R\,R^{-1})\,\boldsymbol{\varphi}_{\square;t}(g\,\cancel{\Dg^{-1}\,\Dg}\,\vx|\Oscene)
\\&=\D (\Delta R)\,\boldsymbol{\psi}_{\square;t}(\vx|\Ograsp)\ \otimes_{\square;t}^{(\rightarrow 1)}\ \D (\Delta R\,R^{-1})\,\boldsymbol{\varphi}_{\square;t}(g\,\vx|\Oscene) \tag{$\because $ \eqref{eqn:edf_steer}}
\\&=\D (\Delta R)\,\boldsymbol{\psi}_{\square;t}(\vx|\Ograsp)\ \otimes_{\square;t}^{(\rightarrow 1)}\ \D (\Delta R)\,\D (R^{-1})\,\boldsymbol{\varphi}_{\square;t}(g\,\vx|\Oscene) \tag{$\because $~\eqref{eqn:homomorphism}}
\\&=\D_{1}(\Delta R)\left[\boldsymbol{\psi}_{\square;t}(\vx|\Ograsp)\ \otimes_{\square;t}^{(\rightarrow 1)}\ \D (R^{-1})\,\boldsymbol{\varphi}_{\square;t}(g\,\vx|\Oscene)\right] 
\tag{$\because\ \left[\D (R)\vv\right] \otimes^{(\rightarrow l)}\left[\D (R)\vw\right]=\D_{l}(R)\left[\vv\otimes^{(\rightarrow l)}\vw\right]$}
\\&=\Delta R\,\widetilde{\vs}_{\square;t}(g,\vx|\Oscene,\Ograsp)
\end{align*}
where in the last line we assume that the degree-$1$ Wigner D-matrix $\D_{1}(\cdot)$ is in the real basis with $x-y-z$ axis ordering. Note that the last line only holds in this specific choice of basis. Therefore, the type-$1$ or higher descriptors of the two EDFs must be defined in this basis.
\end{proof}
\section{Implementation Details}
\label{appndx:imple_detail}

\subsection{Score Field Model Details} 
We assume that the output of the score field model in \eqref{eqn:score_field_model} is a dimensionless quantity. Therefore, we obtain the dimensionful score by taking
\begin{align*}
    \widetilde{\vs}_{\nu;t}\rightarrow\frac{1}{L\sqrt{t}}\widetilde{\vs}_{\nu;t},\qquad \ \widetilde{\vs}_{\omega;t}\rightarrow\frac{1}{\sqrt{t}}\widetilde{\vs}_{\omega;t}
\end{align*}
where $L$ is the characteristic length scale unit. The reason for dividing $1/\sqrt{t}$ is because the norm of the target score tend to scale with $O(1/\sqrt{t})$. Likewise, we divide the linear score field by $L$ because score field is a gradient and thus scales reciprocally to the characteristic length scale.

For computational efficiency, we use identical EDFs for $\square=\omega,\nu$ in \eqref{eqn:score_field_model}.
In addition, we remove the time dependence of the grasp EDF $\boldsymbol{\psi}_t(\vx|\Ograsp)$ so that its field value is computed only once at the beginning of the denoising process.
In conclusion, our actual implementations of \eqref{eqn:score_sum_lin} and \eqref{eqn:score_sum_ang} are as follows:
\begin{align}
    \vs_{\nu;t}(g|\Oscene,\Ograsp) &= \frac{1}{L\sqrt{t}}\sum_{\vq\in Q(\Ograsp)}w(\vq|\Ograsp)\left[\boldsymbol{\psi}(\vq|\Ograsp)\,\otimes_{\nu;t}^{(\rightarrow 1)}\,\D (R^{-1})\,\boldsymbol{\varphi}_t(g\,\vq|\Oscene)\right]\\
    \begin{split}
        \vs_{\omega;t}(g|\Oscene,\Ograsp) &= \frac{1}{\sqrt{t}}\sum_{\vq\in Q(\Ograsp)}w(\vq|\Ograsp)\,\frac{\vq}{L}\wedge\left[\boldsymbol{\psi}(\vq|\Ograsp)\,\otimes_{\nu;t}^{(\rightarrow 1)}\,\D (R^{-1})\,\boldsymbol{\varphi}_t(g\,\vq|\Oscene)\right]\\
        & + \frac{1}{\sqrt{t}}\sum_{\vq\in Q(\Ograsp)}w(\vq|\Ograsp)\left[\boldsymbol{\psi}(\vq|\Ograsp)\,\otimes_{\omega;t}^{(\rightarrow 1)}\,\D (R^{-1})\,\boldsymbol{\varphi}_t(g\,\vq|\Oscene)\right]
    \end{split}
\end{align}

\subsection{Sampling with Annealed Langevin Dynamics}
\label{appndx:sampling}
It is known to be difficult and unstable to train and sample with the score function for a sparse distribution \citep{koehler2022statistical,song2019generative}. To address this issue, \emph{Annealed Langevin Markov Chain Monte Carlo} \citep{song2019generative} leverages the score of the diffused marginal $P_t$ instead of $P_0$. A diffused marginal $P_t(g)$ for a diffusion kernel $P_{t|0}(g|g_0)$ is defined on the $SE(3)$ manifold as
\begin{equation} 
P_t(g)=\int_{SE(3)} dg_0 P_{t|0}(g|g_0)P_0(g_0). 
\end{equation}
We utilize the trained score function $s_{t}(g) =\nabla \log P_{t}(g)$ for the annealed Langevin MCMC %as detailed in section~\ref{sec:AL-MCMC}. 
on $SE(3)$ \citep{urain2022se3dif} as
\begin{equation}
g_{\tau+d\tau}=g_\tau\exp{\left[\frac{1}{2}\vs_{t(\tau)}(g_\tau|\Oscene,\Ograsp)d\tau + dW\right]}.    
\end{equation}
where $t(\tau)$ is the diffusion time scheduling, which is gradually annealed to zero as $\tau\rightarrow\infty$, such that $t(\tau=\infty)=0$. This process will converge to $P_0(g)$ regardless of the initial distribution if it is annealed sufficiently slowly and $\underset{t\rightarrow 0}{\lim}P_{t}=P_0$. This SDE can be discretized using the forward Euler-Maruyama method such that
\begin{equation}
    \label{eqn:annealed_langevin}
    \begin{split}
        g_{n+1}=g_n\exp{\left[\frac{1}{2}\vs_{t[n]}(g_n|\Oscene,\Ograsp)\alpha[n] + \sqrt{\alpha[n]}\vz_n\right]},\ \ \vz_n\sim \mathcal{N}(\boldsymbol{0},I)
    \end{split}
\end{equation}
where $t[n]$ and $\alpha[n]$ are respectively the diffusion time and Langevin step size, both of which are scheduled according to the step count $n$. 
% While the accuracy of this diffusion process can be improved with more sophisticated SDE solvers like [Ref TODO], we find that reasonable accuracy can be obtained with Euler-Maruyama method.
A commonly used scheduling scheme is taking $\alpha[n]\propto t[n]$ with either a linear or log-linear $t[n]$ schedule \citep{song2019generative,ho2020denoising,urain2022se3dif}. 
However, the convergence is very slow with this scheduling. Therefore, we use $\alpha[n]\propto t[n]^{k_1}$ schedule with a hyperparameter ${k_1}<1$. To suppress the instability caused by large step sizes when $t$ is small,
we also gradually lower the \emph{temperature}\footnote{This temperature annealing should not be confused with that of the \emph{`annealed'} Langevin MCMC in which the diffusion time $t$ is decreased.} of the process. This can be done by using $\sqrt{\alpha[n]T[n]}\vz_n$ instead of $\sqrt{\alpha[n]}\vz_n$ for the noise term with the temperature schedule $T[n]=t[n]^{k_2}$, where ${k_2}\geq 0$ is another hyperparameter. Intuitively, this makes the sampling process to smoothly transition into a simple gradient descent optimization as $t[n]\rightarrow 0$, and hence $T[n]\rightarrow 0$. 
We empirically found that this strategy significantly improves the convergence time without compromising the accuracy and diversity of the sampled poses.
The resulting sampling algorithm with a small number $\epsilon$ is
\begin{equation}
    \label{eqn:annealed_langevin_empirical}
    \begin{split}
        g_{n+1}=g_n\exp{\left[\frac{\epsilon}{2}\vs_{t[n]}(g_n|\Oscene,\Ograsp)t[n]^{k_1} + \sqrt{\epsilon}\,t[n]^{\frac{{k_1}+{k_2}}{2}}\vz_n\right]},\ \ \vz_n\sim \mathcal{N}(\boldsymbol{0},I)
    \end{split}
\end{equation}

We use ${k_1}=0.5$ and ${k_2}=1.0$ for the step size and temperature scheduling.
For the diffusion time $t[n]$, we use piecewise linear scheduling. For example, we linearly schedule the diffusion time for $t=1$ to $t=0.1$ and then with $t=0.1$ to $t=0.01$. Similar to diffusion-based image generation models, we separate a low-resolution model and high-resolution model instead of using a single model. We use the low-resolution model for higher $t$ and the high-resolution model for lower $t$.
Similar to \citet{ryu2023equivariant}, we solve \eqref{eqn:annealed_langevin_empirical} in the quaternion-translation parameterization of $SE(3)$ instead of performing the actual exponential mapping in \eqref{eqn:annealed_langevin_empirical}.

\subsection{Architecture details}
\label{appndx:architecture}

\begin{figure*}[ht]
  \centering
  %\framebox{\parbox{\textwidth}{\includegraphics[width=\textwidth]{figure11_solid.png}}}
  \includegraphics[width=0.85\textwidth]{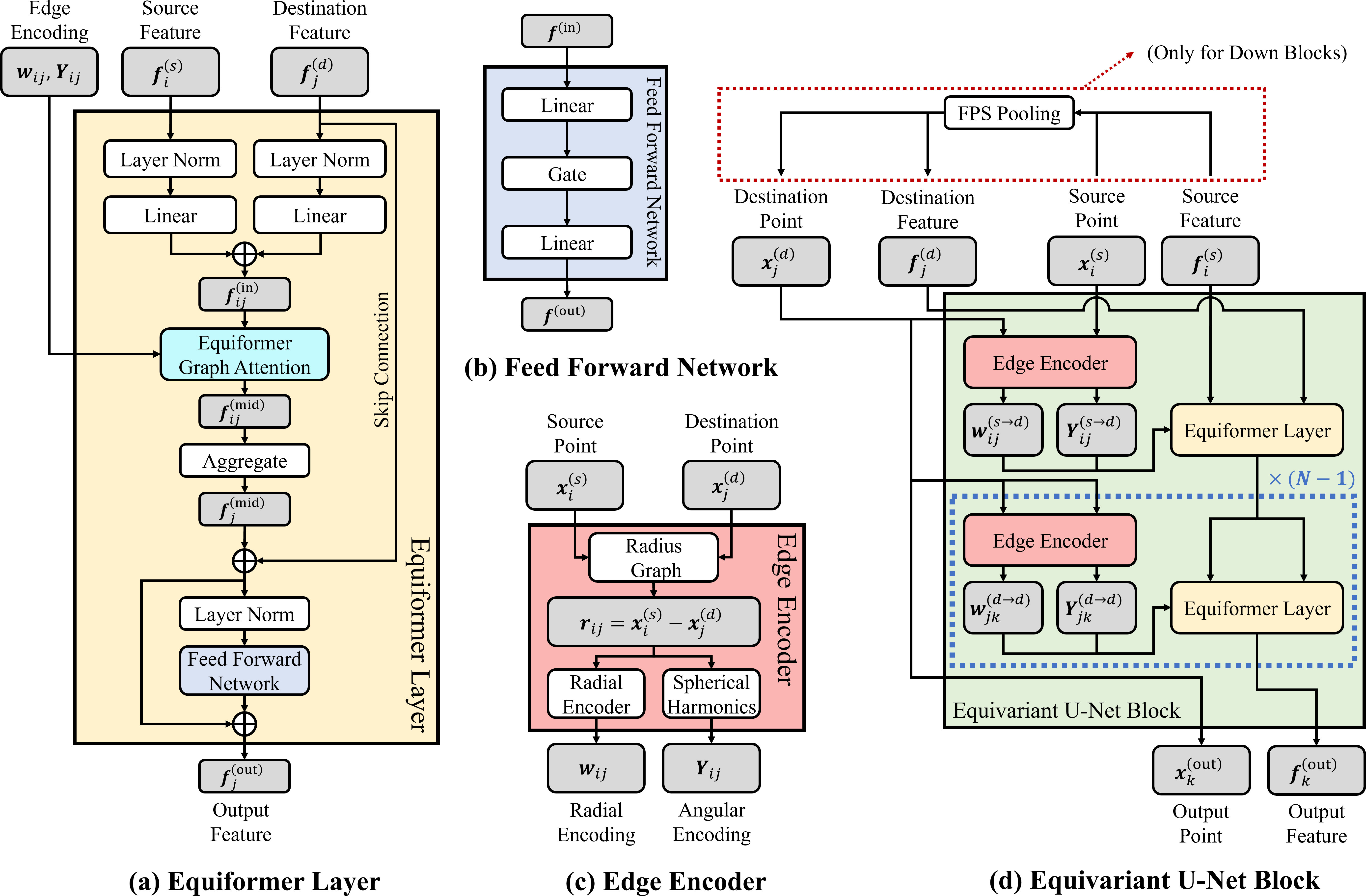}
  \caption{
    \textbf{Overview of Modules Used in Multiscale EDF.} 
    \textbf{(a)} 
    We employ Equiformer~\citep{liao2022equiformer} to achieve $SE(3)$-equivariance in our model.
    % We provide the detailed architecture of the Equiformer graph attention module in \Supp~\ref{appndx:repr_theory_gnn}.
    \textbf{(b)} We use an equivariant feed forward network with gate activation from Equiformer.
    \textbf{(c)} 
    We use radius graph to construct graph from points.
    Graph edge length and orientation are respectively encoded by a radial encoder and spherical harmonics \citep{thomas2018tensor,fuchs2020se,liao2022equiformer}. 
    \textbf{(d)} Multiple equiformer layers are stacked and form the equivariant U-Net Block. FPS pooling is used in downward blocks to obtain coarse-grained destination points from source points in lower scale-space.
  }
  \label{fig:modules}
  % \vspace*{-2mm}
\end{figure*}

See \Fig~\ref{fig:modules} for the illustration of each module used in \Fig~\ref{fig:Unet}.

\subsection{Diffusion Frame Selection Mechanism}
\label{appndx:frame_selection}
In this section, we provide further details on the diffusion frame/origin selection mechanism.

\paragraph{Necessity of Diffusion Frame/Origin Selection Mechanism.}
We first discuss why a diffusion frame/origin selection mechanism is necessary for our diffusion model on the $SE(3)$ manifold.
For simplicity, we confine our argument only to the diffusion origin selection mechanism as Proposition~\ref{propo:only-translation} suggests.

\begin{figure}[t]
  \centering
   \includegraphics[width=1\linewidth]{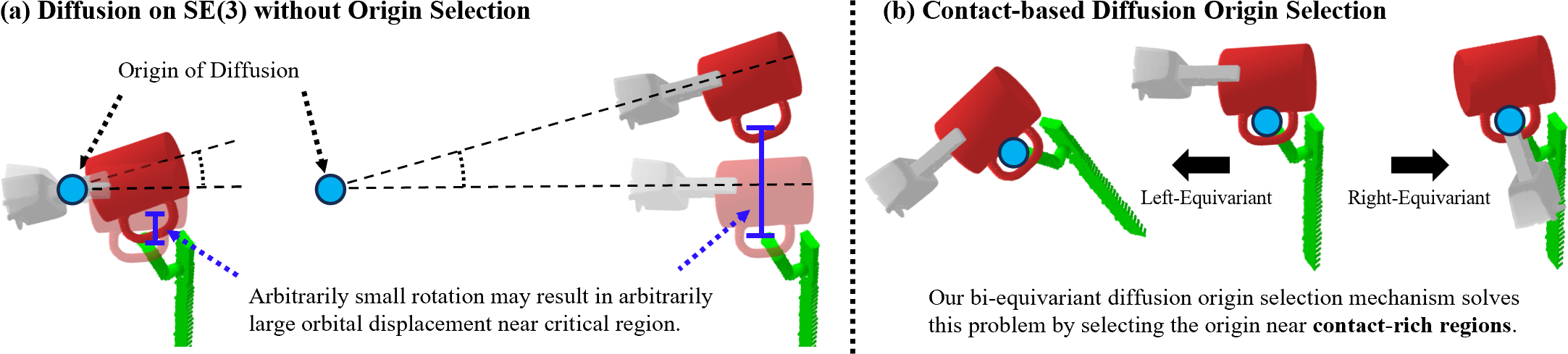}
   \caption{
        \textbf{Necessity of Diffusion Origin Selection Mechanism.} \textbf{(a)}
        % Without rotational origin fixing, pure translations cannot be distinguished from orbital effects. Furthermore, small rotational diffusion may result in arbitrarily large translational displacement near the critical region depending on the diffusion origin.
        A small rotational diffusion may result in arbitrarily large orbital displacement near the critical region depending on the diffusion origin.
        \textbf{(b)} We employ a contact-based diffusion origin selection mechanism.
        This not only allows bi-equivariant diffusion process but also stabilizes learning by minimizing the orbital impact of the rigid body rotation near the critical regions.
   }
   \label{fig:arbitrary_disp}
\end{figure}

In \Sec~\ref{sec:bi-equiv-diffusion}, we introduced the concept of diffusion frame/origin selection mechanism to achieve bi-equivariance in the diffusion process.
However, the diffusion frame/origin selection has further implication, even for non-equivariant diffusion models on the $SE(3)$ manifold.
% In fact, denoising the diffused end-effector pose $g\in SE(3)$ from the two observations $\Oscene$ and $\Ograsp$ without fixing the origin of diffusion is an \emph{ill-posed problem}.
% This is because the pure translation of a rigid body cannot be distinguished from the orbital effect of rotation without specifying the center of rotation.
% Similarly, as illustrated in \Fig~\ref{fig:arbitrary_disp}, an arbitrarily small rotational perturbation may result in an arbitrarily large displacement near the critical region depending on the choice of the origin.
As illustrated in \Fig~\ref{fig:arbitrary_disp}, an arbitrarily small rotational perturbation may result in an arbitrarily large orbital displacement near the critical region depending on the choice of the origin, leading to an unstable diffusion and denoising process.
This is in contrast to typical Euclidean diffusion models because vector addition is a commutative operation, and hence origin fixing has no effect.
% On the other hand, the effect of frame fixing does not cancel out for non-commutative groups such as $SE(3)$.
Therefore, a proper diffusion process for our problem must include a diffusion origin selection procedure to minimize the orbital effect of rotation near critical regions.

A natural selection of the diffusion origin for manipulation tasks is the origin of the end-effector frame itself.
However, this origin selection is not equivariant to the grasped object, making our diffusion kernel only left-equivariant and not right-equivariant.
Another natural diffusion origin is the centroid of the point cloud, which was utilized by \citet{yim2023se} and \citet{corso2023diffdock} for protein docking problems.
Indeed, this is a special case of an equivariant origin selection mechanism that satisfies \eqref{eqn:only-translation}.
However, as pointed out by \citet{ryu2023equivariant} and \citet{kim2023robotic}, centroids are often dominated by the global geometry rather than the critical sub-geometry of the target objects. 
Please recall that this is why R-NDFs suffer without object segmentation.
While the protein-ligand interaction problem in \citet{yim2023se} and \citet{corso2023diffdock} has additional torsional degrees of freedom to debias this centroid artifact, it won't translate to our problem since the points in $\Ograsp$ are only actuated by the end-effector pose $g$.

\paragraph{Equivariant Diffusion Origin Selection Mechanism with Contact Heuristics.}
An important quality of a good diffusion origin selection mechanism is that the selected origin should not be too far away from the critical contact-rich region. 
As illustrated in \Fig~\ref{fig:arbitrary_disp}, even a small rotational diffusion may take the critical region of the grasped object (the handle of the mug) far away from the placement target (the tip of the hanger), making training unstable.
Although this problem can be resolved by reducing the rotational noise scale of the diffusion process, it requires meticulous task-specific hyperparameter tuning.
Furthermore, as can be seen in \eqref{eqn_appndx:right_score}, the rotational score consists of the pure rotational term and the orbital term.
By studying the orbital term, one may notice that this term is non-dimensionalized by the product of the displacement term $\Delta\vp$, which is proportional to the length unit, and the translational score $\vs_{\nu}$, which is reciprocal to the length unit.
Although these two dimensionful quantities neatly cancel out each other's unit, this structure inevitably increases the variance of the score estimation when the displacement term $\Delta\vp$ is too large.
For instance, a small translational score term in the reference frame of the critical region may induce a large rotational score term in the end-effector frame if the displacement $\Delta\vp$ between these two frames is large. 
This is natural because a small rotation in the end-effector frame can dramatically change the probability of the pose if $\Delta\vp$ is large.
Therefore, it is always optimal to work in a diffusion origin near the critical region, such that $\Delta\vp$ is kept minimal.
This is the reason why we propose a contact-based diffusion origin selection mechanism in \eqref{eqn:neighbor_origin_selection}, which selects the origin near the important contact-rich sub-geometries.

\begin{figure*}[t]
  \centering
  \begin{subfigure}{0.35\linewidth}
    \centering
    \includegraphics[width=0.85\linewidth]{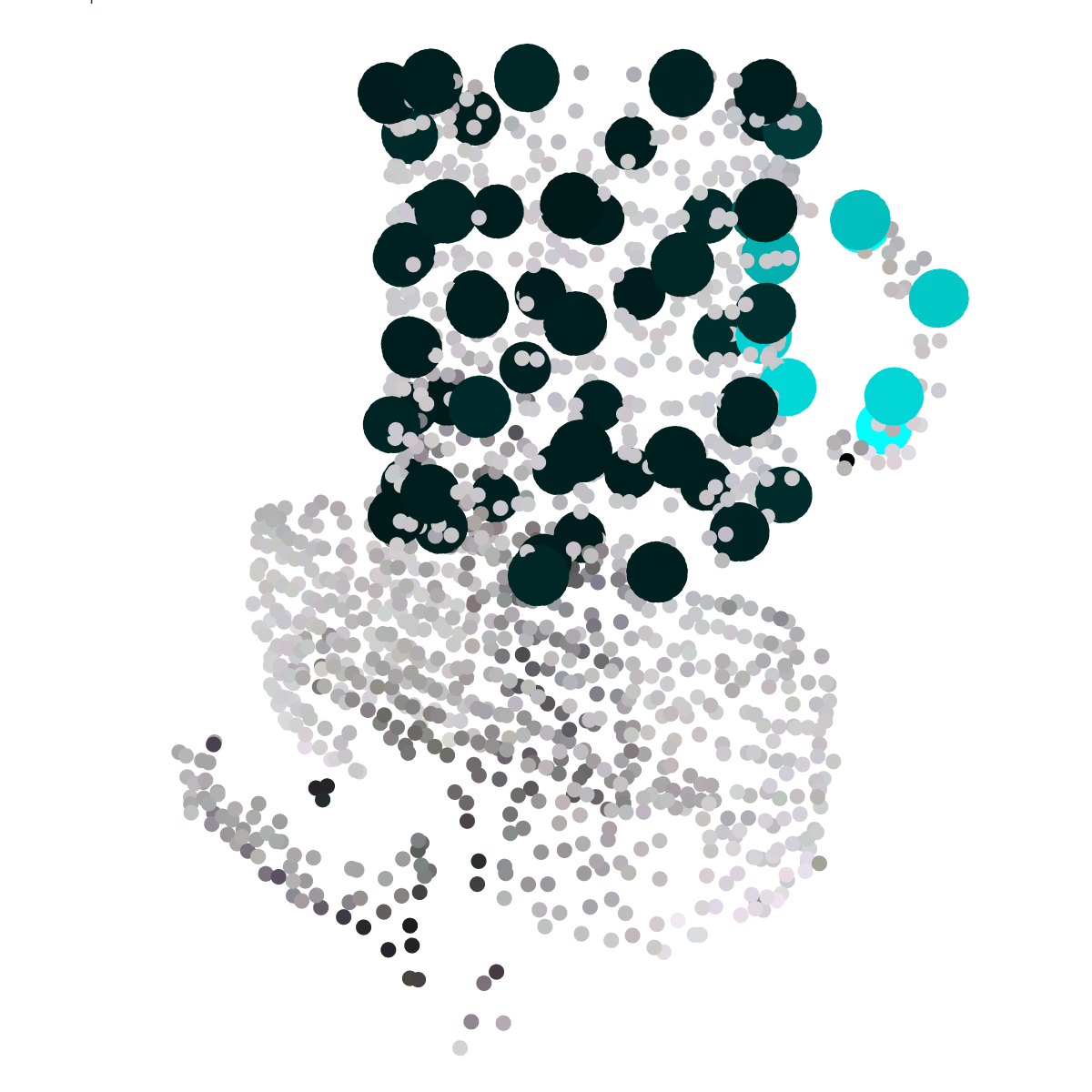}
    \caption{Query points of a real-world mug observation.}
  \end{subfigure}
  \begin{subfigure}{0.1\linewidth}
    \phantom{\includegraphics[width=1\linewidth]{images/mug_real_attn.png}}
  \end{subfigure}
  \begin{subfigure}{0.35\linewidth}
    \centering
    \includegraphics[width=0.85\linewidth]{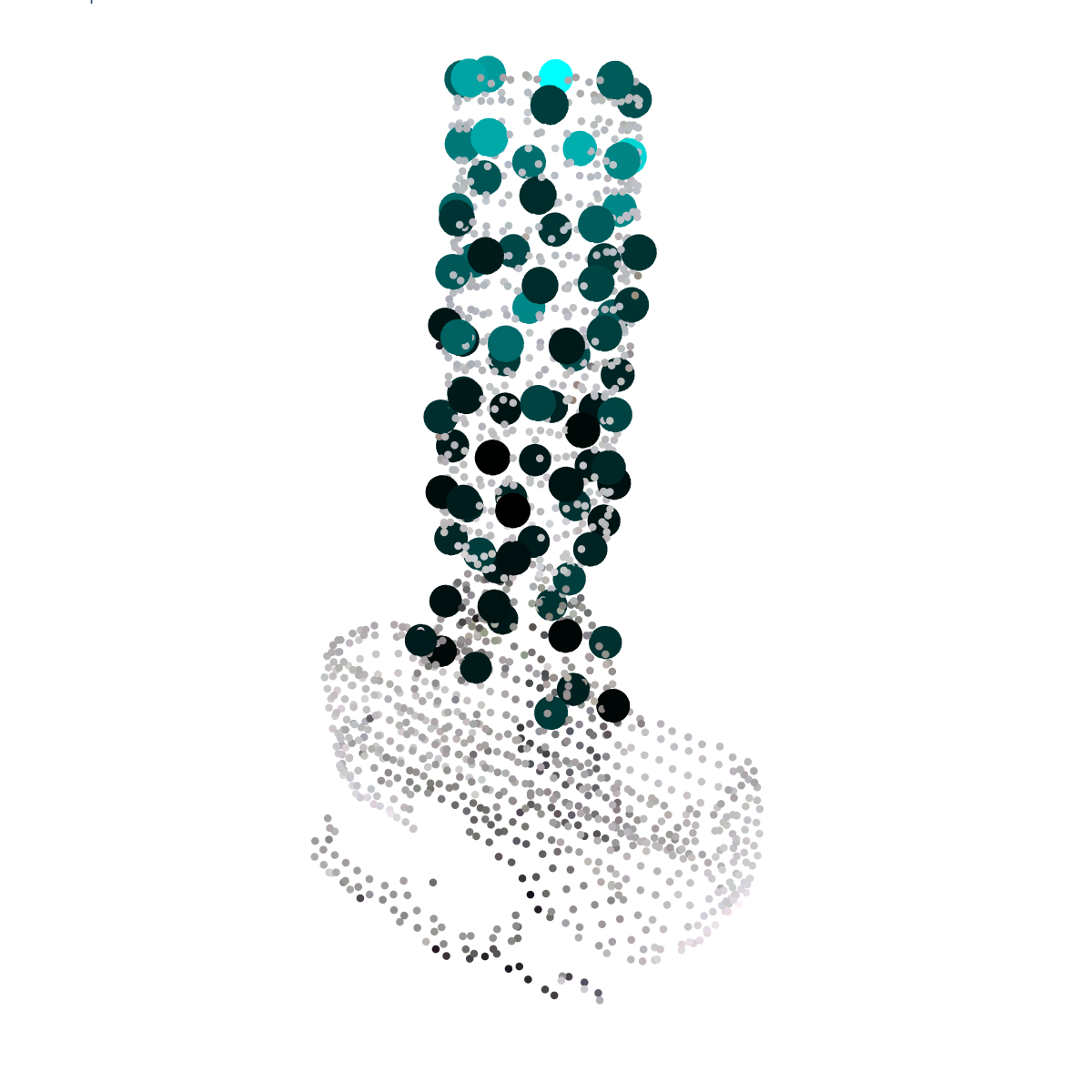}
    \caption{Query points of a real-world bottle observation.}
  \end{subfigure}
  \hfill
   \caption{
        \textbf{Learned Query Points}. The figure depicts the point clouds of a real mug and bottle with their query points visualized in colors according to their weights.
        The query points with the highest weight values are illustrated in cyan.
        The query weight field of the trained Diffusion-EDFs assigns high weight to \textbf{(a)} the mug's handle, which is the most significant sub-geometry when placing it on a hanger, and \textbf{(b)} the bottom of the bottle, which is the most significant sub-geometry when placing it on a shelf.
   }
  \label{fig:query_points}
\end{figure*}

% We introduced the equivariant diffusion origin selection mechanism of \eqref{eqn:neighbor_origin_selection} in \Sec~\ref{sec:contact-heuristic}.
We find that this origin selection mechanism stabilizes training by enabling Diffusion-EDFs to correctly identify important contact rich sub-geometries from the grasp observation $\Ograsp$.
This can be verified by visualizing the strength of the query weight field.
\Fig~\ref{fig:query_points} illustrates the query points in colors according to their query weights. 
Query points with high weights are represented in cyan and those with near-zero weights are in black.
As can be seen in the figure, the query weight field of the trained Diffusion-EDFs successfully assign high weight to the mug's handle, which is the most significant sub-geometry when placing it on a hanger.

% \begin{figure}[t]
%   \centering
%    \includegraphics[width=0.3\linewidth]{images/mug_real_attn.png}
%    \caption{
%         \textbf{Learned Query Points}. The figure depicts the point cloud of a real mug with their query points visualized in colors according to their weights.
%         The cyan query points are the query points with highest weight values.
%         The query weight field of the trained Diffusion-EDFs successfully assign high weight to the mug's handle, which is the most significant sub-geometry when placing it on a hanger. 
%    }
%    \label{fig:query_points}
% \end{figure}

\section{Experiment Details}

\subsection{Simulation Experiment Details}
\label{appndx:sim_exp}
%%%WRITING%%%
In this section, we provide further details on the simulated benchmark experiments in \Sec~\ref{sec:sim_exp}.

\subsubsection{Simulation Environment}
Evaluations are performed in a simulated enviroment using SAPIEN \citep{Xiang_2020_SAPIEN} with nine ceiling-mounted depth cameras.
We assume a perfect observation to remove the influence of point cloud processing pipelines, which is orthogonal to our research.
We also remove the impact of robot's kinematic constraints by using a floating gripper-only robot instead of simulating the full robot. In addition, we turn off the collision between the environment and allow the robot to teleport to the pre-pick/place pose in order to get rid of failures related to motion planning.
% Nonetheless, we fully simulate the grasp, because in reality, the actual grasp pose may differ from the intended pose, especially when a low-quality pose is inferred.
We evaluate the success of pick or place by turning off the collision between the environment (including the table) and the target object to manipulate, and measuring the object's z-axis position.
If the object is not firmly grasped by the gripper or is not placed on the intended placement target, the object will fall after removing the environmental collision.
Therefore, we measure the z-axis position to automatically assess whether the object has not fallen, meaning that the manipulation has succeeded.

\subsubsection{Method Details}
For each task, we train the models using ten human-generated demonstrations, in which five object instances in only upright poses are used.
In other words, each of the five object instances is demonstrated for two different pick/place poses. In the training data set, we do not use distracting objects.
% See \Fig~[TODO] for the illustrations of environmental setups for training and evaluation.
We used a custom-built web-based GUI to collect human demonstrations. 
% for the simulated robot.
% With this GUI, a trained user can provide a demonstration in less than a minute.
% See \Fig~[TODO] for the illustration of this interface.

\paragraph{Diffusion-EDFs.}
We only use ten human demonstrations to train Diffusion-EDFs in a fully end-to-end manner. No additional prior knowledge such as pre-training, object segmentation, pose estimation or data augmentation is used for Diffusion-EDFs.
For preprocessing, we use simple voxel downsampling to reduce the number of points.

\paragraph{R-NDFs.}
For R-NDFs, we use the pre-trained weights from the original implementation of \citet{simeonov2023se}.
These weights were trained with a self-supervised learning method that relies on massive amount (150 gigabytes) object geometry that are specific to the target object categories (mug, bowl, bottle; 50 gigabytes for each).
Although we do not use bowls in our experiment, we still use the weights trained from all three object categories, which achieve better performance than weights trained from only a single object category \citep{simeonov2022neural,simeonov2023se}.
Still, we observe that R-NDFs fail to place the mug on our mug hanger.
We presume that this is due to the discrepancy of the hanger's shape in our experiment and the ones used for pre-training, which were procedurally generated \citep{simeonov2023se}.
Therefore, we do R-NDFs an additional favor of using the pre-trained hanger instances instead of our hanger for the evaluation. 
Lastly, we also tried to naively pre-train the NDFs using the reconstructed meshes from the point clouds in our ten task demonstrations, but resulted in suboptimal performance (less than 5\% success rate).
These attempts show the importance of the end-to-end trainability of EDFs~\citep{ryu2023equivariant} and Diffusion-EDFs. 
R-NDFs cannot be used for uncommon object categories, as they require immense amount of category-specific data for pre-training.
Procedural generation has also turned out to be unable to resolve this problem because it cannot cover all variations in the category, which was evident in the case of the mug hanger mentioned above.

We also evaluate R-NDFs both with and without object segmentation.
It should be noted that the ability to infer without object segmentation is important not only because of its convenience.
As we have demonstrated in our real hardware experiments in \Sec~\ref{sec:real_exp}, it allows the model to understand \emph{scene-level contexts} beyond a single target object.
The experimental results in \Table~\ref{tab:sota} clearly show that R-NDFs are unable to make inference without object segmentation.
As pointed out by \citet{ryu2023equivariant}, we presume this is because of the violation of locality in R-NDFs, such as centroid subtraction.

\paragraph{SE(3)-Diffusion Fields.}
In contrast to R-NDFs, we train $SE(3)$-Diffusion Fields \citep{urain2022se3dif} using only the ten demonstrations as Diffusion-EDFs.
Following \citet{urain2022se3dif}, we jointly train the model to match both the signed distance function and the score function.
We specifically use the \emph{PoiNt-SE(3)-DiF} variant in the original paper \citep{urain2022se3dif}.
Although this model utilizes $SO(3)$-equivariant point cloud encoder based on VN-PointNet \citep{deng2021vector}, the overall architecture is not equivariant.
Therefore, we use $SO(3)$ rotational data augmentation to complement the lack of equivariance.

Similar to R-NDFs, we evaluate $SE(3)$-Diffusion Fields both with and without object segmentation. 
With object segmentation, $SE(3)$-Diffusion Fields could learn to pick up the target object, although the success rates are much lower than Diffusion-EDFs. 
Without object segmentation, they achieve success rates lower than 15\% for all scenarios. 
% Although their success rate is low, $SE(3)$-Diffusion Fields do not completely fail, which is in contrast to R-NDFs.

% \paragraph{Conclusion}
% As a result, R-NDFs \cite{simeonov2023se} and SE(3)-DiffusionFields \citep{urain2022se3dif} demonstrated their necessary for a segmentation pipeline, demonstrating their inability to understand the scene contexts. Moreover, these models required an accurate mesh for the models to operate correctly requiring an additional surface reconstruction pipeline for accurate results.

\subsection{Real Hardware Experiment Details}
\label{appndx:real_exp}
% In this section, we provide further details on the real robot experiments in \Sec~\ref{sec:real_exp}

\subsubsection{Experimental Setup}
We use a Franka Emika Panda robot arm with two Intel RealSense D415 RGB-D cameras. 
The first camera is attached to the wrist of the robot.
The robot moves around the workspace to observe RGB-D images of the scene from multiple viewpoints.
We employ RTAB-Map~\citep{labbe2019rtab}, a 3D SLAM technique, to convert these observations into a point cloud of the scene.
Rather than relying on visual odometry, we take advantage of the forward kinematics solution from the robot's joint encoders, which is more precise.
Although we use 3D SLAM-based approach in our experiments, this procedure can be skipped if multiple well-calibrated external cameras are available.
The second camera is installed on the table to observe the point cloud of the robot's gripper.
This external camera is calibrated to the ArUco marker~\citep{garrido2014automatic} frame attached to the robot's end-effector.
% We take the RGB-D images of the grasped object from 360$^\circ$ view by rotating the end-effector.
% These observations are registered into the grasp point cloud.
All the point clouds are post-processed using Open3D~\citep{Zhou2018}, in which we remove statistical outliers and apply voxel filtering.
We also apply hue and lightness augmentation for the training data to obtain robustness under light condition changes.

In our experimental procedure, the robot first moves along a predefined trajectory to observe the scene.
RTAB-Map is used to convert these observations into the point cloud of the scene in real time. Diffusion-EDFs take this point cloud to generate the end-effector poses to pick the target object.
After picking the object, the robot moves to the predefined grasp observation pose. The robot then rotates its grasped object by 360$^\circ$, and the external camera observes it.
These observations are then registered into the grasp point cloud.
For the scene point cloud, we use the same one that we used to infer the pick pose.
With these two point clouds, Diffusion-EDFs infer the end-effector poses to place the grasped object onto the placement target.
For the collection of human demonstrations, we follow a procedure similar to that in the aforementioned inference pipeline. 
The only difference is that the target pose demonstration is manually provided by a human instead of Diffusion-EDFs.

% In our experimental procedure, we employed RTAB-Map to capture point clouds of the scene, following a predefined scanning trajectory. Subsequently, leveraging the observed point cloud data from the scene, we estimated the end effector's target pose for the 'pick' operation and executed the grasping using the estimated target pose.

% After successful grasping, we repositioned the end effector to a location with a clear view of the object using the camera fixed to the table. We then conducted extrinsic calibration and initiated scanning by rotating the end effector along its z-axis. Following this, we estimated the end effector's target pose for the 'place' operation, using both the scene and the grasped object's point cloud data, and performed the placing using the estimated target pose.

% \subsubsection{Training Details}
% \paragraph{Demo Collection Procedure} For the collection of human demonstrations, we followed a similar pipeline, starting with the initial scene scan. We then instructed the end effector with the 'pick' operation pose and closed the gripper to scan the grasped object. After obtaining the grasp point cloud through scanning, we taught the end effector the pose for the 'place' operation. Each demonstration included point cloud data for the scene and the grasped object, as well as target pose information. In our pipeline, we apply post-processing to the point clouds using Open3D~\citep{Zhou2018}

\subsubsection{System Engineering}
\label{appndx:sys_eng}

\paragraph{Motion Primitives.}
While it is theoretically possible to generate a collision-free motion plan for any reachable goal pose, it is challenging in reality due to the imprecise nature of point cloud observations.
Therefore, determining how to approach the target pose is also an important problem.
As we focus only on the problem of inferring the target pose itself in this work, we simply assume that we already have task-specific motion primitives to approach the generated goal pose.
In all three real-world tasks, we use a simple motion primitive of picking along the end-effector's z-axis direction (the direction in which the gripper is pointing), and placing the target object in the top-down direction.
The robot first moves to the pre-pick/place pose by following the collision-free trajectory found by an off-the-shelf motion planner.
The motion primitives are then used to approach the generated target pick/place pose from the previous pre-pick/place pose.
After successful picking or placing, we initiate post-pick/place primitives.
We simply lift up the end-effector for the post-pick primitive. For the post-place primitive, we retract the end-effector towards the opposite direction that was taken in the pre-pick manuever.
We use MoveIt~\citep{coleman2014reducing} for motion planning and use the TOPP-RA~\citep{toppra} algorithm to time-parameterize our waypoint-based motion primitives.

Although we use predefined motion primitives, not every problem can be solved in this way. Therefore, more general approach should also encompass learning not only the target pose but also the approach direction.
We expect that our score model in \eqref{eqn:score_model} can be used for this purpose with slight modifications. 
The approach direction can be represented as the displacement between the pre-pick/place pose and the target pose.
This displacement can be effectively expressed as an $\mathfrak{se(3)}$ Lie algebra vector.
Therefore, our score model can be modified to equivariantly infer this Lie algebra vector that represents the approach direction.
We leave this research for future studies.

\paragraph{Energy-based Critic.}
Due to the collision and kinematic constraint of the robot, not every pose generated by Diffusion-EDFs are feasible. Although we ignored this problem in our simulation experiment, this problem must be considered in real robot applications.
Therefore, similar to \citet{urain2022se3dif} and \citet{ryu2023equivariant}, we generate multiple samples in parallel and reject infeasible poses one by one until a reachable pose is found.

However, it is difficult to ensure convergence for every generated sample as we use a limited number of Langevin steps to achieve reasonable inference time (5$\sim$17 seconds).
The number of unconverged samples tend to be larger in our real-world experiment with noisy observations than in the simulated ones with perfect observations.
Furthermore, rejecting infeasible poses often leads to the elimination of correct poses and the selection of unconverged wrong poses.
\citet{urain2022se3dif} and \citet{ryu2023equivariant} circumvented this problem by sorting the generated samples according to the learned energy function, which evaluates the quality of the generated poses.
In contrast to these works, however, our method does not have an explicit scalar function that can be utilized.

Therefore, we train an auxiliary energy function to sort the generated poses according to their quality. We first modify the bi-equivariant energy function of \citet{ryu2023equivariant} to allow diffusion time conditioning. We then take the Lie derivatives to obtain the energy-based score model similar to \citet{urain2022se3dif}. This energy-based score model is trained using the loss function in \eqref{eqn:score_matching_loss} with proper non-dimensionalization. 
Although this score-matching model is far less accurate than our original model in \eqref{eqn:score_model} due to the inflexible nature of energy-based diffusion models, the trained energy function is sufficient to distinguish between unconverged samples and converged samples.

With the learned energy function, we first sort the generated samples according to their energy value.
If the energy function is well trained, lower-energy samples should be better than higher-energy samples. 
However, in contrast to the MCMC-based training of \citet{ryu2023equivariant}, our diffusion-based energy function training does not have a contrastive mechanism to penalize the model for assigning low energy to outlier poses.
Therefore, our energy function often assigns too low energy values to outlier poses, although the training is much faster.
Nevertheless, we find that simply rejecting too-low-energy outliers effectively solves this problem.
Therefore, we remove the first few samples from the sorted list and start from samples with moderately low energy.
We then try motion planning for each sample until a feasible pose is found.
This strategy drastically improves the success rate of pick-and-place tasks in our real-world tasks.

\subsubsection{Experimental Results Details}
Note that it is difficult to precisely measure the performance of Diffusion-EDFs for real-world tasks as the success rate is determined not only by the inference quality but also the quality of observation, localization, and motion planning.
For instance, noisy observation and localization cause success rates to drop for subtasks that require high precision, such as mug placement and bottle picking, even though Diffusion-EDFs accurately generated correct target poses.
Challenges associated with motion planning can also reduce the success rate, particularly for subtasks that require difficult 6-DoF manipulation, such as mug placement.
We achieve over 90\% success rate for all subtasks except the mug placement and bottle picking. For these two tasks, the success rates are roughly around 80\%.
% Most of the failures in these tasks were due to a slight miss of sub-centimeter imprecision in the pose.
The majority of the errors in these tasks were caused by a slight lack of accuracy in the position that was less than a centimeter.
Note that these real hardware success rates may largely differ across systems, depending on the quality of observation, calibration, motion planning and control pipelines, which are orthogonal to our research. The video of the real robot manipulation experiments can be found in our project website (\url{https://sites.google.com/view/diffusion-edfs/home}).
In the video, our robot performs 5 to 6 pick-and-place operations in one take without failure,
showcasing that Diffusion-EDFs can solve all three real-world tasks with high success rates.

\begin{figure}[t]
   \vspace{-1.5\baselineskip}
  \centering
   \includegraphics[width=0.95\linewidth]{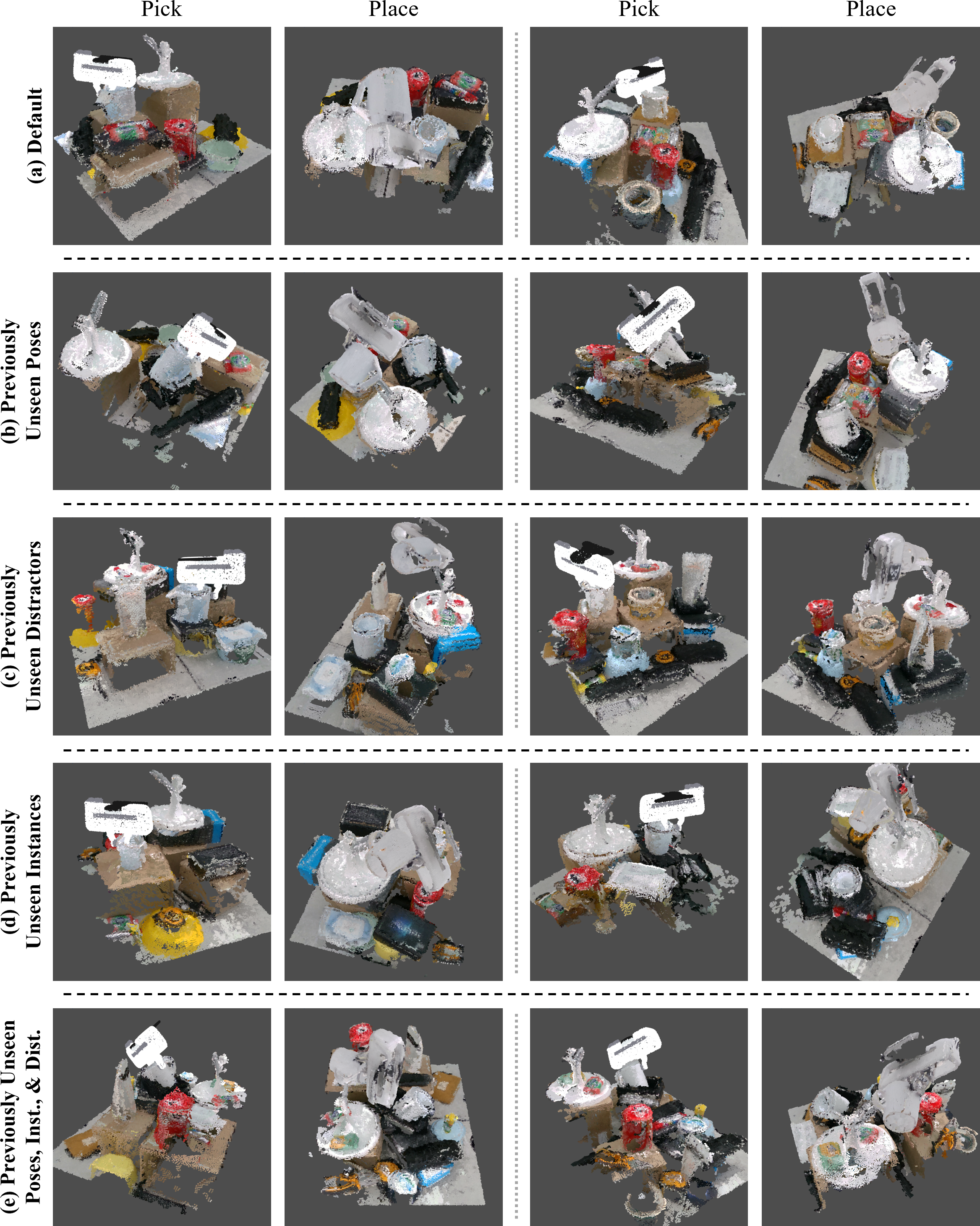}
   \caption{
        \textbf{Samples Generated by Diffusion-EDFs for Real-world Mug-on-a-hanger Task.} The figure depicts the end-effector pose samples for picking and placing a white mug on a white mug hanger. Diffusion-EDFs trained with only ten human demonstrations generated these samples from the real-world point cloud observations of the scene and grasp. 
        Similar to our simulation experiments, we experiment for the \textbf{(a)} default scenario, \textbf{(b)} previously unseen target object poses (oblique; note that we only trained Diffusion-EDFs for upright poses) scenario, \textbf{(c)} previously unseen adversarial distractors (in white color) scenario, \textbf{(d)} previously unseen target object instances scenario, and \textbf{(e)} the all scenarios combined. The video of the denoising diffusion process can be found in \url{https://sites.google.com/view/diffusion-edfs/home}
   }
   \label{fig:mug_result}
\end{figure}

\begin{figure}[t]
    \vspace{-1.5\baselineskip}
  \centering
   \includegraphics[width=0.93\linewidth]{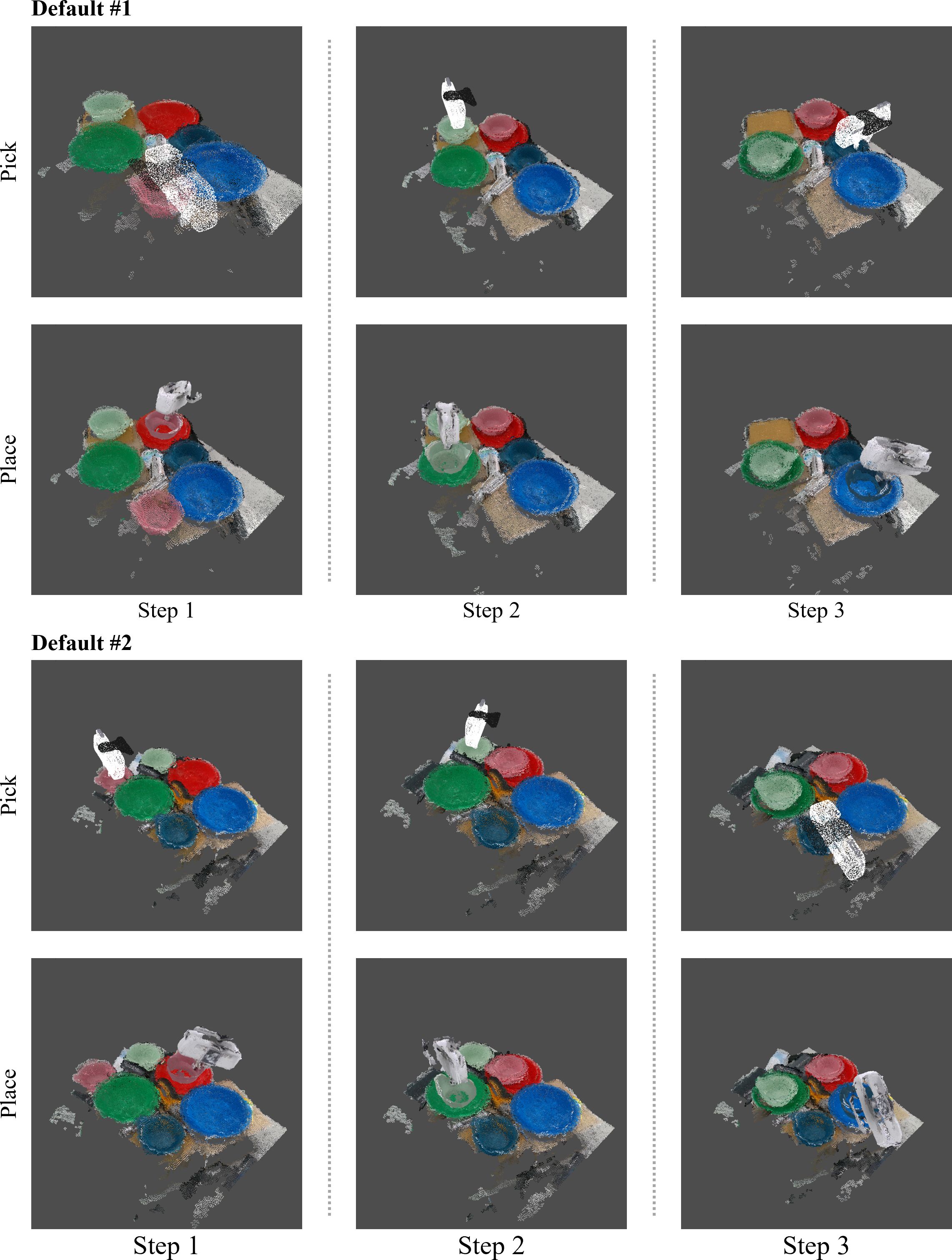}
   \caption{
        \textbf{Samples Generated by Diffusion-EDFs for Real-world Bowls-on-dishes Task.} The figure depicts the end-effector pose samples for picking and placing bowls on the dishes of matching colors in red-green-blue order. Diffusion-EDFs trained with only ten human demonstrations (three colored subtasks for each) generated these samples from the real-world point cloud observations of the scene and grasp. 
        The video of the denoising diffusion process can be found in \url{https://sites.google.com/view/diffusion-edfs/home}
   }
   \label{fig:bowl_result_episodes}
\end{figure}

\begin{figure}[t]
    \vspace{-2\baselineskip}
  \centering
   \includegraphics[width=0.9\linewidth]{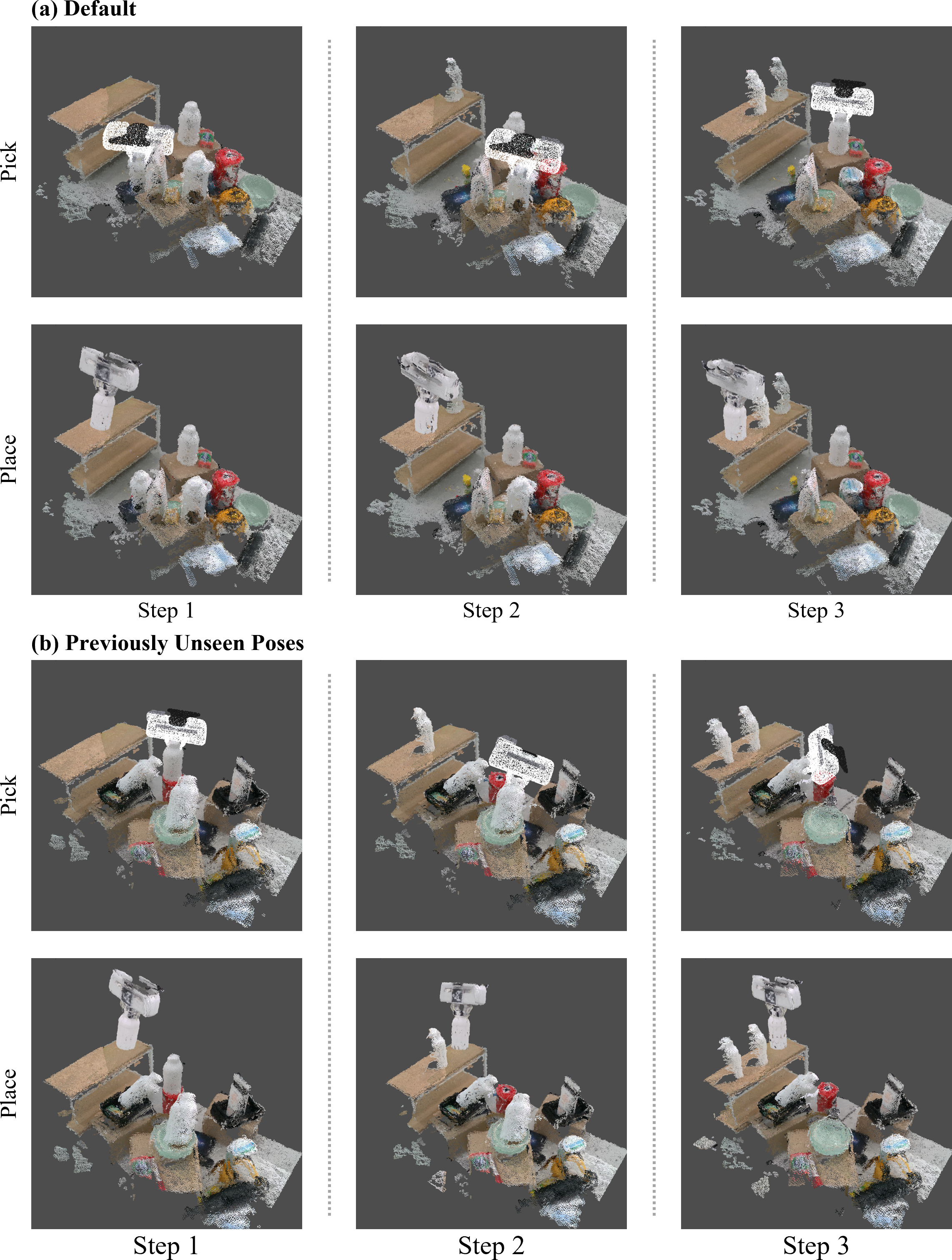}
   \caption{
        \textbf{Samples Generated by Diffusion-EDFs for Real-world Bottles-on-a-shelf Task.} The figure depicts the end-effector pose samples for picking and placing multiples bottles on a shelf. Diffusion-EDFs trained with only four human demonstrations (three sequential subtasks for each) generated these samples from the real-world point cloud observations of the scene and grasp. 
        Similar to our simulation experiments, we experiment for the \textbf{(a)} default scenario and \textbf{(b)} previously unseen target object poses (oblique; note that we only trained Diffusion-EDFs for upright poses) sceneraio.
        The video of the denoising diffusion process can be found in \url{https://sites.google.com/view/diffusion-edfs/home}
   }
   \label{fig:bottle_result}
\end{figure}

\begin{figure}[t]
\vspace{-1.5\baselineskip}
  \centering
   \includegraphics[width=0.6\linewidth]{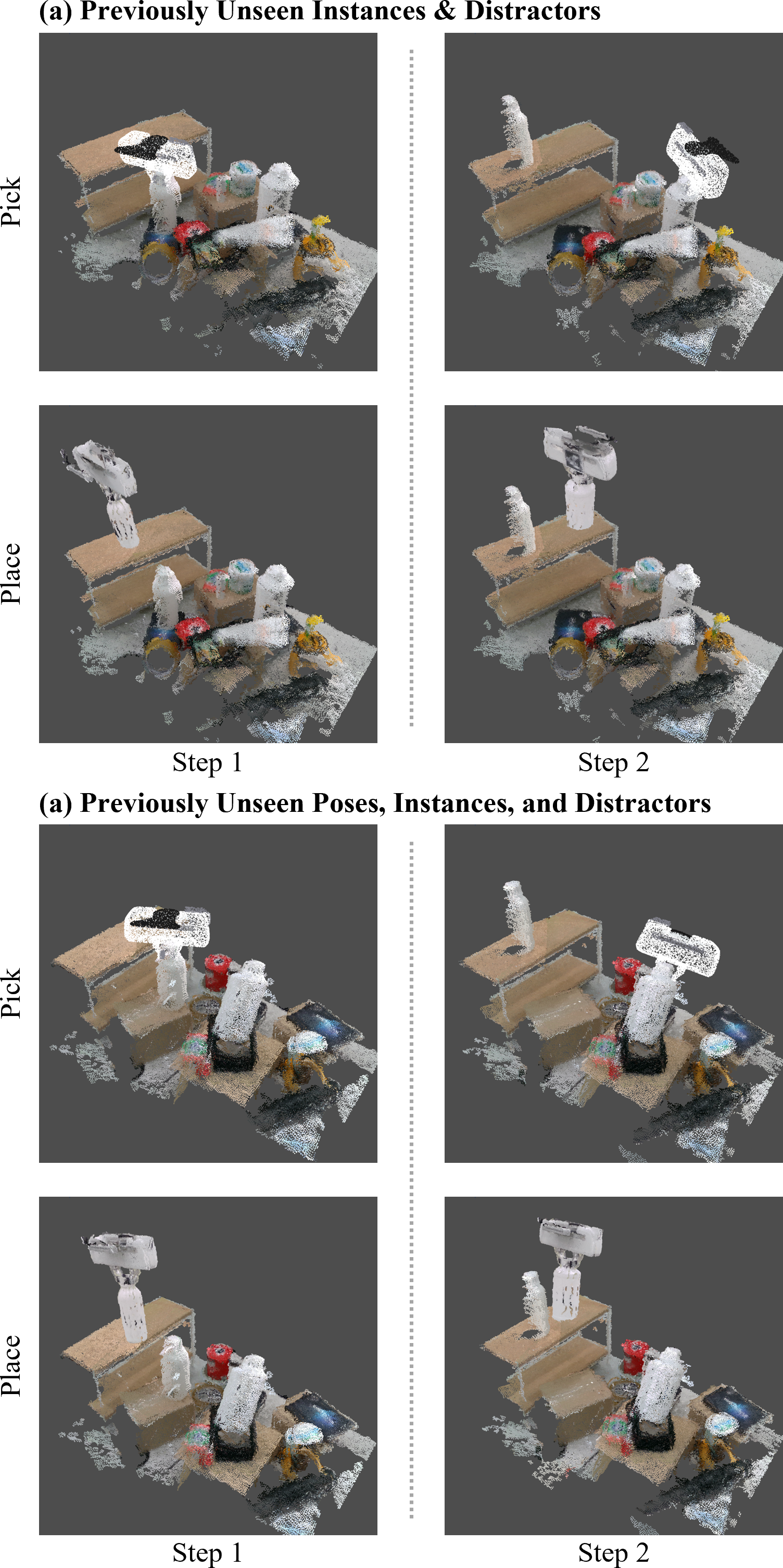}
   \caption{
        \textbf{Samples Generated by Diffusion-EDFs for Real-world Bottles-on-a-shelf Task (Previously Unseen Instances).} The figure depicts the end-effector pose samples for picking and placing multiples bottles on a shelf. 
        In contrast to \Fig~\ref{fig:bottle_result}, we experiment with previously unseen bottle instances.
        Diffusion-EDFs trained with only four human demonstrations (three sequential subtasks for each) generated these samples from the real-world point cloud observations of the scene and grasp. 
        Similar to \Fig~\ref{fig:bottle_result}, we experiment with both the \textbf{(a)} trained poses and \textbf{(b)} previously unseen poses (oblique; note that we only trained Diffusion-EDFs for upright poses).
        The video of the denoising diffusion process can be found in \url{https://sites.google.com/view/diffusion-edfs/home}
   }
   \label{fig:bottle_result_unseen}
\end{figure}

% To provide an idea of Diffusion-EDFs's pure inference performance for noisy real-world observations without the complications related to motion planning and localization, we illustrate the generated samples for the three real-world tasks. 
For more reproducible results, we also provide example input data and codes\footnote{Data and codes can be found in (\url{https://github.com/tomato1mule/diffusion_edf})} that we used to generate end-effector poses for the three real-world tasks with Diffusion-EDFs.
These supplementary materials can provide an idea of Diffusion-EDFs' pure inference performance for noisy real-world observations without the complications related to motion planning and localization.
The samples generated by Diffusion-EDF for the mug-on-hanger and bowls-on-dishes tasks are illustrated in Figs.~\ref{fig:mug_result}~and~\ref{fig:bowl_result_episodes}, respectively.
The samples generated by Diffusion-EDF for the bottles-on-shelf task are illustrated in Figs.~\ref{fig:bottle_result}~and~\ref{fig:bottle_result_unseen}.
% Diffusion-EDFs combined with the energy-based critic in \Sec~\ref{appndx:sys_eng} can successfully infer appropriate poses for all these tasks with more than 90\% of the cases, although this number is subjective to human evaluation.
Diffusion-EDFs combined with the energy-based critic in \Sec~\ref{appndx:sys_eng} can successfully infer appropriate poses for all these tasks in more than 90\% of the cases, although it is important to note that this success rate is subject to human evaluation and may vary based on the individual's criteria.

For mugs and bottles, it takes 5$\sim$6 seconds to generate 20 poses for picking, and 9$\sim$10 seconds to generate 10 poses for placing. For bowls, it takes 7 seconds to generate 20 poses for picking and 17 seconds to generate 10 poses for placing.
The sampling is slower for the bowls-on-dishes task because the point clouds in this task have more points than in the other tasks.
As mentioned in \Sec~\ref{appndx:sampling}, we use two different models for low-resolution and high-resolution denoising. In addition, we use the energy-based critic to sort the sampled poses according to their quality. Therefore, three different models must be trained for each pick and place tasks.
It takes less than 24 minutes to train each model for mug-picking and less than 36 minutes for mug-placing with an RTX3090 GPU.
The bottles-on-a-shelf task requires slightly longer training time, amounting to 27 minutes for picking and 43 minutes for placing with an RTX3090 GPU.
The bowls-on-dishes task requires a much longer training time because it consists of three different subtasks. It takes less than 47 minutes of training for picking and less than 1.3 hours for the placing.
Note that the three models can be trained in parallel.
Therefore, it takes less than an hour with three RTX3090 GPU to train our method for all tasks except for the bowl-placing task.

\end{document}